\documentclass[preprint,12pt]{elsarticle}

\usepackage{amssymb}

\usepackage{amsmath,amsfonts}
\usepackage{algorithm}
\usepackage{algorithmicx}
\usepackage[noend]{algpseudocode}
\usepackage{array}
\usepackage[caption=false,font=normalsize,labelfont=sf,textfont=sf]{subfig}
\usepackage{textcomp}
\usepackage{stfloats}
\usepackage{url}
\usepackage{verbatim}
\usepackage{graphicx}
\usepackage{xcolor}
\usepackage{multirow}

\DeclareMathOperator*{\argmax}{argmax}

\journal{Neurocomputing}

\begin{document}

\begin{frontmatter}




\title{
	SiameseDuo++: Active Learning from Data Streams with Dual Augmented Siamese Networks
	\tnoteref{mytitlenote}}
\tnotetext[mytitlenote]{This paper was supported by the European Research Council (ERC) under grant agreement No 951424 (Water-Futures), the European Union’s Horizon 2020 research and innovation programme under grant agreement No 739551 (KIOS CoE), and the Republic of Cyprus through the Deputy Ministry of Research, Innovation and Digital Policy.}




\author[a]{Kleanthis Malialis\corref{mycorrespondingauthor}}
\cortext[mycorrespondingauthor]{Corresponding author}
\ead{malialis.kleanthis@ucy.ac.cy}

\author[a]{Stylianos Filippou}
\ead{filippou.stylianos@ucy.ac.cy}

\author[a,b]{Christos G. Panayiotou}
\ead{christosp@ucy.ac.cy}

\author[a,b]{Marios M. Polycarpou}
\ead{mpolycar@ucy.ac.cy}

\address[a]{KIOS Research and Innovation Center of Excellence, University of Cyprus, Nicosia, Cyprus}
\address[b]{Department of Electrical and Computer Engineering, University of Cyprus, Nicosia, Cyprus}

\begin{abstract}
Data stream mining, also known as stream learning, is a growing area which deals with learning from high-speed arriving data. Its relevance has surged recently due to its wide range of applicability, such as, critical infrastructure monitoring, social media analysis, and recommender systems. The design of stream learning methods faces significant research challenges; from the nonstationary nature of the data (referred to as concept drift) and the fact that data streams are typically not annotated with the ground truth, to the requirement that such methods should process large amounts of data in real-time with limited memory. This work proposes the SiameseDuo++ method, which uses active learning to automatically select instances for a human expert to label according to a budget. Specifically, it incrementally trains two siamese neural networks which operate in synergy, augmented by generated examples. Both the proposed active learning strategy and augmentation operate in the latent space. SiameseDuo++ addresses the aforementioned challenges by operating with limited memory and limited labelling budget. Simulation experiments show that the proposed method outperforms strong baselines and state-of-the-art methods in terms of learning speed and/or performance. To promote open science we publicly release our code and datasets.
\end{abstract}




\end{frontmatter}


\section{Introduction}
The field of data stream mining, or stream learning, focuses on learning from data which arrive sequentially, typically, in a high-speed manner. The field has been receiving increasing attention in recent years, due to the potential wide applicability of stream learning algorithms in various areas. Examples include, monitoring of critical infrastructure systems (data originated from IoT devices, e.g., sensors) \cite{kyriakides2014intelligent, chang2023attention}, financial monitoring (e.g., data originated from credit card transactions) \cite{dal2015credit}, social media analytics (e.g., data originated from posts and other updates) \cite{wang2018systematic, liu2022multi}, and recommender systems (e.g., data originated from product purchases and reviews) \cite{ditzler2015learning, liu2024integrating, li2024homogeneous}.

Despite significant progress in stream learning, several key challenges exist. The field differs significantly from the traditional learning paradigm as, typically, a large amount of data are produced rapidly, making it challenging to process in real-time. Therefore, stream learning algorithms should operate with limited memory. Importantly, the underlying data distribution is unknown and dynamic, referred to as data nonstationarity or concept drift \cite{ditzler2015learning, gama2014survey, lu2018learning}. Drift can manifest itself in various ways, e.g., seasonality effects and incipient faults can occur in critical infrastructures, while changes in users' interests occur in social media / recommender systems.

Therefore, a learning model should be able to maintain good performance in the presense of concept drift. In doing so though, most existing methods assume that arriving data are annotated with the ground truth (e.g., labels in classification tasks). Active learning \cite{settles2009active} is an effective paradigm to address this challenge. It is very appealing as it involves the human-in-the-loop, specifically, a classifier queries a domain expert for the labels of selected instances. Active learning has been a critical element of many successful applications, e.g., autonomous driving \cite{nvidia} and identifying malicious advertisements \cite{sculley2011detecting}, and it has recently been also embraced by the data mining community \cite{zliobaite2013active, malialis2022nonstationary}.

This work provides new insights into learning from nonstationary and imbalanced data streams under limited availability of labelled data. This constitutes a largely unexplored research area as, typically, the aforementioned challenges are the key reasons which hinder the deployment of stream learning methods in practical applications.
In this work, we propose a method that integrates in a seamless manner three core characteristics; these are, online incremental learning, active stream learning, and data augmentation. Incremental learning helps to address data nonstationarity / concept drift, while active learning and data augmentation help to address the issue of limited availability of labelled data. A table listing the symbols used and a brief description is provided in Table~\ref{tab:symbols}. To promote open science we publicly release our code and datasets\footnote{https://github.com/kmalialis/siameseduo}.

The contributions of this work are as follows:
\begin{itemize}
	\item We propose the SiameseDuo++ method, in which two siamese neural networks operate in synergy to incrementaly learn from data streams to address concept drift. While standard neural networks have been extensively used for incremental learning due to their effective adaptive capabilities, this work pursues a new direction of using specialised neural networks which compare and contrast data points from the stream (similarity learning).
	
	\item SiameseDuo++ uses data augmentation, i.e., synthetic data generation in the latent (rather than the input) space which is incrementally learnt / updated by a Siamese neural network. The encodings learnt from the first siamese network, in conjunction with augmented versions of them, are considered as inputs to a second siamese network, which also learns in an incremental fashion and is responsible for class prediction. Furthermore, a novel density-based active learning strategy is proposed which also operates in the latent space, and it determines when incremental learning will take place. The second siamese network is responsible for the active learning strategy.
	
	\item We conduct an extensive comparative study involving various synthetic and real datasets under concept drift and/or class imbalance. SiameseDuo++ is robust to drift, and operates with limited memory and limited active learning budget. SiameseDuo++ significantly outperforms strong baselines and state-of-the-art methods with respect to learning speed and/or performance.
\end{itemize}

The rest of the paper is structured as follows. Section~\ref{sec:background} provides background material on stream learning, necessary to understand the contributions made by this work. Section~\ref{sec:related} discusses the related work. Section~\ref{sec:method} presents the proposed method. Sections~\ref{sec:exp_setup} and \ref{sec:exp_results} describe the experimental setup and results respectively. Section~\ref{sec:conclusion} concludes the paper.

\begin{table}[t!]
	\centering
	\caption{Nomenclature}
	\label{tab:symbols}
	\resizebox{\columnwidth}{!}{%
		
		\begin{tabular}{ll|lll}
			\textbf{Symbol}           & \textbf{Description}              & \textbf{Symbol} & \textbf{Description}                     &  \\ \cline{1-4}
			$x^t$                     & Arriving example at time $t$      & $Q^t$           & Memory of examples at time $t$           &  \\
			$y^t$                     & Actual class of $x^t$             & $Q^t_{enc}$     & Encoding memory at time $t$              &  \\
			$s_1$                     & First siamese network             & $Q^t_{gen}$     & Augmented encodings memory               &  \\
			$s_1^{enc}$               & Encoder of s\_1                   & $Q^t_{augm}$    & The union of $Q^t_{enc}$ and $Q^t_{gen}$ &  \\
			$s_2$                     & Second siamese network            & $K$             & Number of classes                        &  \\
			$C^t_1$ & Cost function of $s_1$            & $L$             & Number of examples per class in $Q^t$    &  \\
			$C^t_2$ & Cost function of $s_2$            & $c$             & A particular class $c \in \{1, .., K\}$  &  \\
			$s_1(x_i, x_j)$           & Output probability of $s_1$        & $q^t_c$         & A queue within $Q^t$                     &  \\
			$\hat{y}^t$               & Predicted class of $x^t$ by $s_2$ & $x_i$           & An example in $q^t_c$                    &  \\
			$AL$                        & Active learning strategy          & $s_1^{enc}$     & Encoding of $x_i$                        &  \\
			$B$                         & Total budget of AL                & $\alpha_i$      & Augmented encoding                       &  \\
			$b^t$      & AL's labelling spending           & $d(x_i, x_j)$   & Distance between $x_i$ and $x_j$         &  \\
			$G$                         & Set of augmentation functions     & $P^t$           & Pairs created from $Q^t$                 &  \\
			$g_o$                      & An augmentation function          & $P^t_{augm}$    & Pairs created from $Q^t_{augm}$          & 
	\end{tabular}}
\end{table}

\section{Stream Learning}\label{sec:background}
\textbf{Online learning} \cite{wang2018systematic} considers a data generating process that provides a batch of examples $D^t$ at each time step $t$ as follows $D = \{D^t\}_{t=1}^\infty$, where the data are typically sampled from a potentially infinite sequence and $D^t = \{(x^t_i,y^t_i)\}_{i=1}^{|D^t|}$. When the batch size is $|D^t| > 1$ it is termed \textbf{batch-by-batch online} learning. When the batch contains only a single arriving example $|D^t| = 1$, that is, $D = \{(x^t,y^t)\}_{t=1}^\infty$ it is termed \textbf{one-by-one online} learning, also referred to as \textbf{stream learning}. The examples are drawn from an unknown probability distribution $p^{t}(x^t,y^t)$, where $x^t \in \mathbb{R}^d$ is a $d$-dimensional vector in the input space $X \subset \mathbb{R}^d$, $y^t \in \{1, ..., K\}$ is the class label in the target space $Y \subset \mathbb{Z}^+$, and $K \geq 2$ is the number of classes.

This paper focuses on stream learning, which is important for real-time monitoring and decision making. Stream learning requires the model to adapt immediately upon seeing a new example, and algorithms intended for batch-by-batch learning are, typically, not applicable for one-by-one learning tasks \cite{wang2018systematic}. A one-by-one online classifier receives a new instance $x^t$ at time $t$ and makes a prediction $\hat{y}^t$ based on a concept $h: X \to Y$. In \textbf{supervised stream} learning, the classifier receives the true label $y^t$, its performance is evaluated using a loss function and is then trained based on the loss incurred. This process is repeated at each step. The gradual adaptation of the classifier without complete re-training $h^t = h^{t-1}.train(\cdot)$ is termed \textbf{incremental} learning \cite{losing2018incremental, vzliobaite2015towards}.

In tasks involving data streams, however, it is impractical to assume that the ground truth information (i.e., class labels) will be made available at every time step. To address this issue, an alternative approach is \textbf{active} learning \cite{settles2009active}, which deals with strategies to selectively query human experts for labels according to a pre-defined ``budget'' $B \in [0,1]$, e.g., $B=0.01$ means that $1\%$ of the arriving samples can be labelled.

In \textbf{active stream} learning \cite{zliobaite2013active}, a classifier is built that receives a new instance $x^t$ at time $t$. At each time step the classifier calculates the prediction probability $\hat{p}(y | x^t) \in \mathbb{R}^K$ of $x^t$ belonging to each class $y$. The classifier outputs the best prediction probability $h(x^t) = \max_y \hat{p}(y|x^t)$ and the predicted class $\hat{y}^t = \argmax_y \hat{p}(y|x^t)$. An active learning strategy returns True or False to determine whether or not the true label $y^t$ is required respectively, which is assumed that a human will provide. The classifier is evaluated using a loss function and is then trained based on the loss incurred. Training occurs only if the budget allows and when the active learning strategy returns True.

The focus of this article is on active stream learning, where one of its major advantages is that it keeps the human-in-the-loop. Other learning paradigms exist, which are briefly described in Section~\ref{sec:related}, each with their pros and cons, however, they are outside the scope of this work.

A significant challenge encountered in some streaming applications is that of data \textbf{nonstationarity}, typically caused by \textbf{concept drift}, which represents a change in the joint probability \cite{ditzler2015learning, gama2014survey, lu2018learning}. The drift between steps $t_i$ and $t_j$, where $i \ne j$, is defined as:
\begin{equation}
	\quad p^{t_i}(x,y) \neq p^{t_j}(x,y)
\end{equation}

\subsection{Related learning paradigms}
The focus of this work is on stream learning. We discuss below a closely related paradigm, that of online continual learning. Furthermore, within the stream learning area, this work focuses on the use of active learning. Additionally, we describe closely related paradigms to active stream learning, specifically, semi-supervised learning, domain adaptation (a type of transfer learning), and curriculum learning.

\textbf{Online continual learning}. Stream learning aims to learn efficiently from streaming data from one instance at a time. It makes a prediction at each time step, and aims to dynamically adapt to input data distribution changes (or concept drifts). The focus in online continual learning \cite{mai2022online} is to preserve previous knowledge while performing well on the current concept when confronted with concept drift. In online continual learning, the learning process only occurs in a single pass over the data. There have been recent efforts to combine the two in an attempt to find possible synergies \cite{gunasekara2023survey}.

\textbf{Semi-supervised learning}. Methods in this paradigm first learn initial patterns from a set of labelled data, then assigns pseudo-labels to unlabelled data using these patterns (e.g., by considering structure or self-training). Such a method can be iteratively be updated by re-training on both the labelled and the pseudo-labelled data. Examples include \cite{dyer2013compose, khezri2021novel, khezri2020stds}.

\textbf{Domain adaptation}. It is a type of transfer learning which refers to the continual adaption from one (or more) labelled source domain (with or without drifts) to a target domain \cite{yu2022learn}.

\textbf{Curriculum learning}. In this paradigm, a machine learning model is progressively trained from ``easier'' to more ``challenging'' data, an approach which takes inspiration from the meaningful learning order in human curricula. \cite{wang2021survey}.

\section{Related Work}\label{sec:related}
In this section we review related work regarding stream learning and data augmentation.

\subsection{Stream learning}
We start by reviewing work on supervised stream learning, followed by work on active stream learning.

\subsubsection{Supervised stream learning}\label{sec:concept_drift}
Methods to address concept drift are classified as active and passive \cite{ditzler2015learning}. \textbf{Active} methods use explicit mechanisms to detect concept drift, such as, statistical tests (e.g., JIT classifiers \cite{alippi2008justI}) and threshold-based mechanisms (e.g., EDDM \cite{baena2006early}) to compare previous and current performance indicators. \textbf{Passive} methods implicitly address drift using a replay memory or ensembling \cite{krawczyk2017ensemble, gomes2017survey}.

The co-existence of concept drift with class imbalance remains an open research problem \cite{wang2018systematic}. Typically, there are two categories of methods to address imbalance in data streams \cite{aguiar2022survey}. \textbf{Algorithm-level} methods modify directly an algorithm, such as, cost-sensitive learning. \textbf{Data-level} methods, typically, refer to resampling methods, for instance, by using a separate memory per class (e.g., QBR \cite{malialis2018queue}), adaptive rebalancing mechanisms (e.g., AREBA \cite{malialis2020online}), and ensembling (e.g., OOB \cite{wang2015resampling}, ROSE \cite{cano2022rose}, HEEM \cite{siahroudi2021online}, ESOS-ELM \cite{mirza2015ensemble} and GRE \cite{ren2018gradual}).

Typically, while passive methods use incremental learning \cite{elwell2011incremental}, passive methods don’t continually update the classifier but instead perform a complete re-training when drift is detected. Hybrid methods (e.g., \cite{malialis2022hybrid, alippi2017learning}) combine the advantages of both approaches, while an alternative approach is to ``undo'' the effects of concept drift by reverting the data distribution as it had been prior the drift \cite{artelt2022unsupervised}.

The vast majority of the aforementioned methods are effective under the assumption of ground truth availability.
The community has turned into alternative paradigms, such as, semi-supervised learning (e.g., COMPOSE \cite{dyer2013compose} which uses a geometry-based framework), unsupervised learning (e.g., strAEm++DD \cite{li2023autoencoder} which monitors an autoencoder's reconstruction loss to detect drift, and VAE4AS \cite{li2024unsupervised} which monitors the distribution of an autoencoder's encodings to detect drift), and active learning which is the central focus of this paper and is described below.

\subsubsection{Active stream learning}\label{sec:related_ocl_active}
\textbf{Uncertainty sampling}. It is perhaps the most widely used strategy, which queries the most uncertain instances  \cite{settles2009active}. An arriving instance $x^t$ is queried when:
\begin{equation}
	h(x^t) < \theta,
\end{equation}
\noindent where $h(x^t) = \max_y \hat{p}(y|x^t)$ is the best prediction, and $\theta$ is a fixed threshold. This is referred to as fixed uncertainty sampling strategy \cite{zliobaite2013active}. This strategy has been criticised because of the fixed threshold value; if it is set incorrectly, it may allow the classifier's uncertainty to remain above the threshold.

The Randomised Variable Uncertainty Sampling (RVUS) \cite{zliobaite2013active} strategy addresses this limitation. First, it introduces variability so that the threshold changes over time. Second, it introduces randomisation so that the label querying probability is non-zero. Specifically, the threshold is modified as follows:
\begin{equation}\label{eq:rvus}
	\theta =
	\begin{cases}
		\theta (1 - s) & \text{if } v < \theta_{rdm} \text{ \# query label}\\
		\theta (1 + s) & \text{if } v \geq \theta_{rdm} \text{ \# don't query}\\
	\end{cases}
\end{equation}
\noindent where $v = h(x^t)$ is the querying criterion, $s$ is a step size parameter, and $\theta_{rdm} = \theta * \eta$ where $\eta$ follows a Normal distribution $\eta \sim N(1,\delta)$ with a standard deviation of $\delta$. To address catastrophic forgetting and class imbalance, ActiQ \cite{malialis2020data} has been proposed which combines RVUS with a multi-memory component; specifically, one queue per class is maintained to store examples queried by the RVUS strategy. An alternative approach to dynamically tune the threshold is by using meta-learning, where statistical meta-features from adaptive windows are used to meta-recommend a suitable threshold \cite{martins2023meta}.

\textbf{Density sampling}. The concept behind it is that informative instances are those which are representative of the input distribution, i.e., they lie in dense regions of the input space according to a similarity or distance metric. Siamese neural networks have recently been shown to be effective in stream learning.

ActiSiamese \cite{malialis2022nonstationary} considers a multi-memory (one queue per class) with a siamese network. It proposes a density sampling strategy in the latent space, where the querying criterion is the maximum similarity in the predicted class:
\begin{equation}
	v = max_i \ sim(x^t, x_{c, i}),
\end{equation}
\noindent where $sim(\cdot, \cdot)$ denotes a similarity probability, $c$ is the predicted class, and
$x_{c,i} \in q_c$ where $q_c$ is the queue which contains examples of class $c$. It has demonstrated a superior performance under severe class imbalance.

Traditionally, Siamese neural networks (or variants) have been shown to be effective few-shot learners, i.e., capable of learning from few examples per class \cite{koch2015siamese}. The Siamese networks proposed for data stream mining in \cite{malialis2022nonstationary}, have been shown to be effective for two reasons. First, they could handle class imbalance well due to a specific form of oversampling. Second, when used with incremental learning and a memory component, they could learn the concept drift and react to it quickly and effectively.

The proposed SiameseDuo++ significantly differs from ActiSiamese not only because it uses a second siamese network which enables generic augmentation in the latent space, but class prediction and the active learning strategy are now performed in synergy between the two siamese networks.

\textbf{Hybrid sampling and learning paradigms}. Methods have been proposed that combine different active learning strategies to merge their benefits. Examples include DBALStream \cite{ienco2014high} and CogDQS \cite{liu2021online} which, combine uncertainty and density sampling. Furthermore, active stream learning has been used in conjunction with other learning paradigms. For example, density-based stream clustering procedure is used to capture novel concepts with a dynamic threshold, and an effective active label querying strategy to continuously learn the new concepts from the data streams \cite{yan2021clustering}. Another example is the use of both active and semi-supervised learning \cite{dyer2013compose}.

\textbf{Ensemble-based sampling}. Query-by-committee \cite{freund1997selective} is a strategy in which a committee of classifiers predicts the label. The concept behind it is that the most informative query is considered to be the sample which the committee disagrees the most. An example of this method is \cite{korycki2019active}, where uncertain classifiers are temporarily removed by dynamically adjusting an abstaining criterion in favour of minority classes. The Active Weighted Aging Ensemble \cite{wozniak2023active} is a chunk-based method which employs a classifier ensemble approach and utilises the changing ensemble lineup to react to concept drift.

\textbf{Graph active learning}. For graph-structured domains, such as, social networks, molecular structures, and knowledge graphs, an alternative approach is required. Graph representation learning is a learning approach that focuses on encoding graphs into low-dimensional vector spaces while preserving the underlying topology and node relationships, thus enabling the efficient processing of complex networks \cite{wu2020comprehensive}.

The work in \cite{cao2024graph} proposes a novel graph deep active learning framework for data deduplication. A graph active learning strategy is introduced to filter the data that needs to be labelled, which is used to delete duplicate data that retains the most information. The work in \cite{huang2024adaptive} proposes a new active learning approach for graph neural networks that leverages reinforcement learning to select informative nodes for labelling. The method uses mutual information to construct states that capture both the graph structure and node attributes. The work in \cite{ge2025iterative} is concerned with the subgraph matching problem, and proposes a new approach based on a spanning tree that iteratively reduces the solution space by querying nodes in the subgraph.

\subsection{Data augmentation}
Data augmentation increases the size of a dataset by creating artificial variations of its samples to enhance the dataset's diversity, thus improving the performance and generalisation of a learning model \cite{shorten2019survey}.

\textbf{Modality-specific augmentation}. Typically, the process of data augmentation is modality-specific. For example, for images we use techniques which alter the geometrical  \cite{taylor2018improving} and colour  \cite{shorten2019survey} properties of an image, or techniques such as random erasing \cite{zhong2020random}, style transfer \cite{jackson2019style} and image mixing \cite{inoue2018data}. For time series data, representative techniques include window slicing \cite{guennec2016data}, and time warping \cite{umtt2017data}.

\textbf{Generic augmentation}. Despite their effectiveness, the aforementioned need to be manually designed and carefully evaluated, therefore, the community has turned to modality-agnostic data augmentation methods to improve their practical applicability. Examples include \cite{cheung2021modals, kumar2019closer, devries2017dataset} which apply data transformations, e.g., interpolation, extrapolation, and noise injection in the latent space. Our work uses these transformation functions, however, in contrast to these works, SiameseDuo++ falls within the stream learning framework.

\textbf{Augmented-based active learning}. Augmentation has been used in conjunction with traditional active learning, e.g., \cite{tran2019bayesian, hong2020deep, kim2021lada}. Augmentation within the active stream learning framework has also been proposed in \cite{malialis2022data}, however, it concerns modality-specific augmentation. Contrary to all the above, SiameseDuo++ is concerned with generic augmentation for active stream learning in nonstationary environments.

\begin{figure*}[t!]
	\centering
	\includegraphics[scale=0.5]{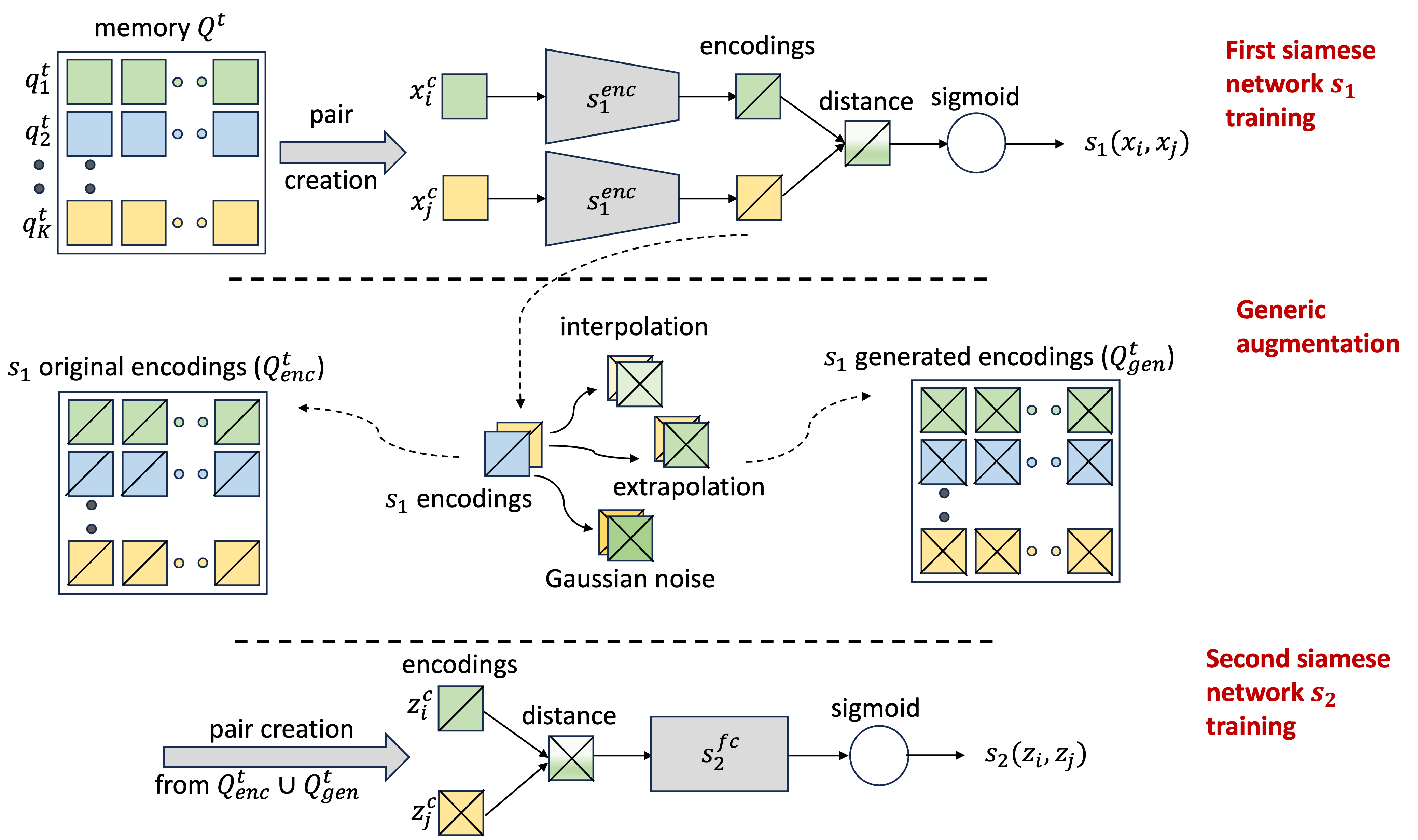}
	
	\caption{An overview of SiameseDuo++: Training process}
	
	\label{fig:siameseduo1}
\end{figure*}

\section{SiameseDuo++}\label{sec:method}
Siamese neural networks are in the backbone of the proposed method. SiameseDuo++ uses two siamese neural networks (hence, the name ``SiameseDuo'') which incrementally learn (hence, the suffix ``++'') from streaming data.

The central concept behind SiameseDuo++ is as follows. The first siamese network is responsible for learning encodings of the data. These encodings are learnt in an incremental fashion, i.e., they are continually updated. Synthetic encodings will be created based on the original encodings. The augmented (original and synthetic) encodings will then be decoded using the first siamese network, and then passed to second siamese network which is resonsible for class prediction.

The overview of SiameseDuo++ regarding the training process is shown in Fig.~\ref{fig:siameseduo1}. In the upper part, on the left we observe the memory $Q^t$  at any time $t$ which contains a queue per class. This is populated by examples for which the active learning strategy queried their label. On the right, we can see the first siamese network $s_1$ which is trained incrementally only when the active learning strategy requests and receives a class label. The middle part depicts the augmentation process in the latent space. The encodings of the first network form the $Q^t_{enc}$ component shown on the left. Transformation functions (such as, interpolation, extrapolation, injecting Gaussian noise) are applied to each encoding to generate the $Q^t_{gen}$ component. The bottom part depicts the second siamese network $s_2$ which is trained incrementally using the original and augmented encodings $Q^t_{enc} \cup Q^t_{gen}$. Like $s_1$, training occurs only when the active learning strategy returns True. 

\begin{figure}[t!]
	\centering
	\includegraphics[scale=0.5]{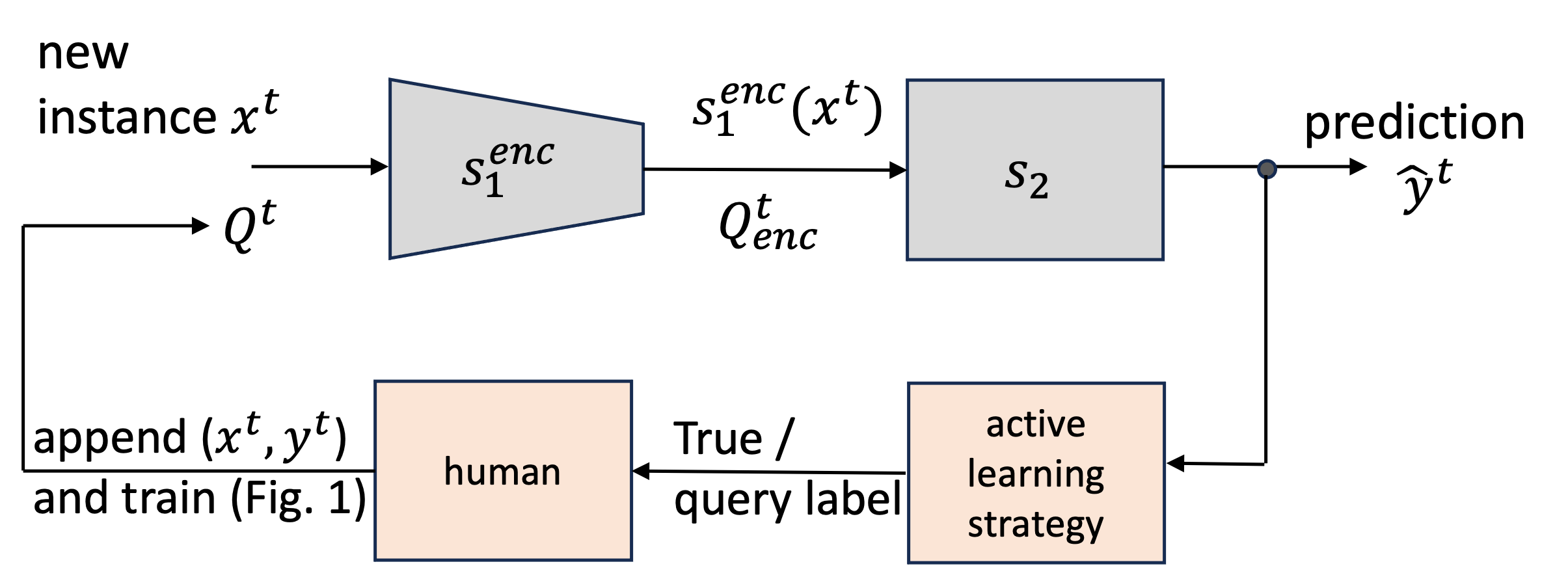}
	
	\caption{An overview of SiameseDuo++: Prediction process and active learning strategy}
	
	\label{fig:siameseduo2}
\end{figure}

The overview of SiameseDuo++ regarding class prediction and the active learning strategy is shown in Fig.~\ref{fig:siameseduo2}. The arriving instance $x^t$, along with the original memory $Q^t$ are first propagated through $s_1$'s encoder to get their encodings, which are then propagated through the second siamese network $s_2$ to predict the class. The prediction probability (or confidence)  is then considered by an active learning strategy to determine if the ground truth will be requested and stored in the memory.

The pseudocode of SiameseDuo++ is shown in Algorithm~\ref{alg:siameseduo}. Each line of the algorithm, as well as each component of Figs. \ref{fig:siameseduo1} and \ref{fig:siameseduo2} are described in detail below.

\subsection{First siamese network}

\textbf{Multi-memory}. Examples queried by the active learning strategy (described later) are stored in a multi-memory. At any time $t$ we maintain $K \geq 2$ first-in-first-out (FIFO) queues of equal size $L \in \mathbb{Z}^+$, one for each class $c \in \{1, ..., K\}$:
\begin{equation}\label{eq:queues}
	\begin{aligned}
		Q^t &= \{q^t_c\}^K_{c=1} \\
		q^t_c &= \{x_{c,i}\}^L_{i=1}
	\end{aligned}
\end{equation}
\noindent where for any two $x_{c,i}, x_{c,j} \in q^t_c$ such that $j > i$, implies that $x_{c,j}$ has been observed more recently in time. In Fig.~\ref{fig:siameseduo1}, each class is shown by a different colour.

The memory $Q^t$ (of size $K \times L$) defined in Eq. (\ref{eq:queues}) can be optionally initialised with historical data, where $K$ is the number of classes and $L$ the number of examples per class.  This is optional as it relies on the availability of even a small amount (say $L=10$) of historical labelled data; this is a reasonable assumption to make in most real-world applications. Alternatively, one should wait until the memory is filled.

\textbf{Pair creation mechanism}. Siamese networks accept pairs of examples. Therefore, pairs are extracted from the multi-memory $Q^t$ to form $P^t$. We create two sets of pairs, the positive set which contains pairs in which the two examples are of the same class, and the negative set which contains pairs in which the two examples are of different classes. To be more effective in addressing class imbalance, we undersample the negative set to ensure the two are balanced.

\textbf{First siamese network}. A siamese network is a special type of neural network which consists of two identical neural networks (known as the ``siamese twins'') \cite{koch2015siamese}. The siamese network learns a function that maps an input pattern into a latent space, thus forming its encoding, in such a way that a distance metric in the latent space approximates the neighbourhood relationships in the original space. In other words, they can perform similarity learning, where they compare and contrast data points from the stream. This is in contrast to traditional models which don't perform similarity learning, and consider neighbourhood relationships in the original input space.

The first network $s_1: X \times X \mapsto [0,1], X \subset \mathbb{R}^d$ in Fig.~\ref{fig:siameseduo1} learns a probability $s_1(x_i, x_j)$ which indicates if the elements of the pair $(x_i,x_j) \in P^t$ belong to the same class. Part of it is the feature learner $s^{enc}_1: X \mapsto Z$ for $Z \subset \mathbb{R}^m$, which maps an input pattern $x \in \mathbb{R}^d$ into a latent space, thus forming its ``encoding'' $s_1^{enc}(x) \in \mathbb{R}^m$, where $d$ and $m$ are the dimensionality of the input and latent spaces respectively. Let $(x_i,x_j)$ be a pair of examples, the distance metric used is the element-wise absolute difference as shown below. The distance is then provided to a sigmoid output unit, as shown in the upper part of Fig.~\ref{fig:siameseduo1}.
\begin{equation}
	d(x_i, x_j) = | s_1^{enc}(x_i) - s_1^{enc}(x_j) |
\end{equation}

\textbf{Incremental learning of first network}. The siamese network is incrementally trained only when the active learning strategy (described in Section~\ref{sec:method_siamese2}) requests the ground truth (i.e., label) of the arriving instance. The cost function $C_1^t$ at time $t$ is provided below:
\begin{equation}\label{eq:cost_s1}
	\begin{aligned}
		C_1^t &= \frac{1}{|P^t|} \sum_{(x_i, x_j) \in P^t} l(y_{i,j}, s_1(x_i, x_j))\\
		&= - \frac{1}{|P^t|} \sum_{(x_i, x_j) \in P^t} y_{i,j} log(s_1(x_i, x_j))\\
		&+ (1 - y_{i,j}) log(1 - s_1(x_i, x_j))
	\end{aligned}
\end{equation}

\noindent where $y_{i,j} \in \{0,1\}$ is the ground truth which states if the pair $(x_i,x_j)$ belongs to the same class ($y_{i,j} = 1$) or not ($y_{i,j} = 0$). The loss function $l$ used is the binary cross-entropy.

Learning is performed using incremental stochastic gradient descent (or any other gradient descent-based algorithm) and backpropagation \cite{rumelhart1986learning}. Let $t$ be the current time, and also the time where the active learning strategy had requested and received the class label. Each network's parameter or weight $w^{t-1}$ is updated according to the formula:
\begin{equation}\label{eq:gd_first}
	w^t \leftarrow w^{t-1} - \xi \frac{\partial C_1^t}{w^{t-1}},
\end{equation}
\noindent where $\frac{\partial C_1^t}{w^{t-1}}$ is the partial derivative with respect to $w^{t-1}$, and $\xi$ is the learning rate.

\subsection{Generic augmentation}

\textbf{Transformation functions}.
Let $x_{c,i} \in q^t_c \subset \mathbb{R}^d$ ($d$ is the dimensionality of the input space) be an example in the multi-memory belonging to class $c$. Let $s_1^{enc}(x_{c,i}) \in \mathbb{R}^m$  ($m$ is the dimensionality of the latent space) and $Q^t_{enc}$ be its encoding. Let $G = \{g_o\}_{o=1}^{|G|}$ be a set of transformation functions $g_o: Z \mapsto Z$, such that, an encoding is augmented to $a_{c,i} = g_o(s_1^{enc}(x_{c,i}))$. Data augmentation occurs in the latent space by applying one (or more) of the following three transformation functions, randomly or in an informed way. This is shown in the middle of Fig.~\ref{fig:siameseduo1}.

The first one is interpolation shown in Eq. (\ref{eq:interpolation}):
\begin{equation}\label{eq:interpolation}
	a_{c,i} = s_1^{enc}(x_{c,i}) + \beta_{1} \big(\epsilon - s_1^{enc}(x_{c,i})\big), 
\end{equation}
\noindent where $\epsilon \in Q^t_{enc}$ is the nearest example of the same class $c$ to $s_1^{enc}(x_{c,i})$ based on some distance metric (e.g., Euclidean or cosine), and $\beta_{1} \in (0,1)$ is a scaling factor.

The second one is extrapolation shown in Eq. (\ref{eq:extrapolation}):
\begin{equation}\label{eq:extrapolation}
	a_{c,i} = s_1^{enc}(x_{c,i}) + \beta_{2} (s_1^{enc}(x_{c,i}) - \mu_c)
\end{equation}
\noindent where $\mu_c \in \mathbb{R}^m$ denotes the class mean $c$ at the given time (also referred to as the class prototype),  and $\beta_{2} \in (0,1)$ is a scaling factor.

The third transformation occurs by injecting noise from a Gaussian distribution with zero mean and per-element standard deviation of examples in the class $c$. This is shown in Eq. (\ref{eq:gaussian_noise}):
\begin{equation}\label{eq:gaussian_noise}
	a_{c,i} = s_1^{enc}(x_{c,i}) + \beta_{3} \eta
\end{equation}
\noindent where $\eta \sim N(0, \sigma^2_c \mathbb{I})$, and $\beta_{3} \in (0,1)$ is a scaling factor.

\textbf{Augmented memory}. All generated examples derived from the previous transformation functions are stored in $Q^t_{gen}$. We merge this memory with the encoded $Q^t_{enc}$ to form the augmented memory as follows:
\begin{equation}\label{eq:augm_memory}
	Q^t_{augm} = Q^t_{enc} \cup Q^t_{gen}
\end{equation}

Based on the pair creation mechanism described earlier, we create the augmented pairs memory $P^t_{augm}$.

\subsection{Second siamese network}\label{sec:method_siamese2}

\textbf{Second siamese network}.
The inputs to the second siamese network are the learnt encodings from the first siamese network. It is defined as $s_2: Z \times Z \mapsto [0,1]$, where $Z \subset \mathbb{R}^m$ and learns a probability $s_2(z_i, z_j)$ which indicates if the elements of the pair $(z_i,z_j) \in P^t_{augm}$ belong to the same class. Unlike the first siamese network, there is no feature extraction component. It directly computes the element-wise absolute difference between the encodings, followed by a standard fully-connected neural network ($s_2^{fc}$), depicted in the bottom part of Fig.~\ref{fig:siameseduo1}.

\textbf{Incremental learning of second network}. Like the first siamese network, it is incrementally trained only when the active learning strategy requests the class label of the arriving instance. The cost function $C_2^t$ at time $t$ is provided below:
\begin{equation}\label{eq:cost_s2}
	C_2^t= \frac{1}{|P^t_{augm}|} \sum_{(z_i, z_j) \in P^t_{augm}} l(y_{i,j}, s_2(z_i, z_j))
\end{equation}
\noindent where $y_{i,j} \in \{0,1\}$ is the actual label and the loss function $l$ used is the binary cross-entropy. Training the second network is done similarly to the first one, analogous to Eq. (\ref{eq:gd_first}).

\textbf{Class prediction}. This stage is shown in Fig.~\ref{fig:siameseduo2}. To predict the class of the arriving instance $x^t$, we consider its encoding $s_1^{enc}(x^t)$, as well as the encodings in $Q^t_{enc}$.

\begin{algorithm}[t!]
	\caption{SiameseDuo++}
	\label{alg:siameseduo}
	\begin{algorithmic}[1]
		
		\Statex \textbf{Input:}
		\Statex $AL$: active learning strategy
		\Statex $B$: labelling budget
		\Statex $K$: number of classes or queues in $Q^t$
		\Statex $L$: length of each queue in $Q^t$
		\Statex $D$: initial labelled data \Comment Optional
		
		\Statex \textbf{Initialisation:}
		\State init queues $Q^0 = FIFO(K, L, D)$ \Comment Eq. (\ref{eq:queues})
		\State init budget expenses $b^0 = 0$
		\State init siamese networks $s_1^0, s_2^0$
		
		\Statex \textbf{Start:}
		\For{each time step $t \in [1, \infty)$}
		\State receive instance $x^t \in \mathbb{R}^d$
		\State predict class $\hat{y}^t$ \Comment Eq.~(\ref{eq:prediction})
		
		\If{$b^{t-1} < B$}\Comment budget suffices
		\State calculate query criterion value $v$ \Comment Eq.~(\ref{eq:al_value})
		\If{$AL(v) == True$} \Comment ground truth request
		\State receive ground truth $y^t$
		\State append example $Q^t = Q^{t-1}.append((x^t,y^t))$
		\State update first network \Comment Eq. (\ref{eq:gd_first})
		\State Create augmented memory $Q^t_{augm}$ \Comment Eq. (\ref{eq:augm_memory})
		\State update second network
		\EndIf
		\EndIf
		
		\State update budget expenses $b^t$ \Comment Eq. (\ref{eq:budget_spending})
		\EndFor
		
	\end{algorithmic}
\end{algorithm}

For each queue in $Q^t_{enc}$, $s_2$ calculates the average similarity of $s_1^{enc}(x^t)$ to its elements. We choose the queue with the highest similarity:
\begin{equation}\label{eq:prediction}
	\hat{y}^t = \argmax_{c \in \{1, ..., K\}} \sum_{i=1}^L s_2(s_1^{enc}(x^t), s_1^{enc}(x_{c,i})),
\end{equation}
\noindent where $K$ is the number of classes, and $x_{c,i} \in Q^t, s_1^{enc}(x_{c,i}) \in Q^t_{enc}$.

Notice that $Q^t_{gen}$ does not take part in the class prediction stage, and that $s_2$ is responsible for the prediction stage as $s_1$'s role is solely for extracting the encodings.

\textbf{Active learning}. SiameseDuo++ uses a novel selection criterion value for the active learning strategy which is the maximum similarity in the predicted class:
\begin{equation}\label{eq:al_value}
	v_{proposed} = \max_{i \in \{1, ..., L\}}s_2(s_1^{enc}(x^t), s_1^{enc}(x_{c,i})),
\end{equation}
\noindent where $L$ is the queue length, and $c$ is the class selected using Eq.~(\ref{eq:prediction}). It then uses a randomised variable sampling scheme, similar to \cite{zliobaite2013active} and described in Eq.~(\ref{eq:rvus}), specifically:
\begin{equation}\label{eq:proposed_scheme}
	\theta =
	\begin{cases}
		\theta (1 - s) & \text{if } v_{proposed} < \theta_{rdm} \text{ \# query label}\\
		\theta (1 + s) & \text{if } v_{proposed} \geq \theta_{rdm} \text{ \# don't query}\\
	\end{cases}
\end{equation}

This prioritises samples with high similarity to existing class representations. At the same time, it incorporates a randomised threshold to ensure variability in queries, thus promoting coverage across different data regions. Furthermore, as our method guarantees a non-zero probability of querying all instances (randomisation element), it maintains adaptation to concept drift over time, ensuring a diverse and representative labelled set.

Lastly, a budget spending mechanism adopted from \cite{zliobaite2013active} ensures that the labelling spending $b^t$ at any time $t$ does not exceed the allocated budget $B$. The labelling spending $b^t$ is approximated by:
\begin{equation}\label{eq:budget_spending}
	\hat{b}^t = \frac{\hat{u^t_W}}{W},
\end{equation}
\noindent where $W$ is a sliding window size, and $\hat{u}^t$ is the approximated number of queried instances within the sliding window, which is updated using $\hat{u}^t = \lambda \hat{u}^{t-1}_W + AL(x^t)$, where $\lambda = \frac{W}{W-1}$ and $AL(x^t)$ is a Boolean value that indicates if the true label for $x^t$ is queried or not. The scheme follows the incremental learning paradigm and it was proved that $\hat{b}$ is an unbiased estimate of $b$ \cite{zliobaite2013active}.

\subsection{Computational aspects}

\textbf{Memory requirements}. The proposed method relies on two permanent memory components, these are, the memory $Q^t$ and the encodings memory $Q^t_{enc}$ (as discussed below, the memory $Q^t_{gen}$ is not a permanent storage component). Each is of size $K \times L$, where $K$ is the number of classes and $L$ is the number of examples per queue. The $K$ parameter is entirely dependent on the task at hand, while SiameseDuo++ has been shown to be effective with a small number of $L$ (e.g., 10).  The space complexity is thus $O(K \times L)$ for memory $Q^t$, and $O(K \times L)$ for $Q^t_{enc}$. The combined complexity remains $O(K \times L)$. Since $L$ is typically small and SiameseDuo++ has a constant memory requirement, these make the proposed method scalable in terms of memory requirements for many streaming applications. In addition to the memory required for storing data, the neural network also requires space to store its weights, which is proportional to the number of model's parameters. Importantly, SiameseDuo++ satisfies one of the most desired properties of learning in nonstationary environments, which is having a fixed amount of memory for any storage \cite{gama2014survey}.

\textbf{Prediction part}. A prediction should be made before the arrival of the next example in the stream, i.e., it has to operate in less than the example arrival time. The prediction part is depicted in Fig.~\ref{fig:siameseduo2}. SiameseDuo++ relies on a single forward propagation step through $s_1^{enc}$ to get $s_1^{enc}(x^t)$. Notice that repeated forward propagation steps from $s_1^{enc}$ are avoided since the encodings are pre-computed in $Q^t_{enc}$. The complexity depends on the neural architecture of $s_2$ (number of neurons per layer, number and type of layers, etc.). Assuming a standard fully-connected neural network for the siamese network $s_2$, the time complexity is simplified to $O(d \times e \times$ $d_{size}^2$), where $d$ is the number of hidden layers each of size $d_{size}$, and $e = K \times L$ is the batch size. Since $d$ and $b$ are typically small in data stream applications, the dominant factor is $d_{size}$, making the method scalable for many high-dimensional problems.

\textbf{Training part}. The proposed method requires two siamese networks to be incrementally trained. First, although desirable, training can span over several time steps and shouldn't necessarily operate in less than the example arrival time (as with the prediction part). Second, SiameseDuo++ has several characteristics which allow it to be trained in real-time in many applications. These are: (i) Training occurs only when the active learning strategy requests the ground truth information. In conjunction with the fact that SiameseDuo++ is effective with a small active learning budget (e.g., 10\%), training is not expected to occur frequently; (ii) The memory $Q^t_{gen}$ is not only created during training, but the augmentations can be created efficiently in a parallel manner; and (iii) Both siamese networks are incrementally trained using a single update step (1 epoch), which also helps to prevent overfitting.

\section{Experimental Methodology}\label{sec:exp_setup}
In this section we describe the datasets used in this study, the compared methods, the evaluation methodology and the performance metrics.

\subsection{Datasets}
Our experimental study considers both synthetic and real datasets.

\subsubsection{Synthetic datasets}
Our study first considers synthetic datasets, which are of lower dimensionality, nevertheless they allow us to examine the behaviour of the compared methods under different characteristics of concept drift, class imbalance levels, and combinations of both.

\begin{figure}[t!]
	\centering
	
	\subfloat[Sea]{\includegraphics[scale=0.16]{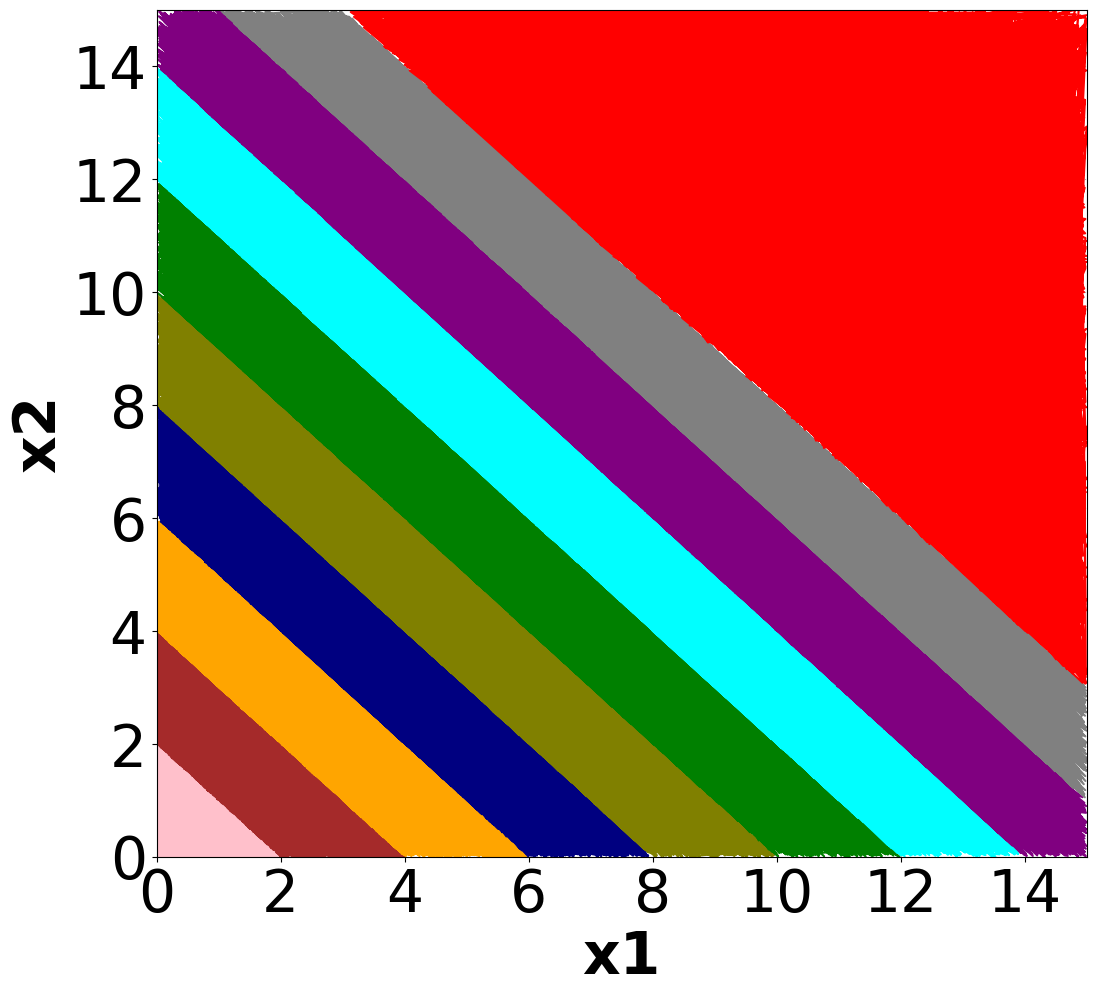}%
		\label{fig:sea}}
	\subfloat[Circles]{\includegraphics[scale=0.16]{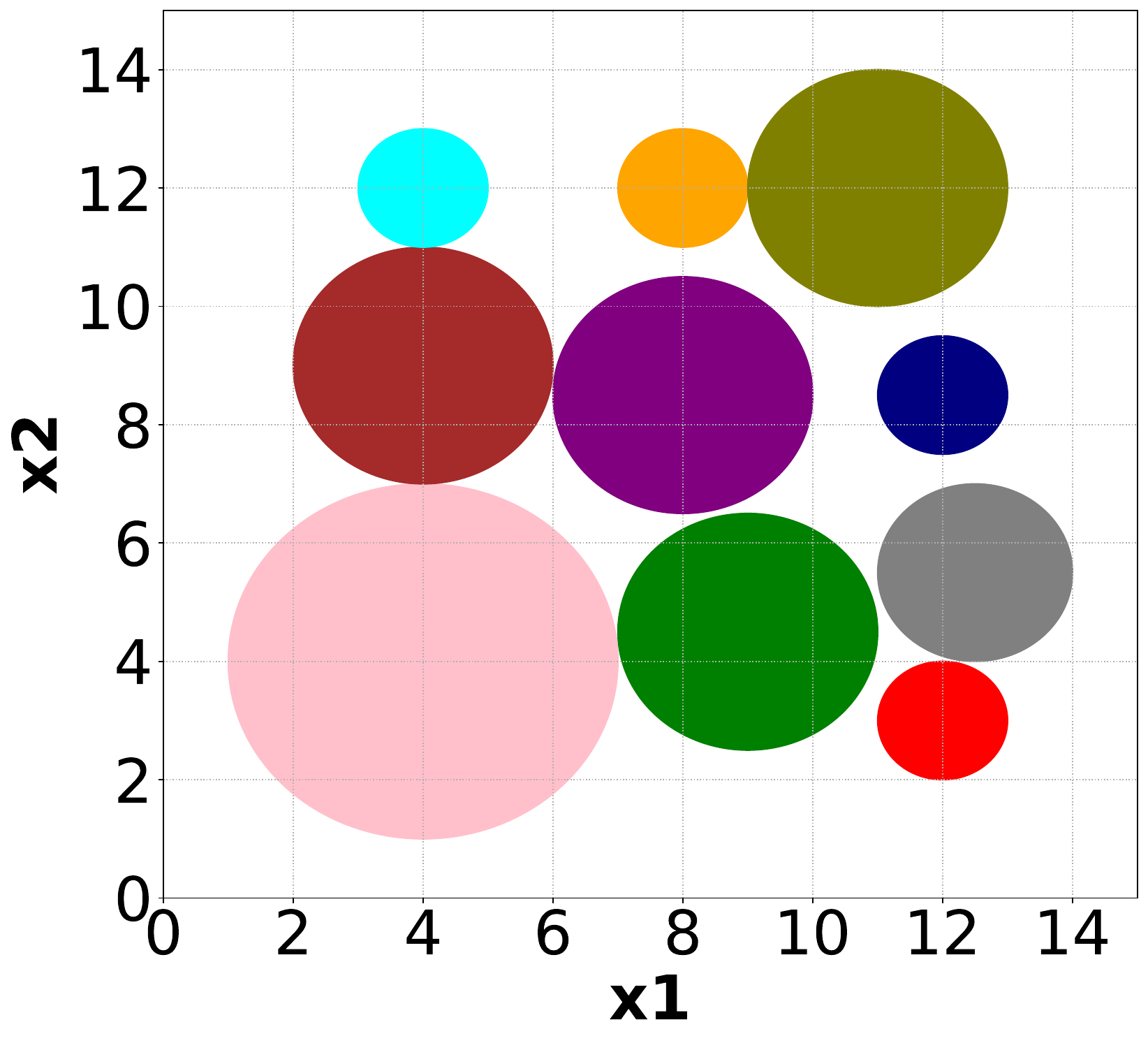}%
		\label{fig:circles}}
	\subfloat[Blobs]{\includegraphics[scale=0.4]{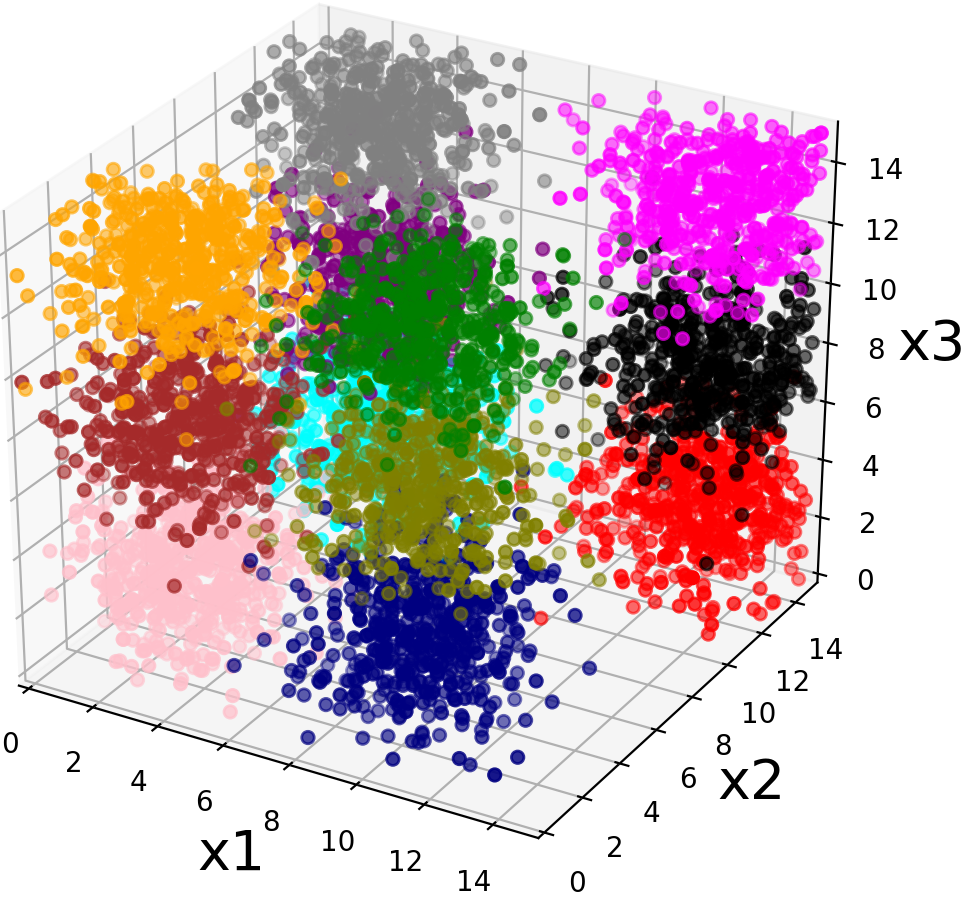}%
		\label{fig:blobs}}
	
	\caption{Synthetic datasets (original)}
\end{figure}

\begin{figure}[t]
	\centering
	
	\subfloat[Sea]{\includegraphics[scale=0.16]{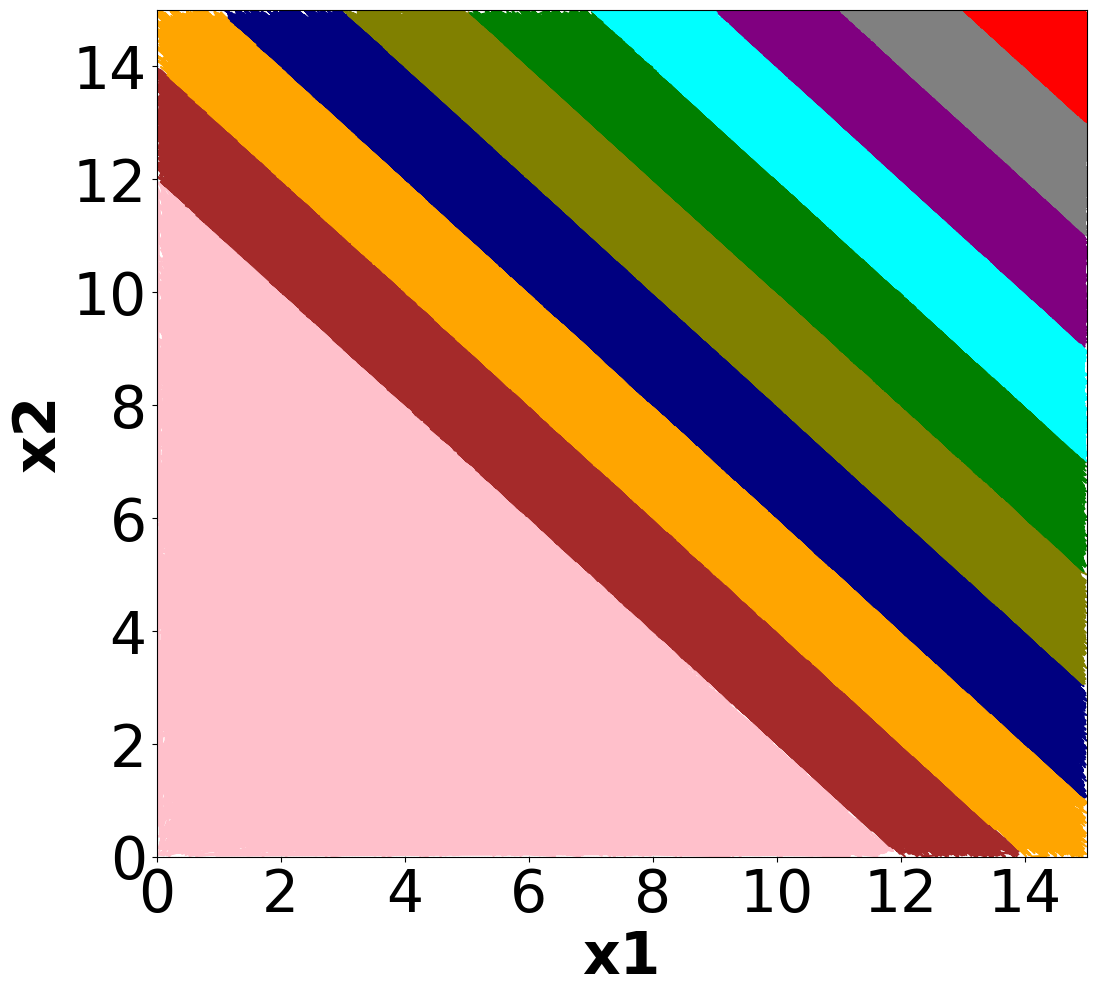}%
		\label{fig:sea_drifted}}
	\subfloat[Circles]{\includegraphics[scale=0.16]{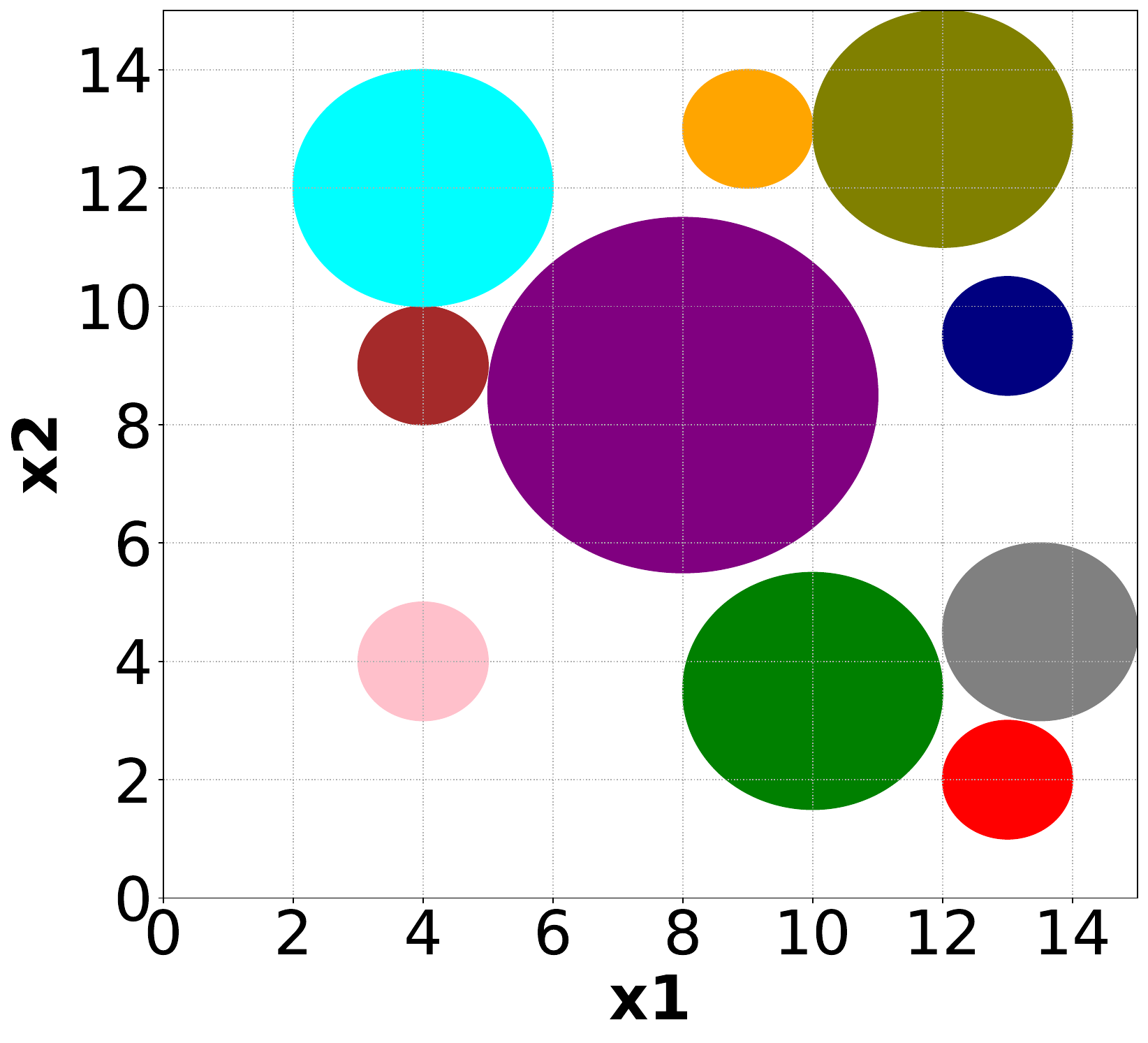}%
		\label{fig:circles_drifted}}
	\subfloat[Blobs]{\includegraphics[scale=0.4]{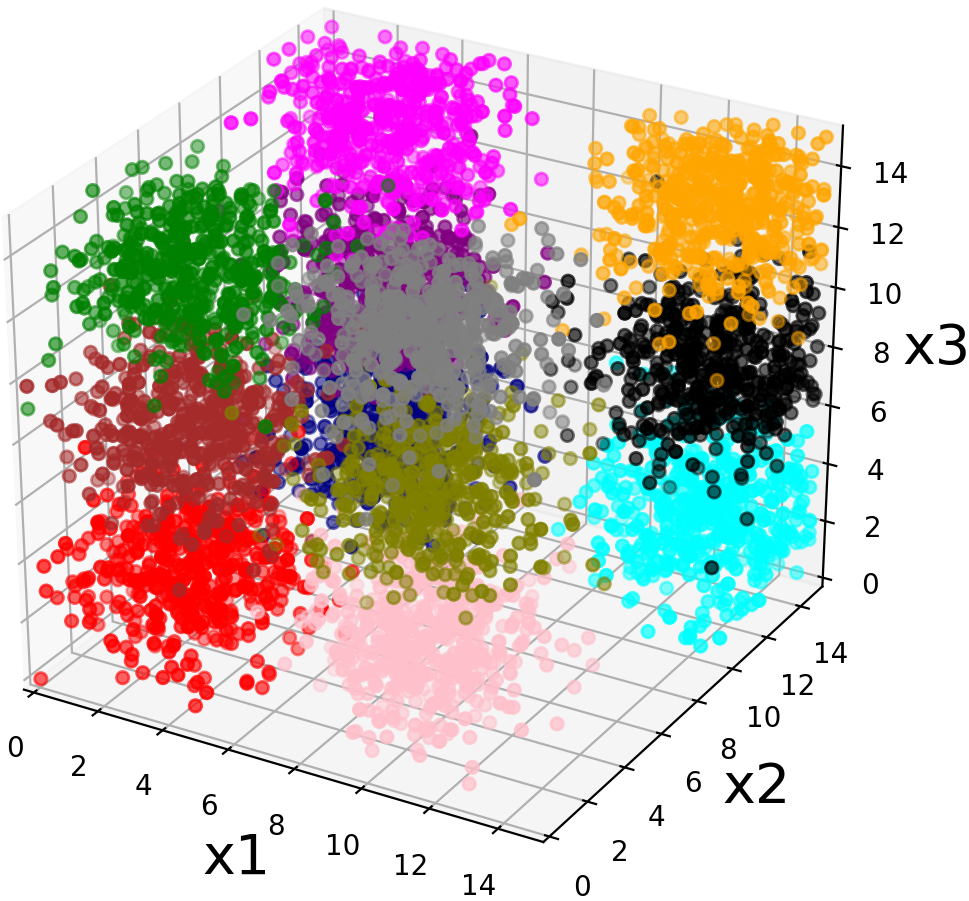}%
		\label{fig:blobs_drifted}}
	
	\caption{Synthetic datasets after concept drift}
\end{figure}

We consider the three datasets, Sea, Circles, and Blobs from \cite{malialis2022nonstationary}, and create five variations of each, i.e., a total of 15 datasets. The three datasets are depicted in Figs.~\ref{fig:sea}-\ref{fig:blobs}, their characteristics are shown in the upper part of Table~\ref{tab:datasets}, and the characteristics of their variations are shown in Table~\ref{tab:datasets_synth}. A detailed description of each is presented below:

\textbf{Sea} \cite{malialis2022nonstationary}. It has two features $x_1, x_2 \in [0, 15]$, and has ten classes as shown in Fig.~\ref{fig:sea}. The drifted version of the dataset suffers from a posterior concept drift as depicted in Fig.~\ref{fig:sea_drifted}. In our experiments, concept drift occurs either abruptly or recurrently, as shown in Table~\ref{tab:datasets_synth}.

\textbf{Circles} \cite{malialis2022nonstationary}. It has two features $x_1, x_2 \in [0,15]$ and ten classes as shown in Fig.~\ref{fig:circles}. The drifted version of the dataset suffers from a posterior concept drift as depicted in Fig.~\ref{fig:circles_drifted}. In our experiments, concept drift occurs either abruptly or recurrently as shown in Table~\ref{tab:datasets_synth}.

\textbf{Blobs} \cite{malialis2022nonstationary}. It has three features $x_1, x_2, x_3 \in [0,15]$ and 12 classes as shown in Fig.~\ref{fig:blobs}. Each class is an isotropic Gaussian blob and noise exists due to the standard deviation of the blobs. The drifted version of the dataset suffers from a posterior concept drift as depicted in Fig.~\ref{fig:blobs_drifted}. In our experiments, concept drift occurs either abruptly or recurrently as shown in Table~\ref{tab:datasets_synth}.

Additionally, we create variations of the synthetic datasets with multi-minority class imbalance; typically, this is considered more challenging compared to multi-majority imbalance \cite{wang2012multiclass}. Specifically, in all cases the pink class is chosen to be the majority class, and the rest constitute minority classes with the same imbalance ratio. The imbalance ratios are shown in Table~\ref{tab:datasets_synth}.

\subsubsection{Real datasets}
Our study also considers real-world datasets. These are significantly more challenging as they are of higher dimensionality (varying from 10 to 1024 dimensions), including time-series in which the arriving data are not assumed to be independent and identically distributed (iid), and the nature of drift may be unknown. All datasets are summarised in the bottom part of Table~\ref{tab:datasets}, and a description of each real-world dataset is provided below.

\textbf{Keystroke} \cite{souza2015data}. More than 50 users were requested to type a specific password which was captured over eight sessions on different days. Ten features were extracted based on the difference between the times when a key is released and when the next one is pressed. The task is to classify four specific users according to their typing rhythm. The number of arriving examples is 1400.

\textbf{Hand-written digits} \cite{misc_optical_recognition_of_handwritten_digits_80}. It has in total 1697 instances, with 10 classes. Each example is an 8x8 image of a hand-written digit, i.e., a feature vector with length 64.

\textbf{uWave gestures} \cite{LIU2009657}. A set of simple gestures generated from accelerometers, each described by 945 features. The data represent 8 classes, with total 4478 samples.

\textbf{Starlight curves} \cite{Rebbapragada_2008}. A starlight curve is a time series of brightness of a celestial object as a function of time. This dataset has in total 9206 phase-aligned starlight curves arriving examples of length 1024. All examples were classified into three classes by an expert.

\textbf{Electromyography} \cite{gestures}. This dataset represents human gestures from electrical activity of muscles (electromyography). To collect the data for each instance, eight sensors are placed on skin surface to collect 64 features. There are four classes with 2500 samples in each class.

\begin{table}[h!]
	\centering
	\caption{Overview of the 20 datasets (15 synthetic, 5 real-world)}
	\label{tab:datasets}
 	\resizebox{\columnwidth}{!}{%
	\begin{tabular}{|c|c|c|c|c|c|}
		\hline
		\textbf{Type}       & \textbf{Dataset}              & \textbf{Modality}   & \textbf{\#Features} & \textbf{\#Classes} & \textbf{\#Arriving} \\ \hline\hline
		\multirow{3}{*}{Synthetic} & Sea (5 variations)      & Tabular      & 2           & 10          & 18000         \\ \cline{2-6} 
		& Circles (5 variations)  & Tabular      & 2           & 10          & 18000         \\ \cline{2-6} 
		& Blobs (5 variational)   & Tabular      & 3           & 12          & 18000         \\ \hline\hline
		\multirow{5}{*}{Real} & Keystroke              & Time-series  & 10          & 4           & 1400          \\ \cline{2-6} 
		& Hand-written digits     & Images       & 64          & 10          & 1697          \\ \cline{2-6} 
		& uWave gestures          & Time-series  & 945         & 8           & 4478          \\ \cline{2-6} 
		& Starlight curves        & Time-series  & 1024        & 3           & 9206          \\ \cline{2-6} 
		& Electromyography        & Time-series  & 64          & 4           & 10000         \\ \hline
	\end{tabular}}
\end{table}

\begin{table}
	\centering
	\caption{Synthetic data variations}
	\label{tab:datasets_synth}
 \resizebox{\columnwidth}{!}{%
	\begin{tabular}{|c|c|c|} 
		\hline
		\textbf{Sea, Circles, Blobs variations}                & \textbf{Time of drift(s)}                                        & \textbf{Imbalance ratio}  \\ 
		\hline\hline
		Original                                               & None                                                             & None                      \\ 
		\hline
		\begin{tabular}[c]{@{}c@{}}Abrupt drift\\\end{tabular} & 3000                                                             & None                      \\ 
		\hline
		\begin{tabular}[c]{@{}c@{}}Imbalance\\\end{tabular}    & None                                                             & 0.1\%     \\ 
		\hline
		Abrupt drift \& Imbalance              & 3000                                                             & 1\%       \\ 
		\hline
		Recurrent drift                                        & \begin{tabular}[c]{@{}c@{}}3000, 6000, 9000, 12000\end{tabular} & None                      \\
		\hline
	\end{tabular}}
\end{table}

\subsection{Methods}

\textbf{Baseline} \cite{zliobaite2013active}. This refers to the seminal work which introduced the RVUS active learning strategy (Section~\ref{sec:related}). The method has been highly influential to the field, it is effective in addressing concept drift, and it is computationally cheap as it performs one-pass learning, i.e., learning occurs on the most recent example. The fact that it is memory-less though, makes it more challenging to address imbalance. The active learning strategy is of type ``uncertainty sampling''. A standard neural network is used.

\textbf{NN} \cite{malialis2020data}. This refers to the ActiQ method (Section~\ref{sec:related}), which builds upon the previous method and proposes the integration of a memory component which allows to address class imbalance, and also improve the performance in scenarios with drift.

\textbf{Siamese} \cite{malialis2022nonstationary}. This refers to the ActiSiamese method (Section~\ref{sec:related}) which proposed for the first time the use of siamese neural networks in online stream learning. With its density sampling active learning strategy it has been shown to achieve state-of-the-art results, particularly under severe class imbalance.

\textbf{SiameseDuo++}. The proposed method described in Section~\ref{sec:method} addressing both class imbalance and concept drift.

Our comparative study examines several critical aspects of stream learning, such as, one-pass learning (Baseline) vs memory-based (NN, Siamese, SiameseDuo++), standard (Baseline, NN) vs siamese (Siamese, SiameseDuo++) neural network, uncertainty (Baseline, NN) vs density (Siamese, SiameseDuo++) sampling, and without (Baseline, NN, Siamese) vs with (SiameseDuo++) data augmentation. These characteristics are summarised in Table~\ref{tab:methods}.

Given also the diverse set of dataset characteristics, it will allow us to examine the behaviour of the compared methods under various conditions. For example, memory-less methods will be stress tested under conditions of imbalance, memory-based methods will be tested on how quickly they can re-act to drift (particularly to recurrent drift), more complex models (e.g., siamese-based) will be tested under streaming conditions, etc.

\begin{table}
	\centering
	\caption{Key features of compared methods}
	\label{tab:methods}
	 \resizebox{\columnwidth}{!}{%
	\begin{tabular}{|c|c|c|c|c|c|c|}
		\hline
		\textbf{Method} & \textbf{Memory} & \textbf{Neural network} & \textbf{Active learning strategy} & \textbf{Augmentation} & \textbf{Concept drift} & \textbf{Class imbalance} \\
		\hline\hline
		\textbf{Baseline}     & No (one-pass learning) & standard network        & uncertainty sampling              & No                    & Yes                    & No                       \\
		\textbf{NN}           & Yes                    & standard network        & uncertainty sampling              & No                    & Yes                    & Yes                      \\
		\textbf{Siamese}      & Yes                    & one siamese network     & density sampling                  & No                    & Yes                    & Yes                      \\
		\textbf{SiameseDuo++} & Yes                    & two siamese networks    & density sampling                  & Yes                   & Yes                    & Yes                      \\
		\hline
	\end{tabular}}
\end{table}

Furthermore, we assume the initial availability of 10 examples per class. For the memory-based methods (NN, Siamese and SiameseDuo++), their memory component was initialised by this initial labelled set, although no pre-training is performed and learning starts from the first time step. Furthermore, hyper-parameter tuning has been performed using this initial labelled set. While some parameters are indeed tailored to specific datasets, SiameseDuo++ appears to be a robust method overall. While we provide all implementation details in our released code, for completeness, we briefly mention them here:
\begin{itemize}
	\item Tables \ref{tab:hyperparams} and \ref{tab:hyperparamsDuo} show the hyper-parameters of the first and second siamese network respectively. Irrespective of the dataset used, most hyper-parameters remain the same. Importantly, the number of epochs which is crucial for incremental learning, but also the mini-batch size, L2 rate, weight initialiser, optimiser, hidden activation function, and the number of hidden layers.
	
	\item The learning rate is set either to 0.01 or 0.001 for all synthetic and real-world datasets, for both siamese networks. Moreover, in Table~\ref{tab:hyperparams}, the number of layers remains fixed to two, and the number of neurons ranges between [32, 32] and [512, 512]. Similarly in Table~\ref{tab:hyperparamsDuo}, there is only a single layer, and the number of neurons is either [16] or [32].
	
	\item For SiameseDuo++'s augmentation transformations, we have observed that lower values were more effective, therefore, for robustness we fix $\beta_1 = \beta_2 = \beta_3 = 0.1$ in all datasets and experiments. For interpolation, the cosine distance is used.
	
	\item Lastly, apart from the active learning strategy's querying criterion, all methods share the same values for the rest of the strategy's hyper-parameters; these are, step size $s=0.01$, randomisation threshold $\delta = 1.0$ and sliding window size $w=300$, as suggested by \cite{zliobaite2013active}.
\end{itemize}

\begin{table}[]
		\centering
	\caption{\label{tab:hyperparams} Hyper-parameter values of Baseline, NN, Siamese and SiameseDuo++'s first network}
	\label{table:hyperparams}
	\resizebox{\columnwidth}{!}{%
	\begin{tabular}{c|cccccccc|}
		\cline{2-9}
		& \multicolumn{3}{c|}{\textbf{Synthetic}}                                                                        & \multicolumn{5}{c|}{\textbf{Real}}                                                                                                                                                                                      \\ \cline{2-9} 
		& \multicolumn{1}{c|}{\textbf{Sea}} & \multicolumn{1}{c|}{\textbf{Circle}} & \multicolumn{1}{c|}{\textbf{Blobs}} & \multicolumn{1}{c|}{\textbf{Keystroke}} & \multicolumn{1}{c|}{\textbf{Digits}} & \multicolumn{1}{c|}{\textbf{Starlight}} & \multicolumn{1}{c|}{\textbf{uWave}} & \textbf{Electromyog.} \\ \hline
		\multicolumn{1}{|c|}{\textbf{Learning rate}}      & \multicolumn{3}{c|}{0.01}                                                                                      & \multicolumn{5}{c|}{0.001}                                                                                                                                                                                              \\ \hline
		\multicolumn{1}{|c|}{\textbf{Hidden layers}}      & \multicolumn{3}{c|}{{[}32, 32{]}}                                                                              & \multicolumn{3}{c|}{{[}128, 64{]}}                                                                                                           & \multicolumn{1}{c|}{{[}512, 128{]}}          & {[}512, 512{]}            \\ \hline
		\multicolumn{1}{|c|}{\textbf{Mini-batch size}}    & \multicolumn{8}{c|}{64}                                                                                                                                                                                                                                                                                                                  \\ \hline
		\multicolumn{1}{|c|}{\textbf{L2 regulariser}}     & \multicolumn{8}{c|}{0.0001}                                                                                                                                                                                                                                                                                                              \\ \hline
		\multicolumn{1}{|c|}{\textbf{Weight initialiser}} & \multicolumn{8}{c|}{He Normal \cite{he2015delving}}                                                                                                                                                                                                                                                                      \\ \hline
		\multicolumn{1}{|c|}{\textbf{Optimiser}}          & \multicolumn{8}{c|}{Adam \cite{kingma2014adam}}                                                                                                                                                                                                                                                                          \\ \hline
		\multicolumn{1}{|c|}{\textbf{Hidden activation}}  & \multicolumn{8}{c|}{Leaky ReLU(0.01) \cite{maas2013rectifier}}                                                                                                                                                                                                                                                           \\ \hline
		\multicolumn{1}{|c|}{\textbf{Num. of epochs}}     & \multicolumn{8}{c|}{1}                                                                                                                                                                                                                                                                                                                   \\ \hline
		\multicolumn{1}{|c|}{\textbf{Output activation}}  & \multicolumn{8}{c|}{Sigmoid (Siamese, SiameseDuo++) / Softmax}                                                                                                                                                                                                                                                                           \\ \hline
		\multicolumn{1}{|c|}{\textbf{Loss function}}      & \multicolumn{8}{c|}{Binary (Siamese, SiameseDuo++) / Categorical cross-entropy}                                                                                                                                                                                                                                                          \\ \hline
	\end{tabular}}
\end{table}

\begin{table}[]
		\centering
	\caption{\label{tab:hyperparamsDuo} Hyper-parameter values of SiameseDuo++'s second network}
	\label{table:hyperparamsDuo}
	\resizebox{\columnwidth}{!}{%
	\begin{tabular}{c|cccccccc|}
		\cline{2-9}
		& \multicolumn{3}{c|}{\textbf{Synthetic}}                                                                        & \multicolumn{5}{c|}{\textbf{Real}}                                                                                                                                                                                      \\ \cline{2-9} 
		& \multicolumn{1}{c|}{\textbf{Sea}} & \multicolumn{1}{c|}{\textbf{Circle}} & \multicolumn{1}{c|}{\textbf{Blobs}} & \multicolumn{1}{c|}{\textbf{Keystroke}} & \multicolumn{1}{c|}{\textbf{Digits}} & \multicolumn{1}{c|}{\textbf{uWave}} & \multicolumn{1}{c|}{\textbf{Starlight}} & \textbf{Electromyog.} \\ \hline
		\multicolumn{1}{|c|}{\textbf{Learning rate}} & \multicolumn{8}{c|}{0.001}                                                                                                                                                                                                                                                                                                               \\ \hline
		\multicolumn{1}{|c|}{\textbf{Hidden layers}} & \multicolumn{3}{c|}{{[}16{]}}                                                                                  & \multicolumn{1}{c|}{{[}32{]}}           & \multicolumn{4}{c|}{{[}16{]}}                                                                                                                                                 \\ \hline
	\end{tabular}}
\end{table}

\subsection{Evaluation}

\subsubsection{Performance metrics}
In balanced datasets the standard metric classifiers are evaluated against is the accuracy metric. However, in imbalanced datasets it tends to become biased towards the majority class (or classes); therefore, the accuracy metric becomes unsuitable in scenarios with class imbalance \cite{he2008learning}.

\textbf{G-mean}. A widely accepted metric which is less sensitive to class imbalance is the geometric mean \cite{sun2006boosting}: 
\begin{equation}\label{eq:gmean}
	G\text{-}mean = \displaystyle\sqrt[K]{\prod_{c=1}^K r_c},
\end{equation}
\noindent where $r_c$ is the recall of class $c$ and $K$ is the number of classes. In addition, this metric has the desirable property that it is high when all recalls are high and their difference is small.

The popular \textit{prequential evaluation with fading factors} method \cite{gama2013evaluating} is used to assess the performance of the compared methods. The method offers two great advantages and, therefore, it is widely adopted by the community. Specifically, it has been proven to converge to the Bayes error when learning in stationary data, and that it does not require a holdout set as the predictive algorithm is tested on unseen examples. The fading factor is set to $0.99$.

\textbf{PMAUC}. Another popular metric is the Area Under the Curve (PMAUC), which is invariant to changes in class distribution and provides a statistical interpretation for scoring classifiers \cite{aguiar2024survey}. The Prequential Multi-class AUC (PMAUC) version is calculated as follows \cite{wang2020auc}.
\begin{equation}\label{eq:auc}
	PMAUC = \frac{2}{K (K - 1)} \sum_{i < j} A(i, j),
\end{equation}
\noindent where $A(i, j)$ is the pairwise AUC when treating class i as the positive class and class j as negative, and $K$ is the number or classes. Prequential evaluation using a sliding window of size $500$ was used to calculate it.

The two metrics complement each other as the G-mean can better reflect the performance on minority classes, while the AUC better reflects the performance on majority classes \cite{wang2020auc}.

\subsubsection{Statistical analysis}
All experiments are repeated ten times, and we provide the average prequential performance. Also, on plots we present the error bars displaying the standard error around the mean, while in tables we provide the standard deviation. Furthermore, we test for statistical significance using a one-way repeated measures ANOVA and then using posthoc multiple comparisons tests with Fisher’s least significant difference correction procedure to show which of the compared method is significantly different from the others.

\section{Experimental Results}\label{sec:exp_results}
This section consists of two parts, an empirical analysis of the proposed method, and a comparative study of SiameseDuo++ to existing methods.

\subsection{Empirical analysis of SiameseDuo++}
\textbf{Active learning budget}. The active learning budget is critical to the success of a real-world application as it essentially determines the amount of data to be labelled. On one hand, a budget too large would result in a larger amount of labelled data which is likely to improve the performance. On the other hand, it would correspond to too frequent interactions with a human expert, i.e., requesting ground truth information too often, which could potentially render a method impractical. This constitutes an important trade-off. Furthermore, the budget will vary according to the data characteristics. Some datasets may require a larger budget to match the performance of a fully supervised approach due to factors like complex decision boundaries, and early biases from small initial labelled data. Moreover, a small budget can cause an AL strategy to struggle to sample enough minority class instances (class imbalance). Moreover, frequent changes in data distribution (concept drift) may necessitate a larger budget for continuous adaptation.

\begin{figure*}[t!]
	\centering
	
	\subfloat[t = 8000]{\includegraphics[scale=0.16]{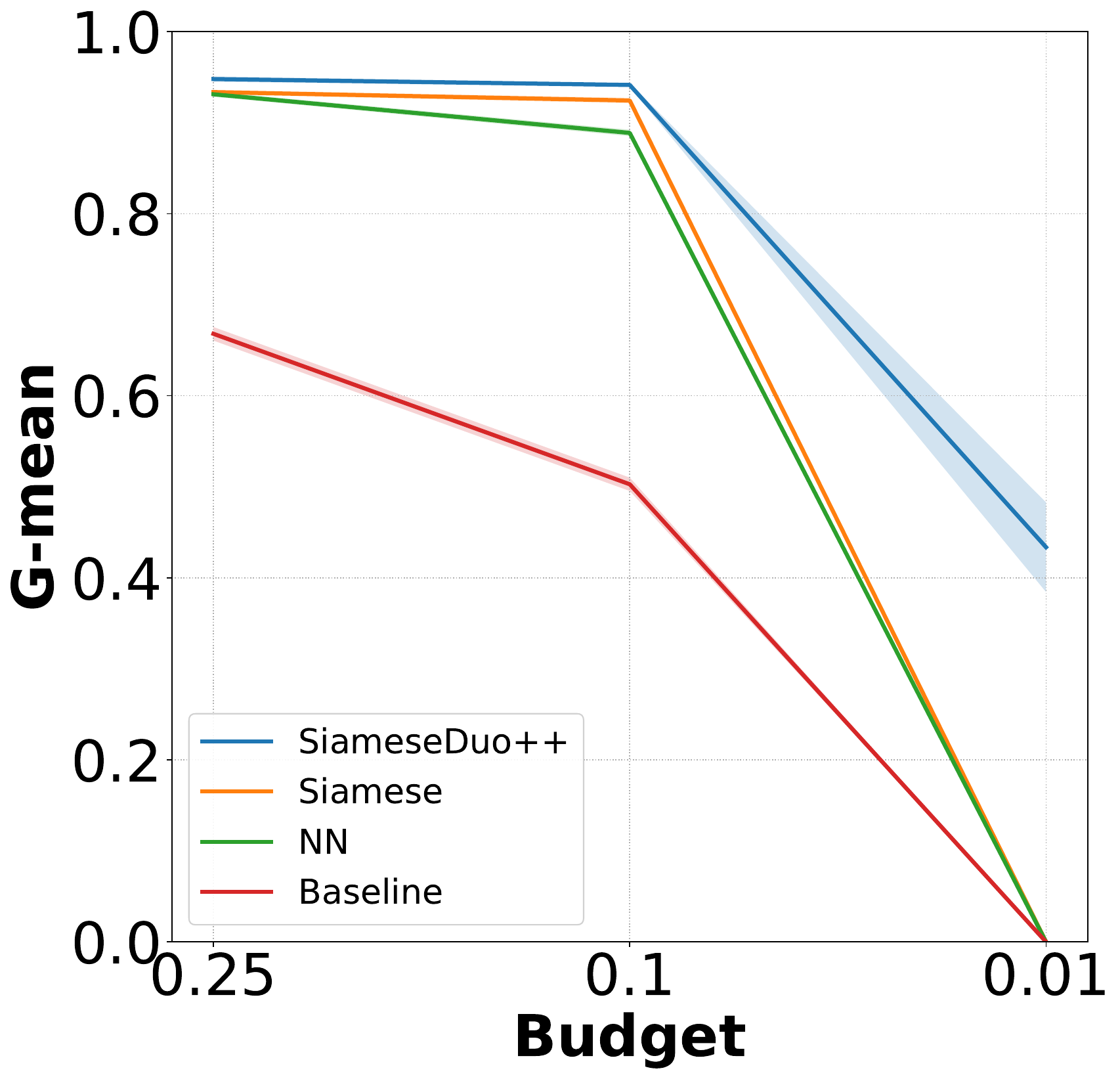}%
		\label{fig:budget_time8k}}
	\subfloat[t = 12000]{\includegraphics[scale=0.16]{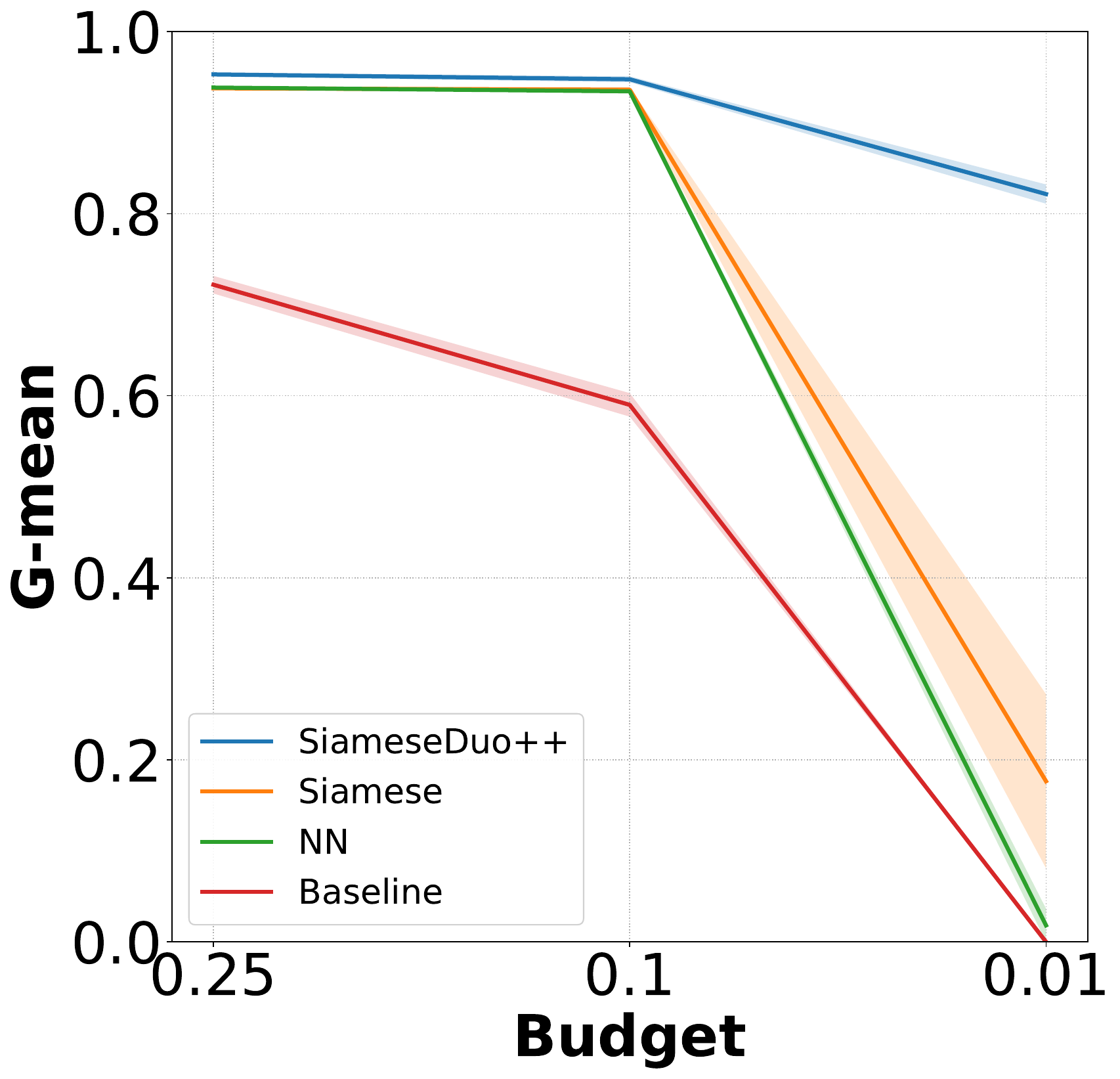}%
		\label{fig:budget_time12k}}
	\subfloat[t = 18000]{\includegraphics[scale=0.16]{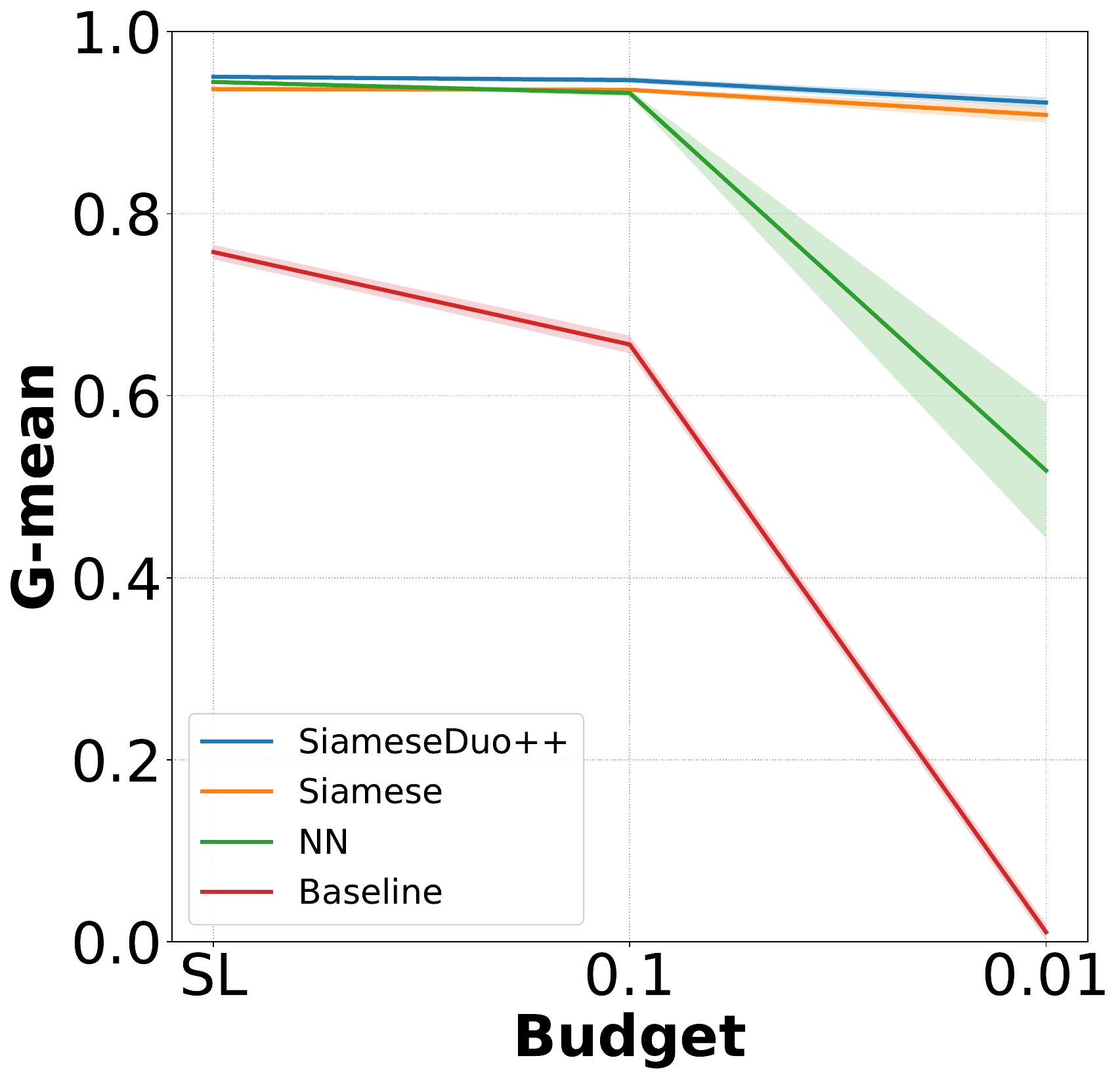}%
		\label{fig:budget_time18k}}
	
	\caption{Performance with different active learning budgets (25\%, 10\%, 1\%) at time steps t = 8000, 12000, 18000 in the Sea dataset. Concept drift occurred at t = 3000.}
\end{figure*}

In this experimental series we examine the robustness of methods with different active learning budgets  (25\%, 10\%, 1\%). We consider the dataset Sea with abrupt drift at time step t = 3000. Figs.~\ref{fig:budget_time8k}, \ref{fig:budget_time12k}, and \ref{fig:budget_time18k} display the performance at times t = 8000, 12000, and 18000 (last time step) respectively. SiameseDuo++ is significantly more robust to the choice of active learning budget compared to the rest. Notable cases are at t = 8000 (Fig.~\ref{fig:budget_time8k}) with budget 1\% where the performance of SiameseDuo++ is around 45\% while for the rest is 0\%. Similarly at t = 12000 (Fig.~\ref{fig:budget_time12k}), the performance of SiameseDuo++ is around 80\%, of Siamese is around 20\%, while for the rest is 0\%. At the last time step (Fig.~\ref{fig:budget_time18k}), reducing the budget from 25\% to 1\% only slightly affects the performance for SiameseDuo++ and Siamese. 

\begin{figure*}[h!]
	\centering
	
	\subfloat[Drift (sea)]{\includegraphics[scale=0.15]{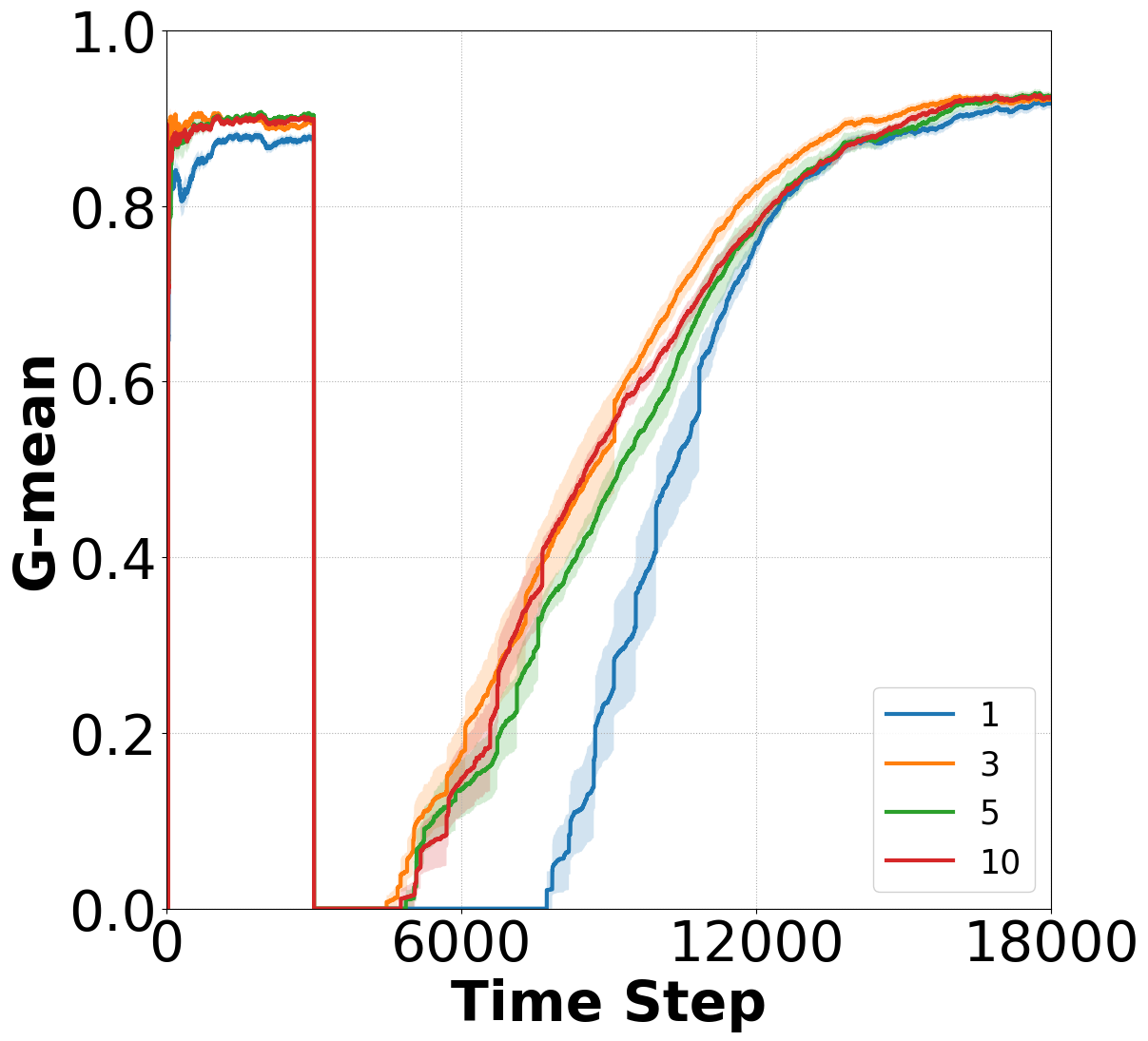}%
		\label{fig:num_sea_abrupt}}
	\subfloat[Drift (circles)]{\includegraphics[scale=0.15]{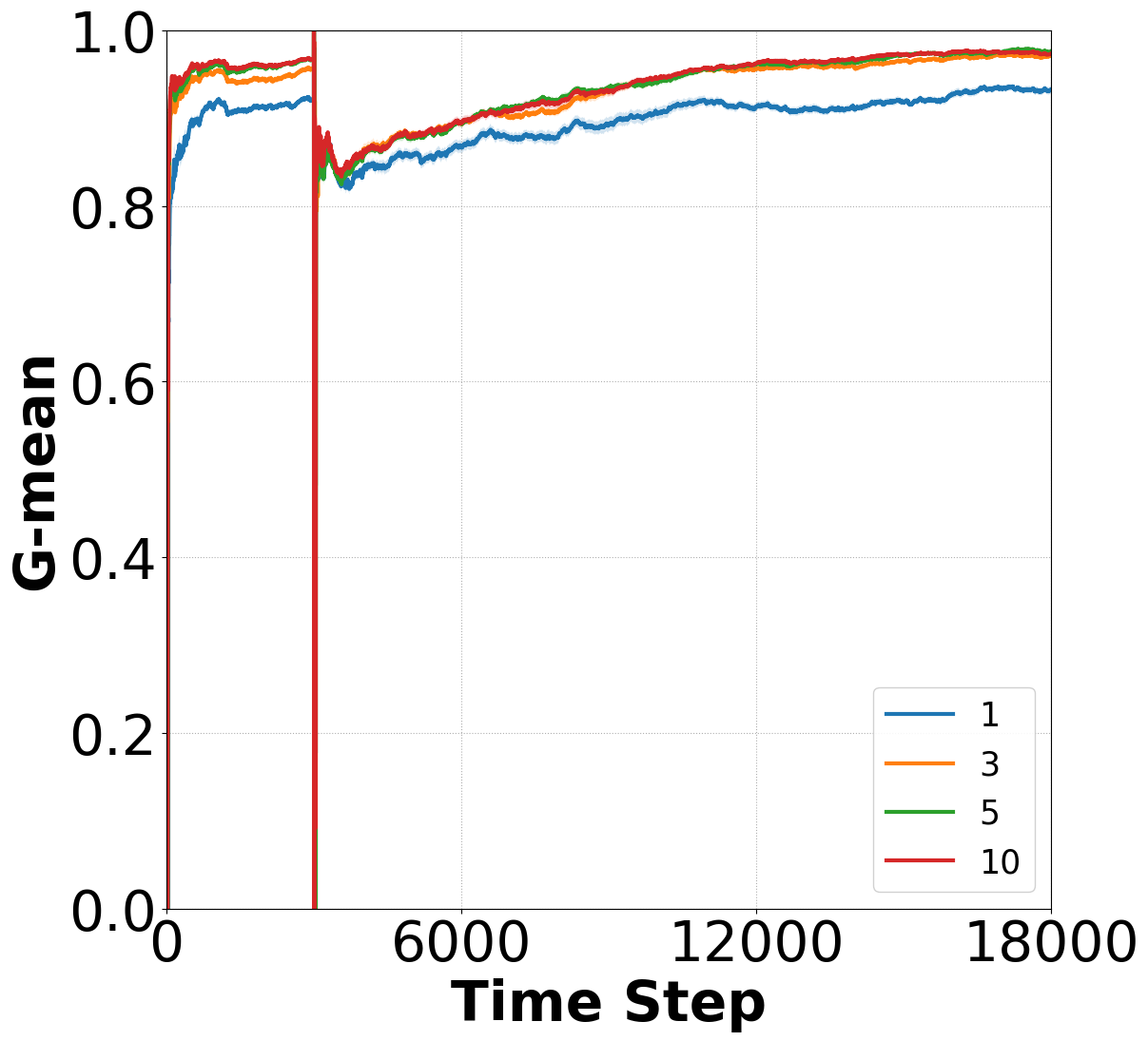}%
		\label{fig:num_circles_abrupt}}
	\subfloat[Drift (blobs)]{\includegraphics[scale=0.15]{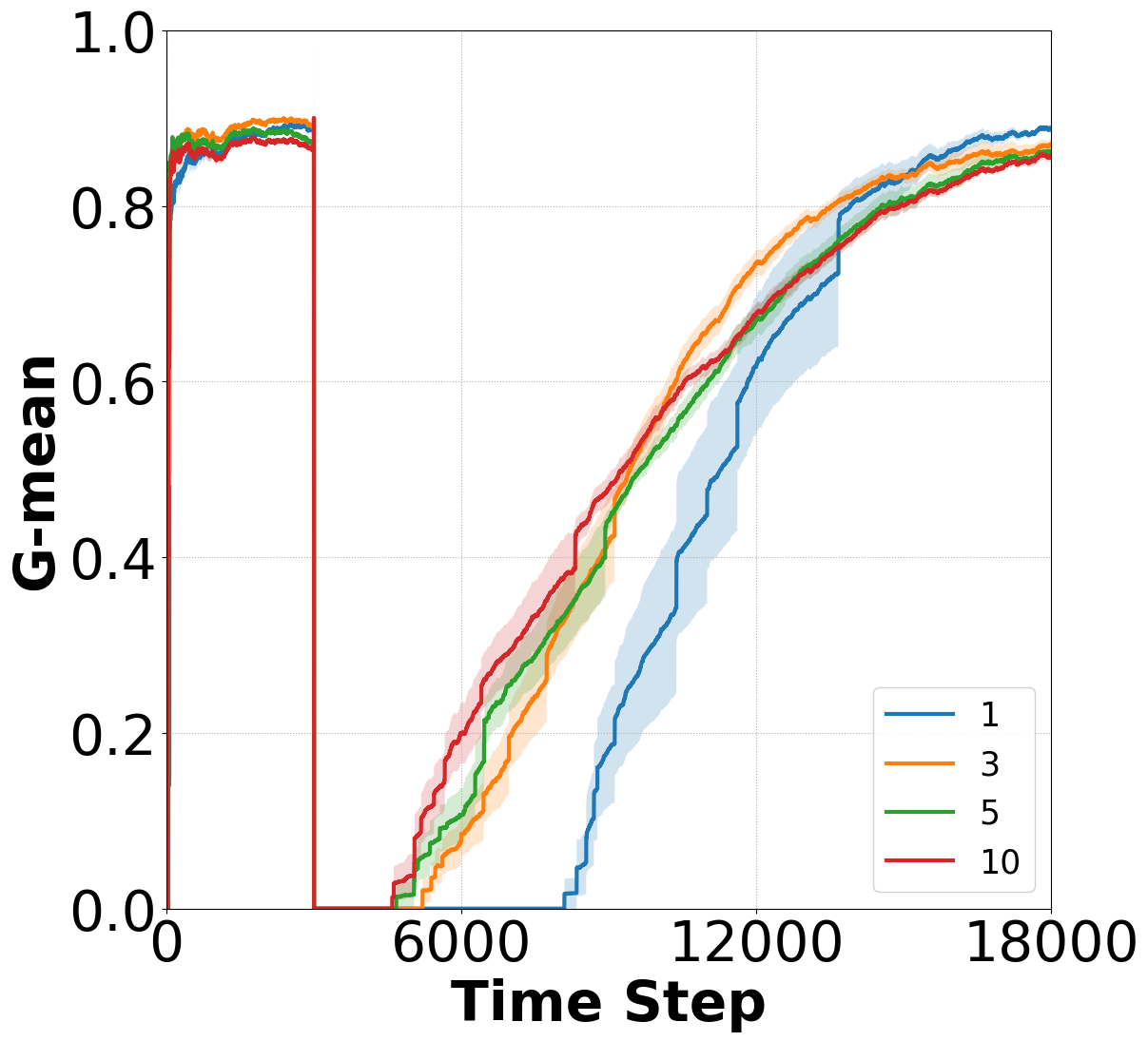}%
		\label{fig:num_blobs_abrupt}}
	
	\caption{SiameseDuo++'s performance with a different number (1, 3, 5, 10) of augmentations per example using interpolation in the three synthetic datasets (Sea, Circles, Blobs) with concept drift.}
\end{figure*}

\begin{figure*}[h!]
	\centering
	
	\subfloat[Sea]{\includegraphics[scale=0.15]{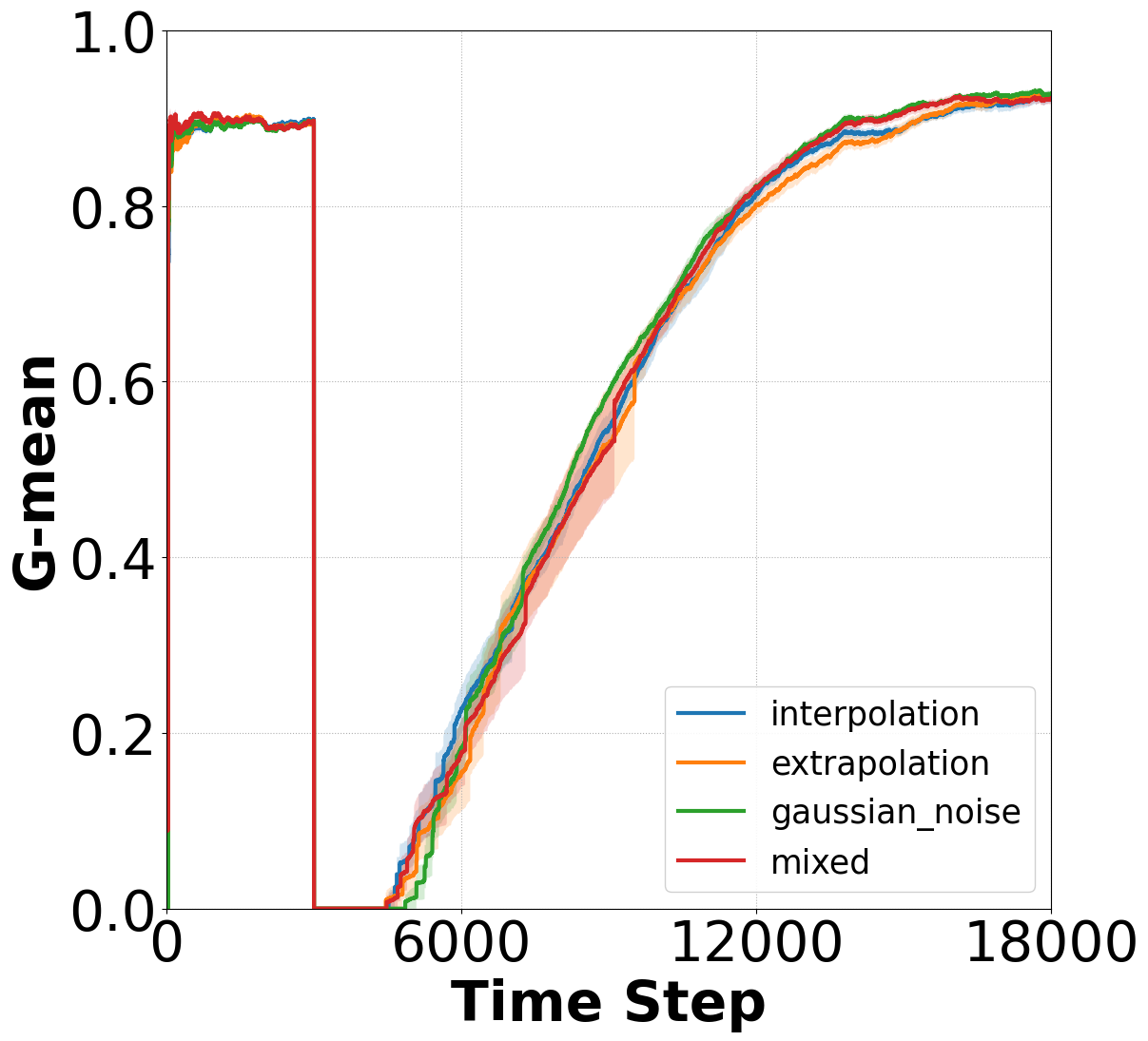}%
		\label{fig:augm_sea_abrupt}}
	\subfloat[Circles]{\includegraphics[scale=0.15]{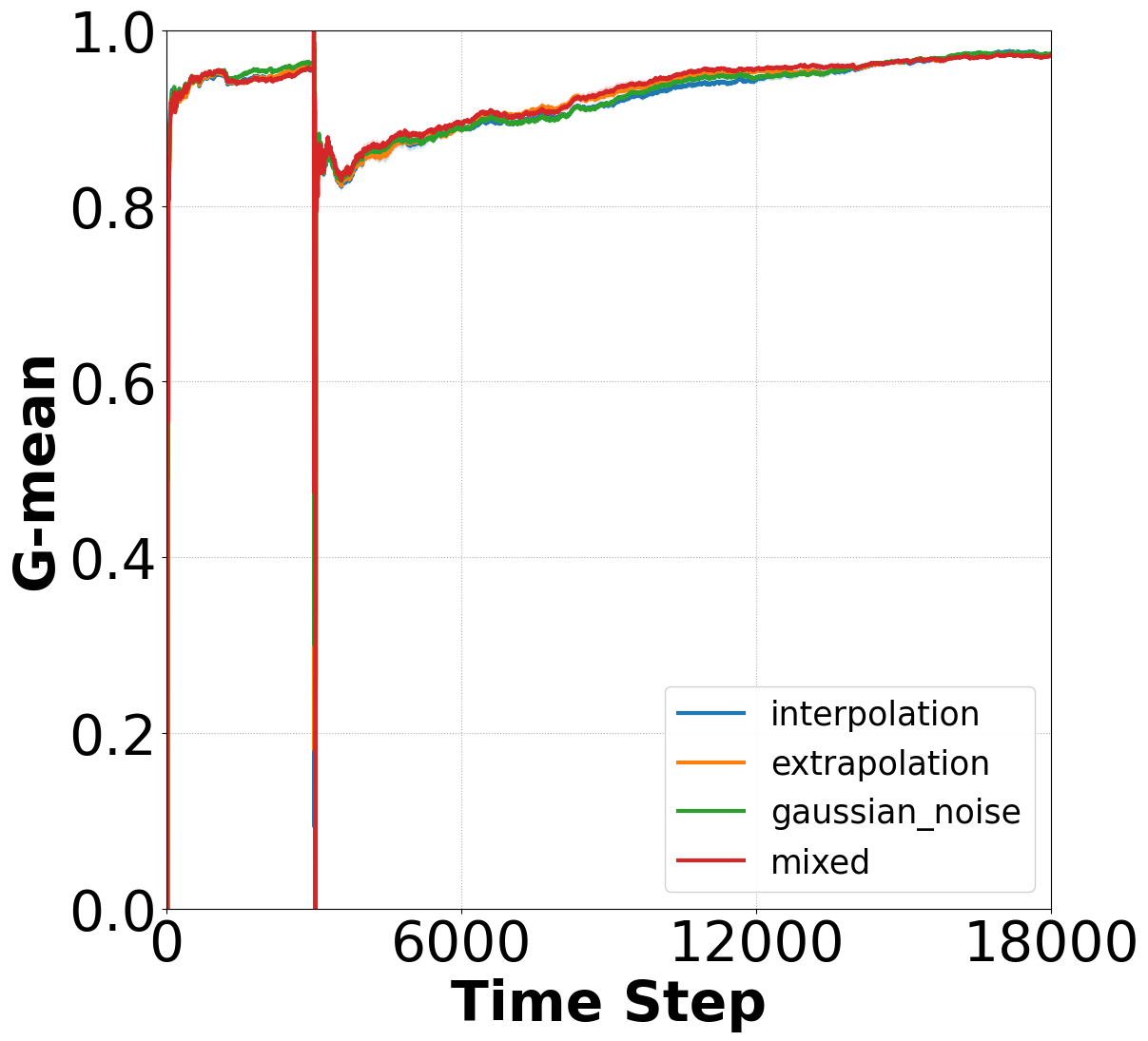}%
		\label{fig:augm_circles_abrupt}}
	\subfloat[Blobs]{\includegraphics[scale=0.15]{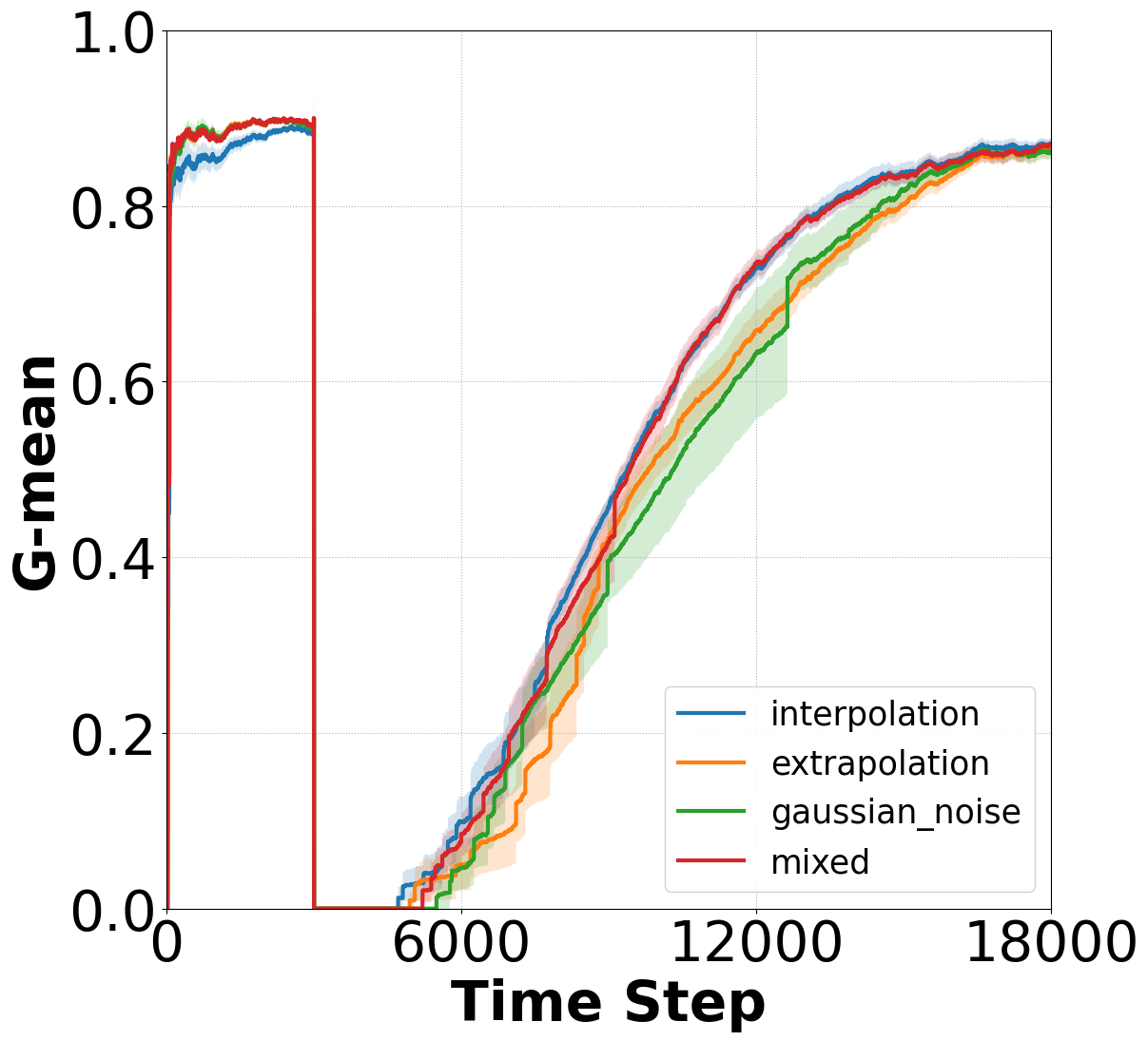}%
		\label{fig:augm_blobs_abrupt}}
	
	\caption{SiameseDuo++'s performance with interpolation, extrapolation, Gaussian noise, and mixed augmentation types in the three synthetic datasets (Sea, Circles, Blobs) with concept drift.}
\end{figure*}

\textbf{Amount of generated examples}. We examine now the number of generated examples per original example. For instance, if 10 examples are generated for each original example in $Q^t$, the generated memory grows by a factor of 10, i.e., $|Q^t_{gen}| = 10 |Q^t|$, and the size of the augmented memory (Eq. (\ref{eq:augm_memory})) is $|Q^t_{augm}| = |Q^t_{gen}| + |Q^t| = 11 |Q^t|$. In these experiments we vary the number of generated examples (1, 3, 5, and 10) per original example. The labelling budget is set to 1\% and the interpolation function is used. Figs.~\ref{fig:num_sea_abrupt}, \ref{fig:num_circles_abrupt} and \ref{fig:num_blobs_abrupt} display the performance of SiameseDuo++ in the Sea, Circles and Blobs datasets with abrupt drift respectively. The first observation is that as we increase the amount of augmentation the performance is improved. In all Figures, when only a single example is generated per original example, the performance is the lowest. The second observation is that as the number of generated  examples increases, either diminishing returns are observed (Fig.~\ref{fig:num_circles_abrupt}) or in some cases a smaller amount of augmentation is preferred (Figs.~\ref{fig:num_sea_abrupt} and \ref{fig:num_blobs_abrupt}). We attribute this to the noisy generated examples; we re-visit this in Section~\ref{sec:conclusion}.

\textbf{Transformation functions}. In these experiments we examine the role of three transformation functions (interpolation, extrapolation, Gaussian noise) for generating new examples. The active learning budget is set to 1\%. The number of generated examples per original example is set to nine, that is, the generated memory is nine times larger than the original ($|Q^t_{gen}| = 9 |Q^t|$) and the augmented memory is 10 times larger ($|Q^t_{augm}| = 10 |Q^t|$). We also experiment with a mixed transformation which generates three examples per original example for each of the three transformation functions, therefore, the generated memory remains the same for a fair comparison. Figs.~\ref{fig:augm_sea_abrupt}, \ref{fig:augm_circles_abrupt} and \ref{fig:augm_blobs_abrupt} display the performance of SiameseDuo++ in the Sea, Circles and Blobs datasets with abrupt drift respectively. The choice of the transformation function for Sea (Fig.~\ref{fig:augm_sea_abrupt}) and Circles (Fig.~\ref{fig:augm_circles_abrupt}) doesn't appear to be significant. The most interesting observation comes from Fig.~\ref{fig:augm_blobs_abrupt}. While interpolation seems to be a better choice than extrapolation and Gaussian noise, the mixed strategy appears to ``inherit'' the advantages of interpolation; the two perform the same. From now in our comparative study, we will be using the mixed strategy unless otherwise stated.

\begin{figure*}[t!]
	\centering
	
	\subfloat[Original]{\includegraphics[scale=0.15]{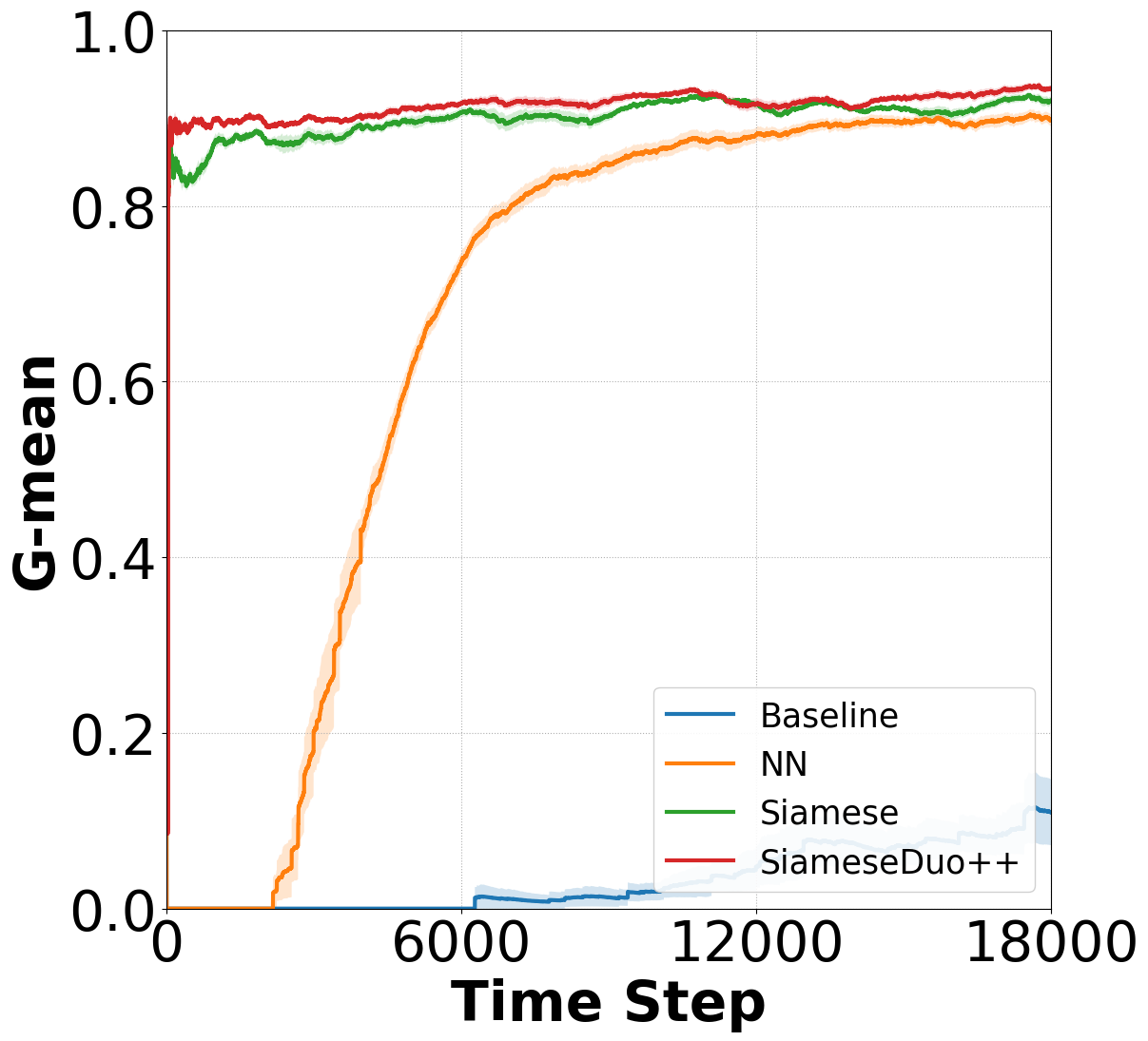}%
		\label{fig:perf_sea_original}}
	\subfloat[Abrupt drift]{\includegraphics[scale=0.15]{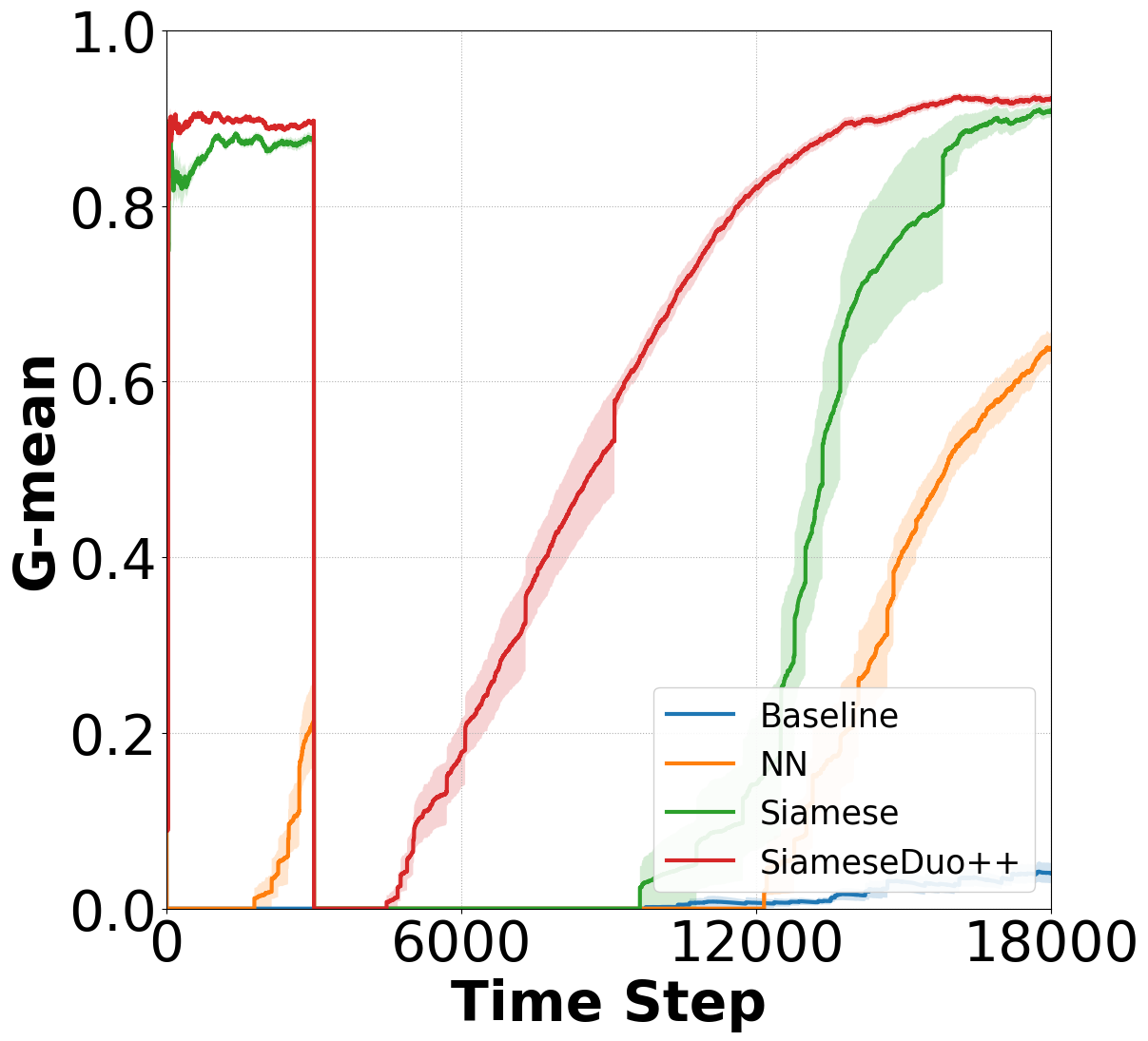}%
		\label{fig:perf_sea_abrupt}}
	\subfloat[Imbalance]{\includegraphics[scale=0.15]{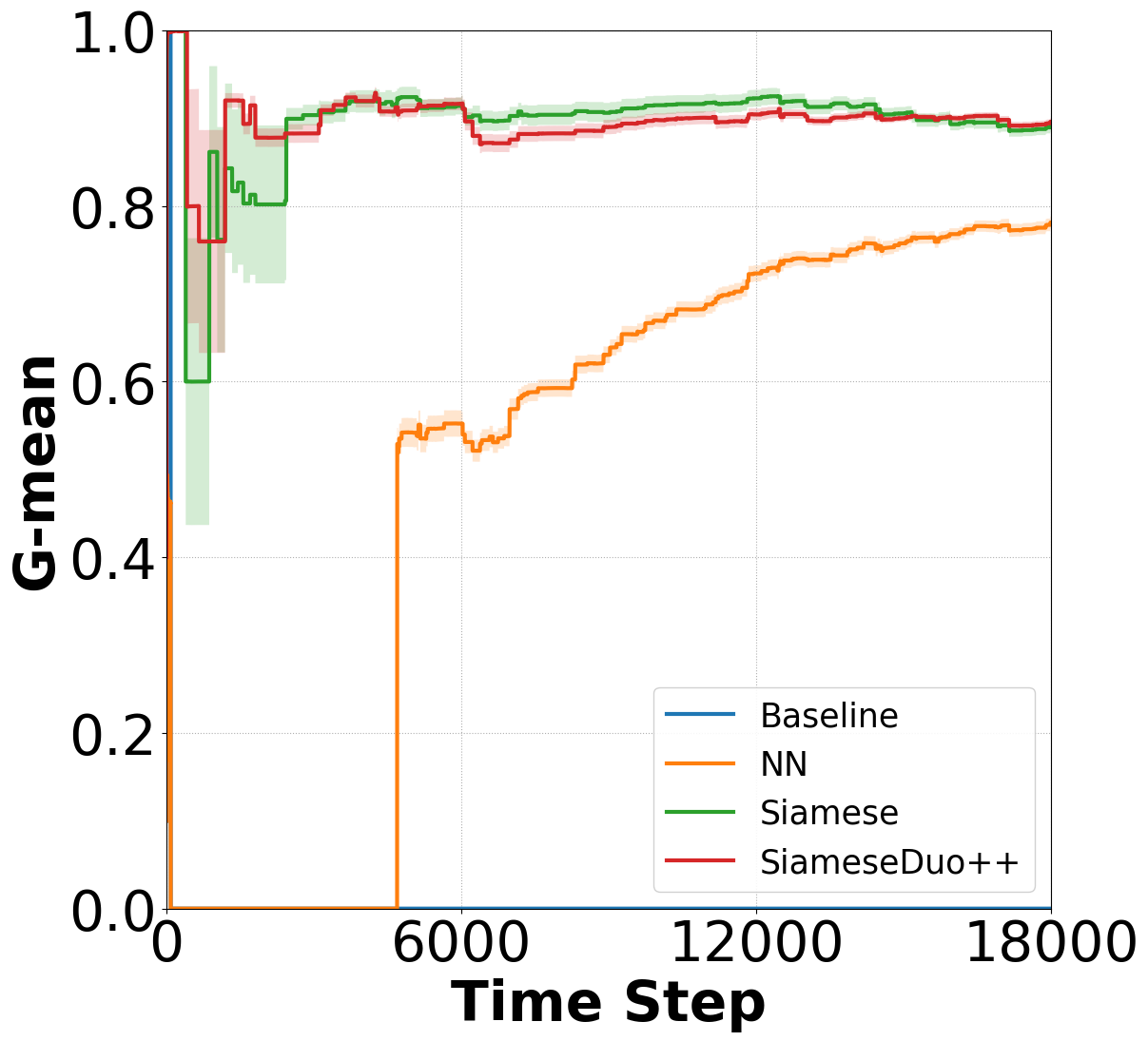}%
		\label{fig:perf_sea_mm_extreme}}
		
	\subfloat[Abrupt + Imbalance]{\includegraphics[scale=0.15]{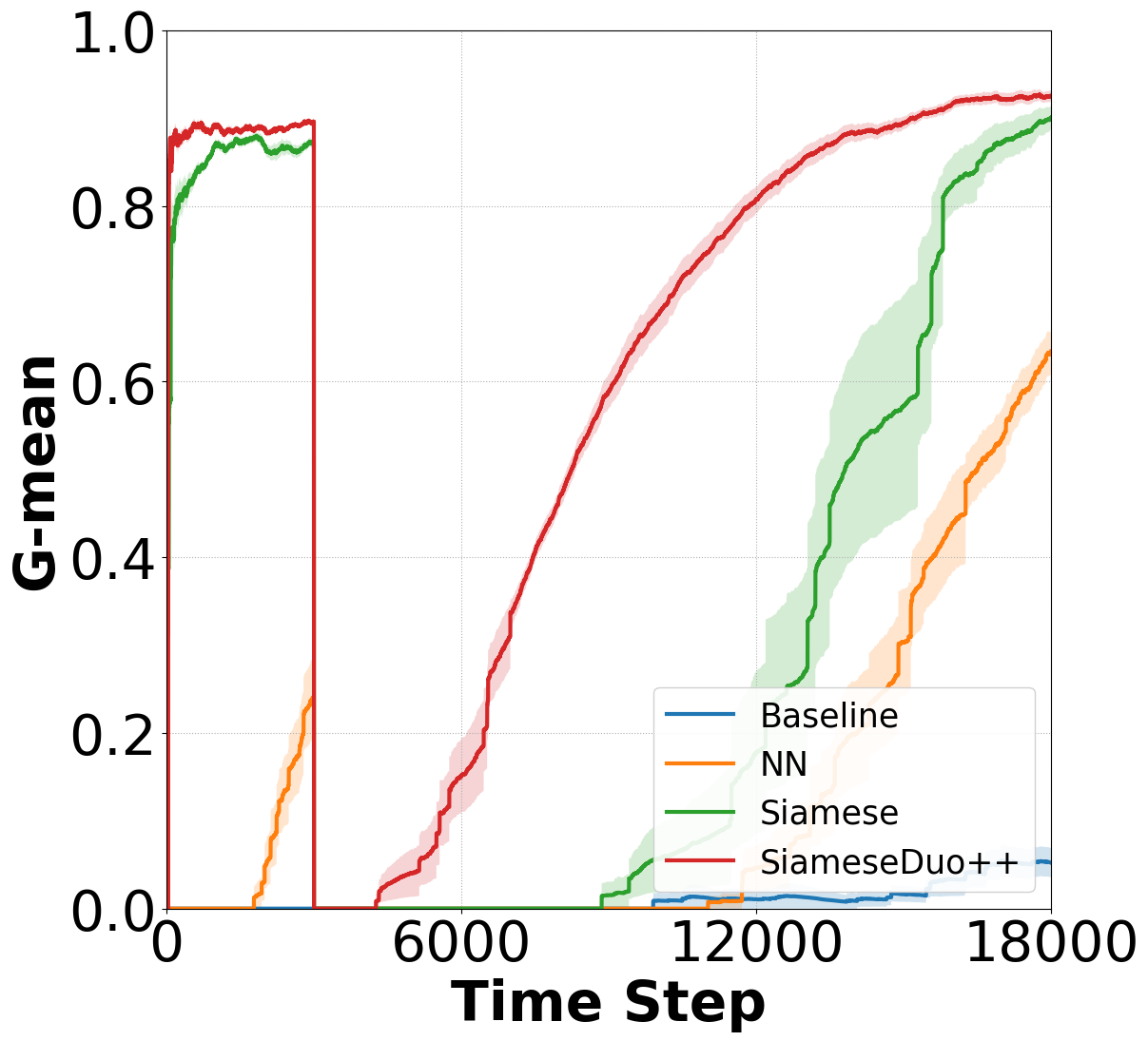}%
		\label{fig:perf_sea_abrupt_mm_severe}}
	\subfloat[Recurrent drift]{\includegraphics[scale=0.15]{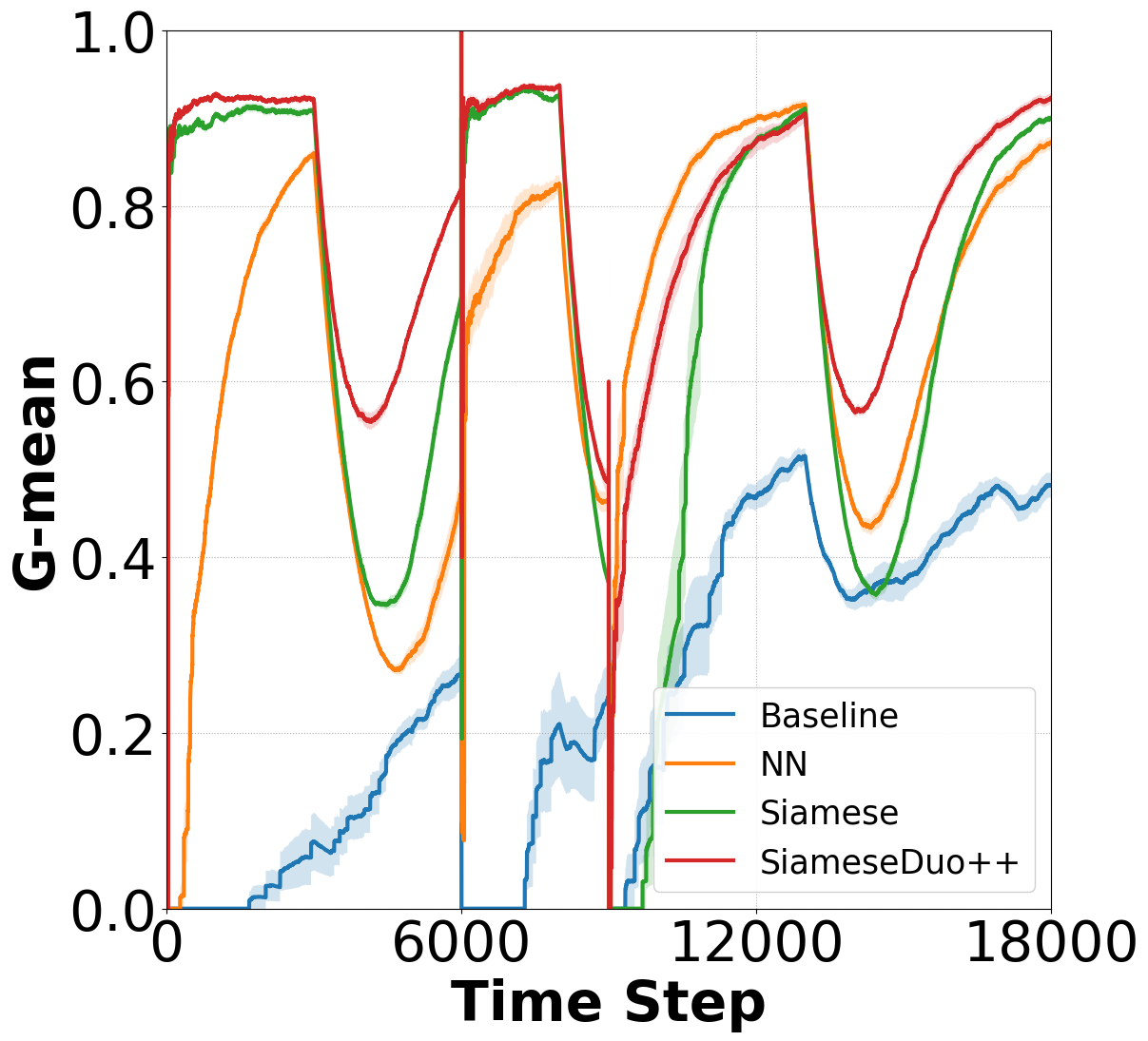}%
		\label{fig:perf_sea_recurrent}}
	
	\caption{SiameseDuo++'s performance in the five variations of the Sea dataset.}
	\label{fig:perf_sea}
\end{figure*}

\begin{figure*}[t!]
	\centering
	
	\subfloat[Original]{\includegraphics[scale=0.15]{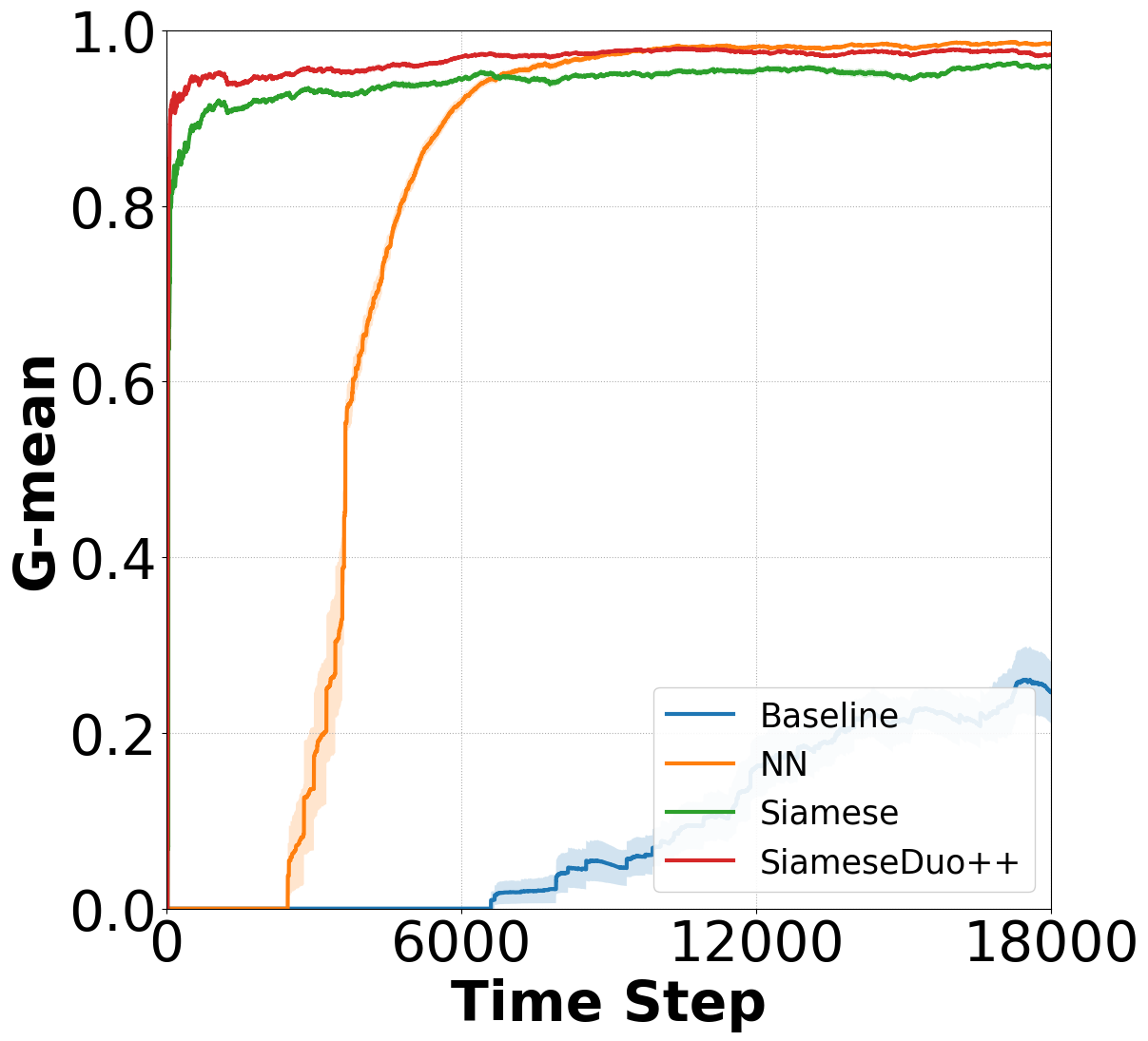}%
		\label{fig:perf_circles_original}}
	\subfloat[Abrupt drift]{\includegraphics[scale=0.15]{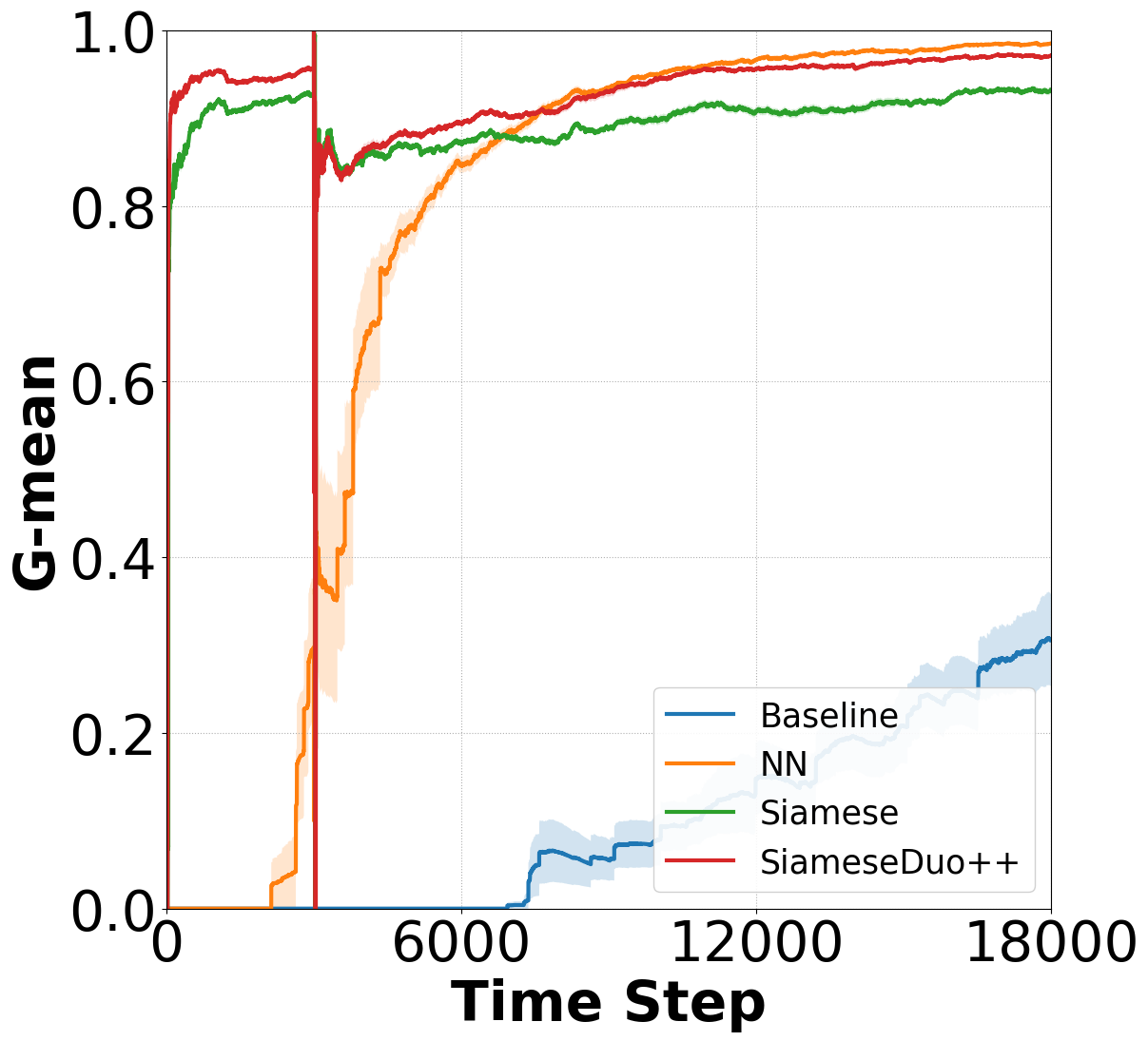}%
		\label{fig:perf_circles_abrupt}}
	\subfloat[Imbalance]{\includegraphics[scale=0.15]{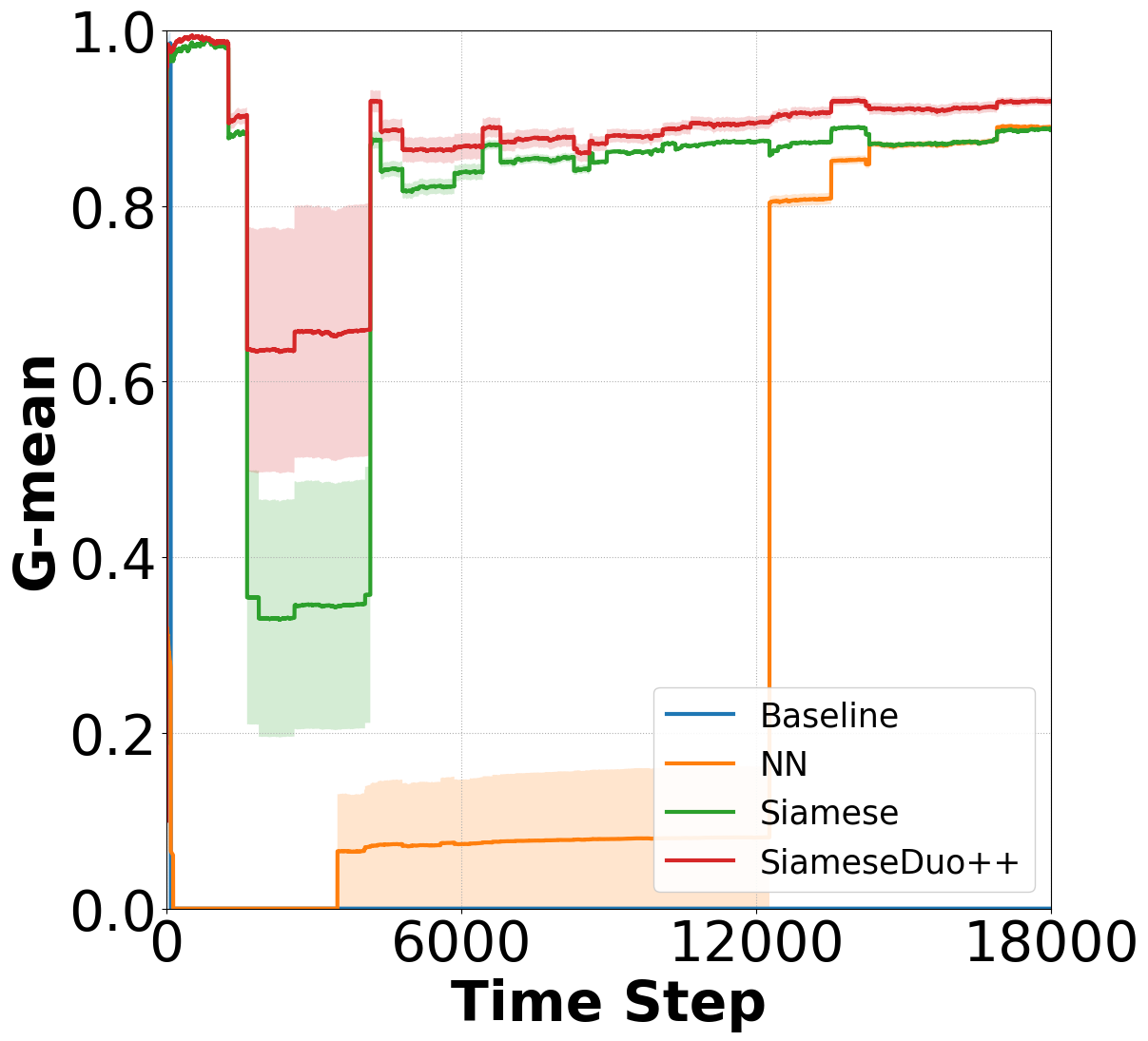}%
		\label{fig:perf_circles_mm_extreme}}
		
	\subfloat[Abrupt + Imbalance]{\includegraphics[scale=0.15]{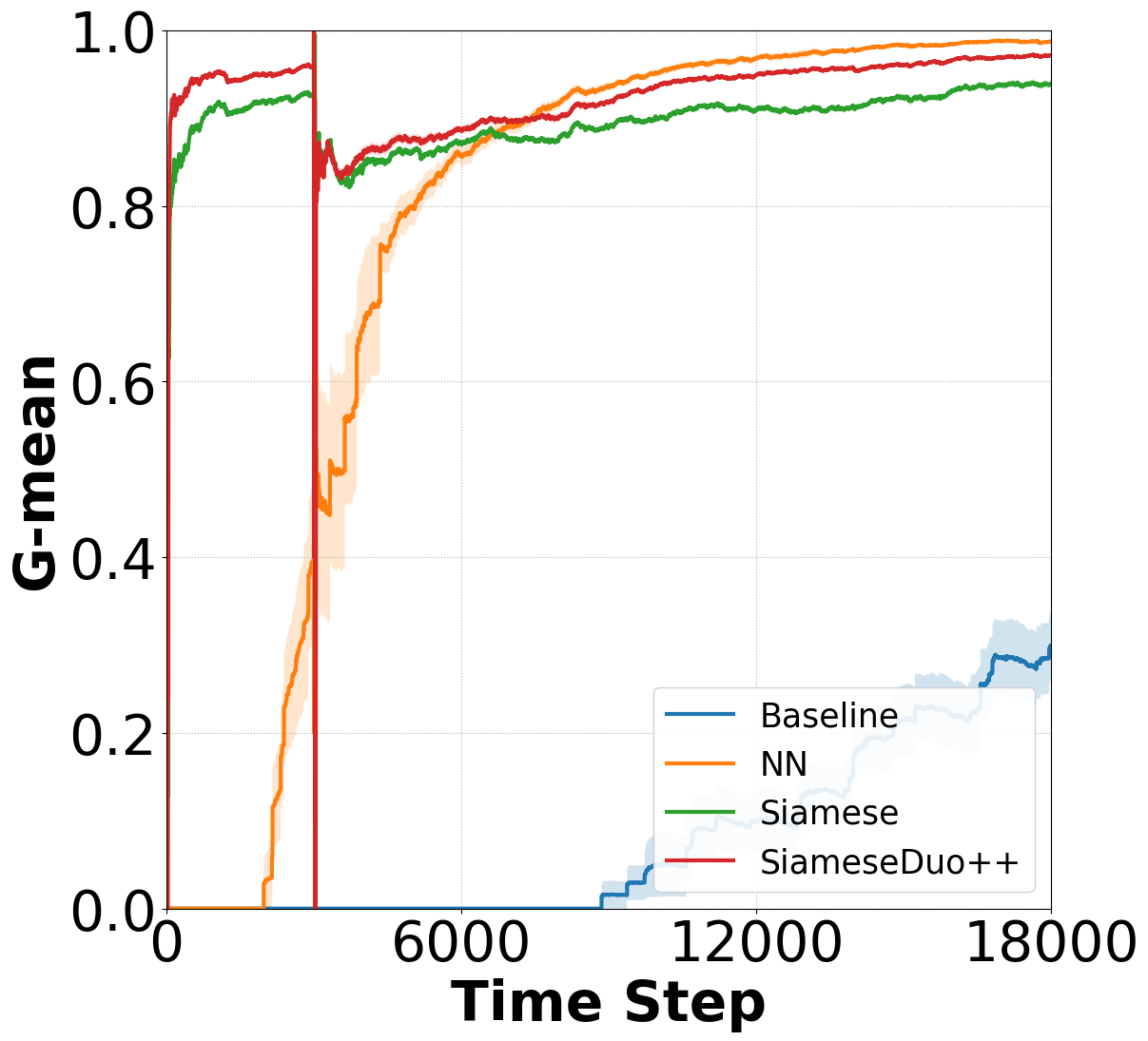}%
		\label{fig:perf_circles_abrupt_mm_severe}}
	\subfloat[Recurrent drift]{\includegraphics[scale=0.15]{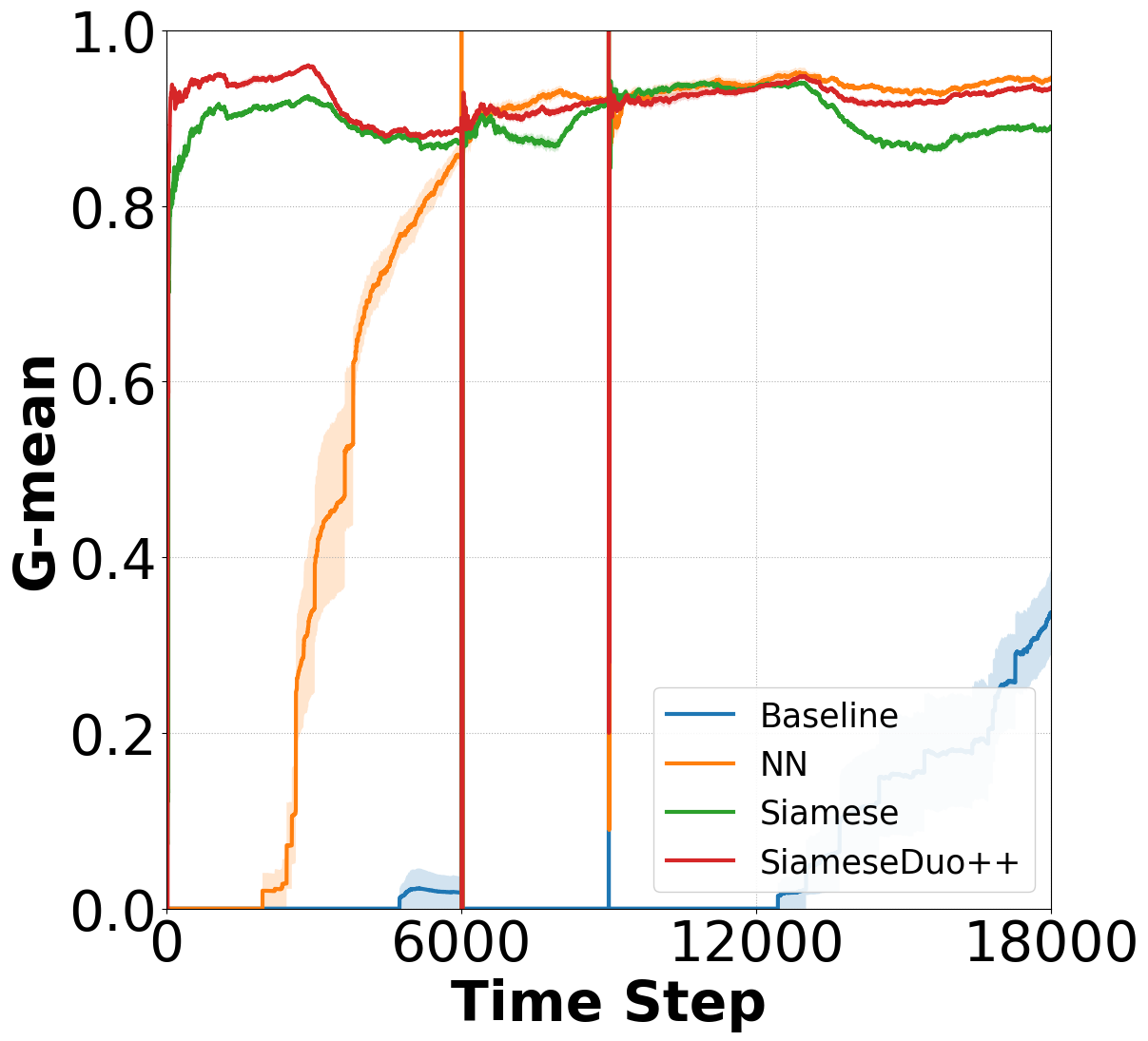}%
		\label{fig:perf_circles_recurrent}}
	
	\caption{SiameseDuo++'s performance in the five variations of the Circle dataset.}
	\label{fig:perf_circles}
\end{figure*}

\begin{figure*}[t!]
	\centering
	
	\subfloat[Original]{\includegraphics[scale=0.15]{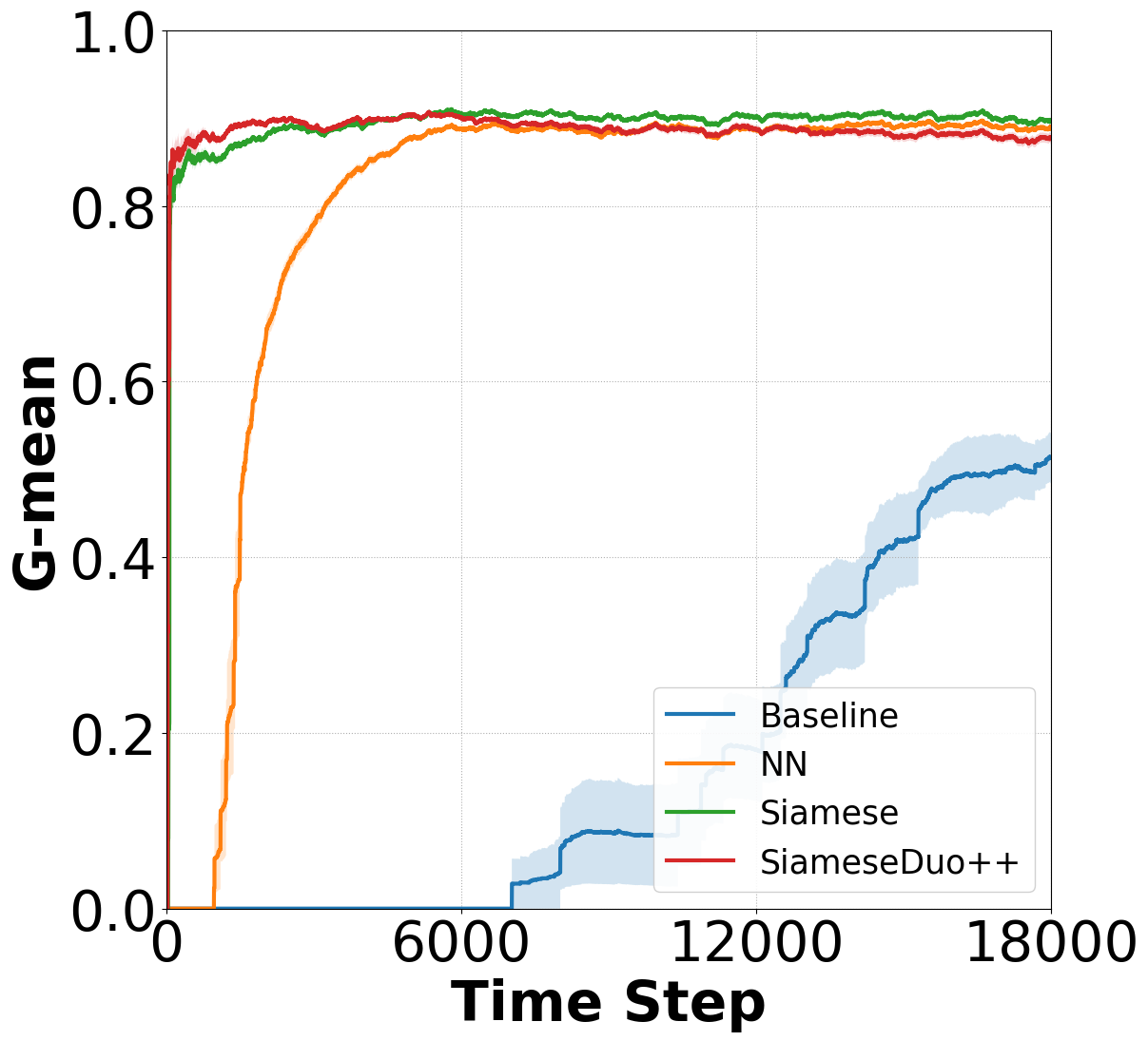}%
		\label{fig:perf_blobs_original}}
	\subfloat[Abrupt drift]{\includegraphics[scale=0.15]{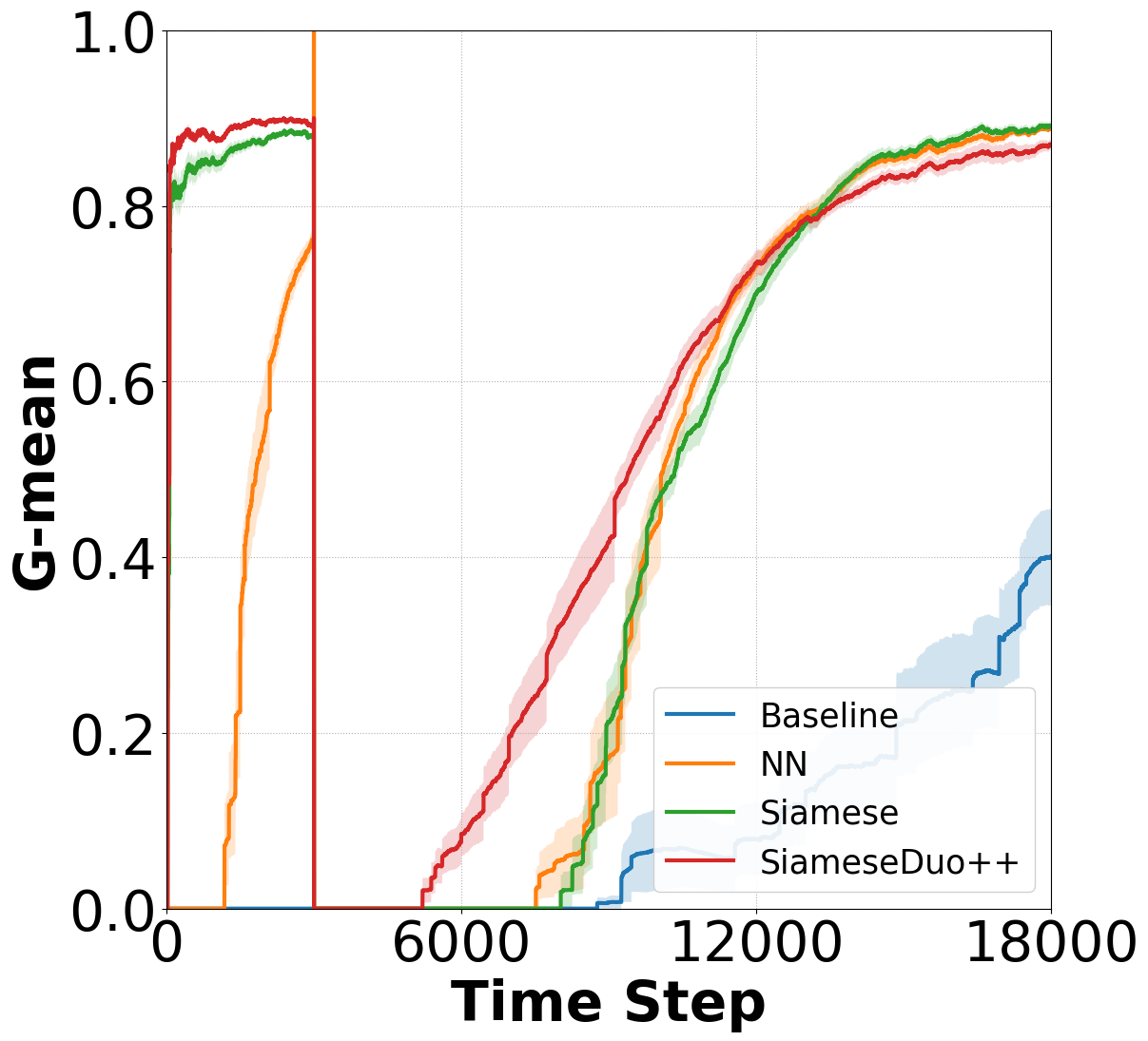}%
		\label{fig:perf_blobs_abrupt}}
	\subfloat[Imbalance]{\includegraphics[scale=0.15]{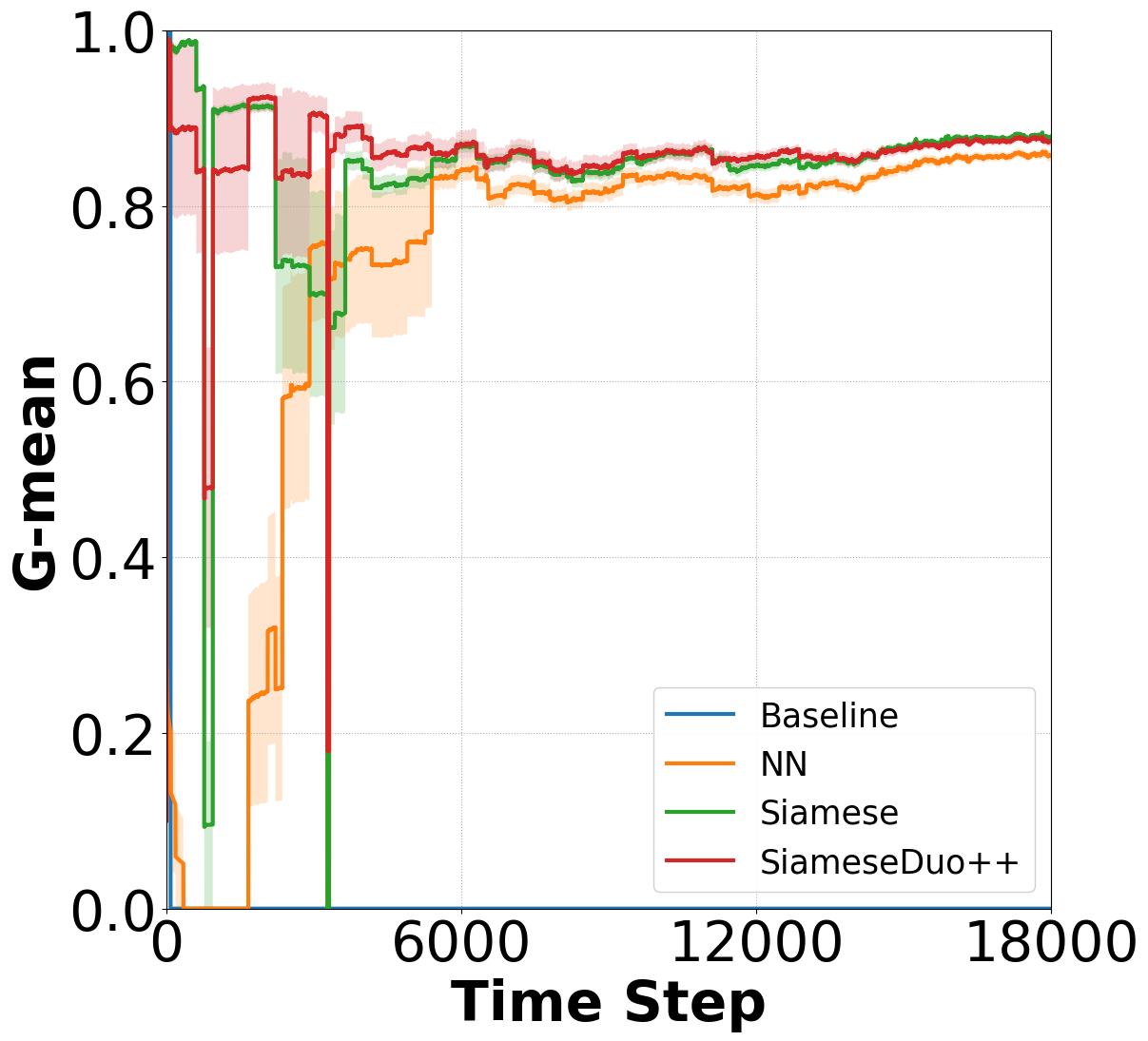}%
		\label{fig:perf_blobs_mm_extreme}}
		
	\subfloat[Abrupt + Imbalance]{\includegraphics[scale=0.15]{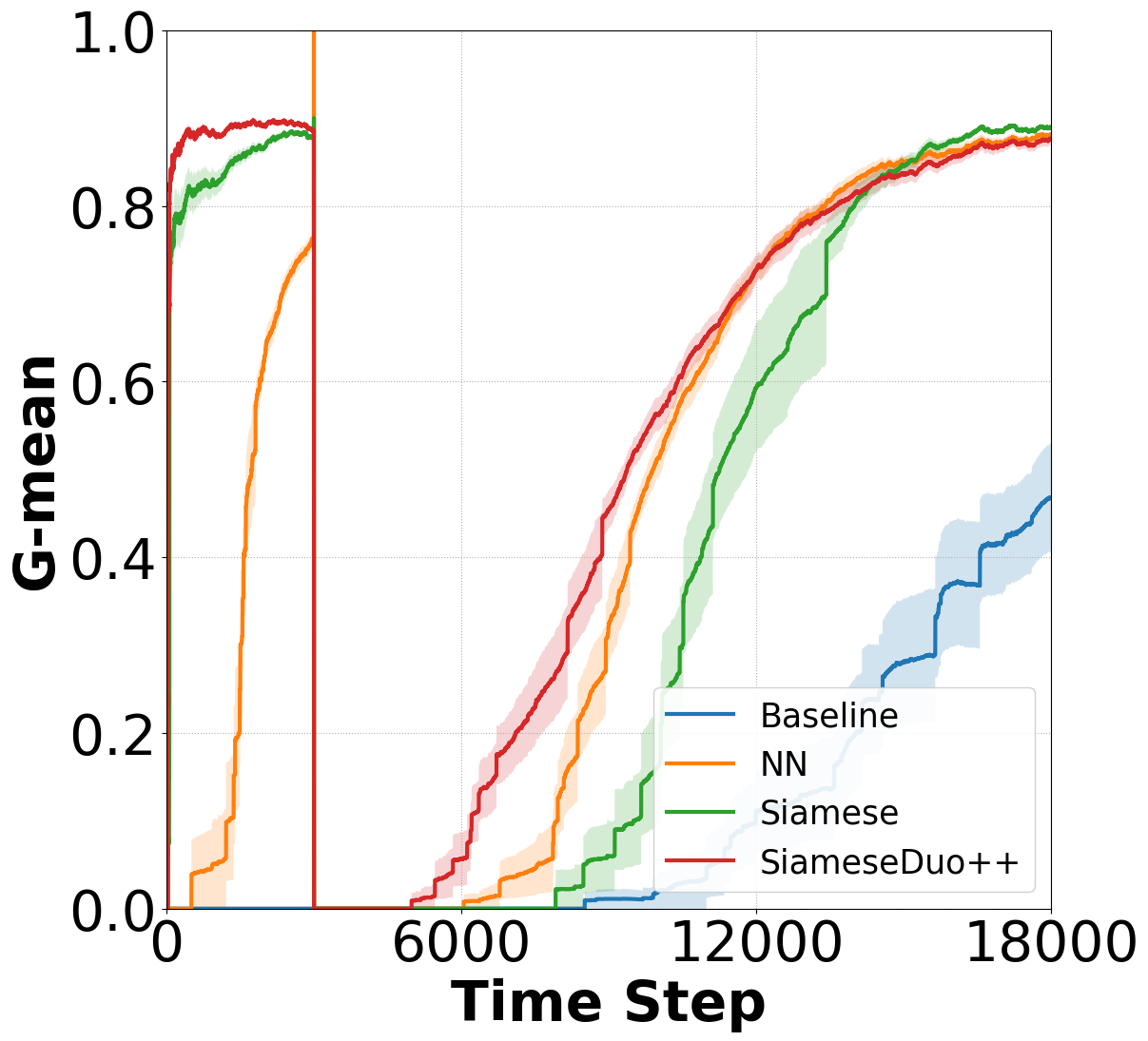}%
		\label{fig:perf_blobs_abrupt_mm_severe}}
	\subfloat[Recurrent drift]{\includegraphics[scale=0.15]{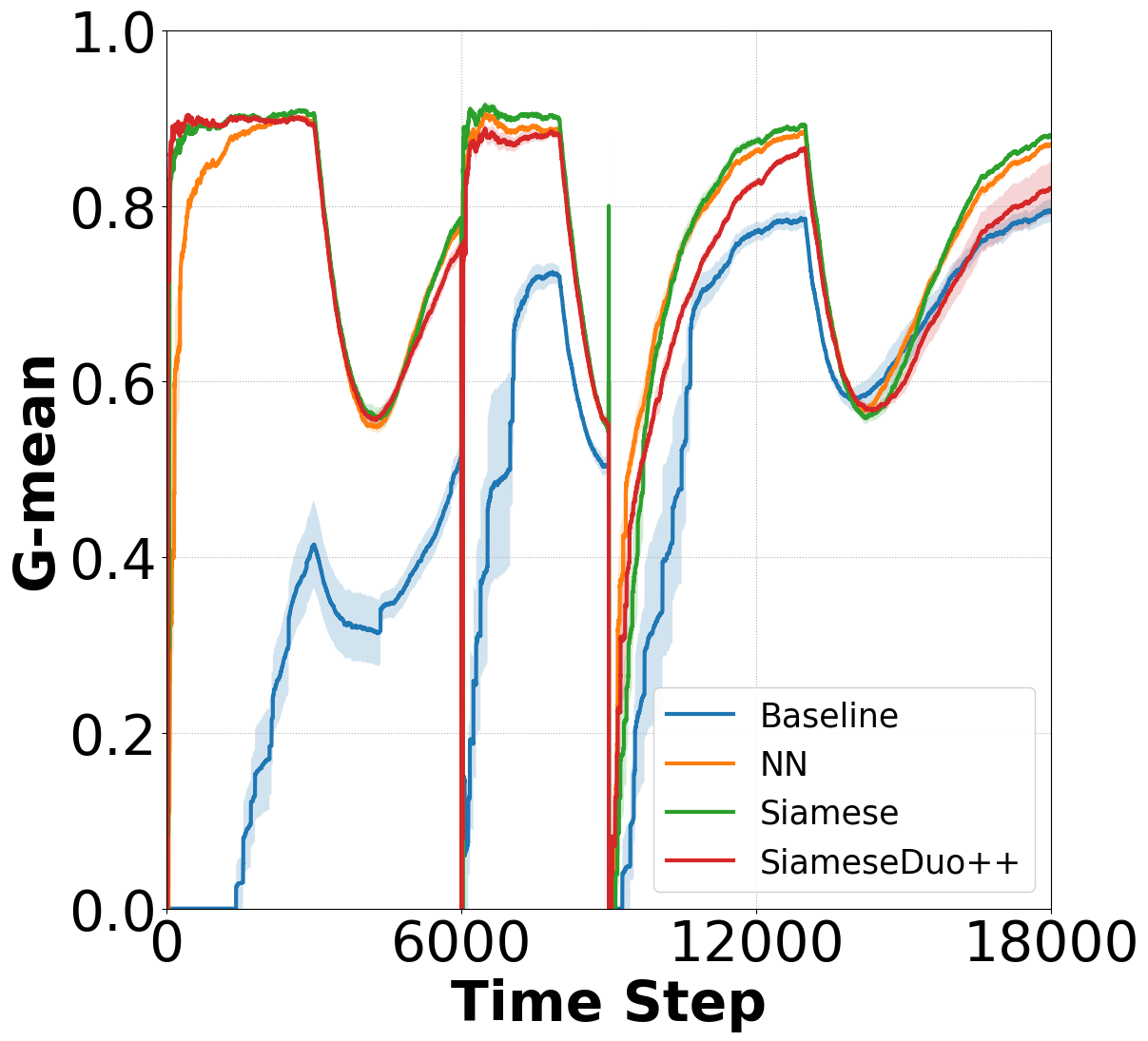}%
		\label{fig:perf_blobs_recurrent}}
	
	\caption{SiameseDuo++'s performance in the five variations of the Blobs dataset.}
	\label{fig:perf_blobs}
\end{figure*}

\subsection{Comparative study}

\textbf{Synthetic data}. We examine the performance of all methods in the five variations of each synthetic dataset, i.e., the original version, with abrupt drift, with 0.1\% imbalance, with both abrupt drift and 1\% imbalance, and with recurrent drift. The active learning budget is 1\% in all cases.

Fig.~\ref{fig:perf_sea} shows the performance for Sea. The performance of SiameseDuo++ is superior to the rest particularly in the cases with abrupt drift (Fig.~\ref{fig:perf_sea_abrupt}), drift and imbalance (Fig.~\ref{fig:perf_sea_abrupt_mm_severe}), and recurrent drift (Fig.~\ref{fig:perf_sea_recurrent}). The Baseline is unable to cope under these challenging conditions; sometimes it achieves a 0\% performance or close to it. The NN method which uses a standard neural network also performs poorly. In the less challenging case (Fig.~\ref{fig:perf_sea_original}), NN equalises the performance of siamese-based methods but with a significantly slower learning speed. The state-of-the-art Siamese method, as expected, it performs well in some cases. Siamese and SiameseDuo++ yield a similar performance in Figs.~\ref{fig:perf_sea_original} and \ref{fig:perf_sea_mm_extreme}, however, SiameseDuo++ significantly outperforms it in the other cases.

Similar results are obtained in Fig.~\ref{fig:perf_circles} for Circles. Interestingly in all plots, NN's final performance is greater than Siamese's, although the latter learns significantly faster. SiameseDuo++ equalises the performance of NN while it inherits the learning speed of Siamese. Lastly, similar results are obtained in Fig.~\ref{fig:perf_blobs} for Blobs. Lastly, Table~\ref{tab:perfSynth} shows the mean G-mean and the standard deviation at the last time.

\begin{figure*}[h!]
	\centering
	\subfloat[Keystroke]{\includegraphics[scale=0.15]{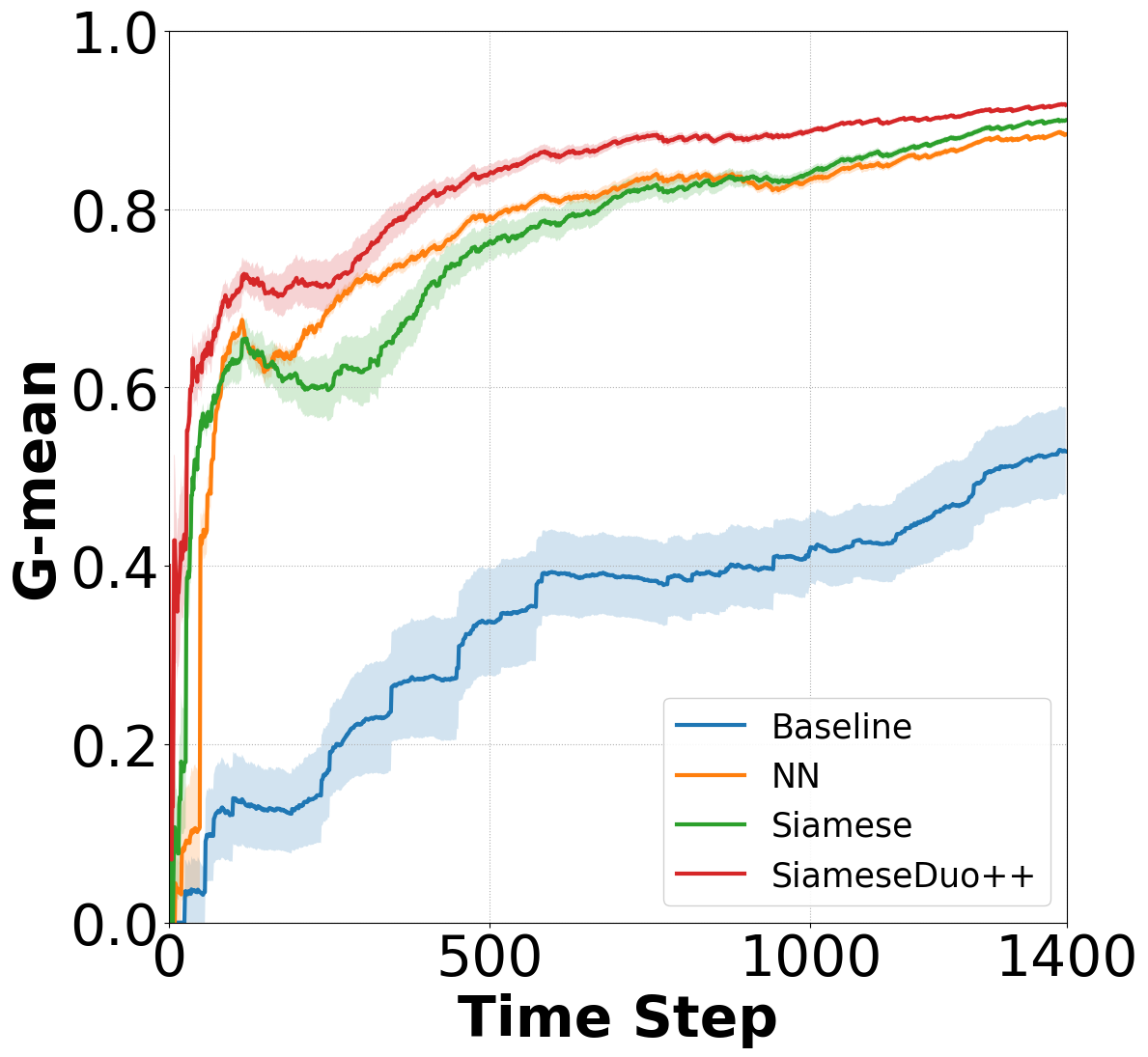}%
		\label{fig:augm_keystroke}}
	\subfloat[Hand-written digits]{\includegraphics[scale=0.15]{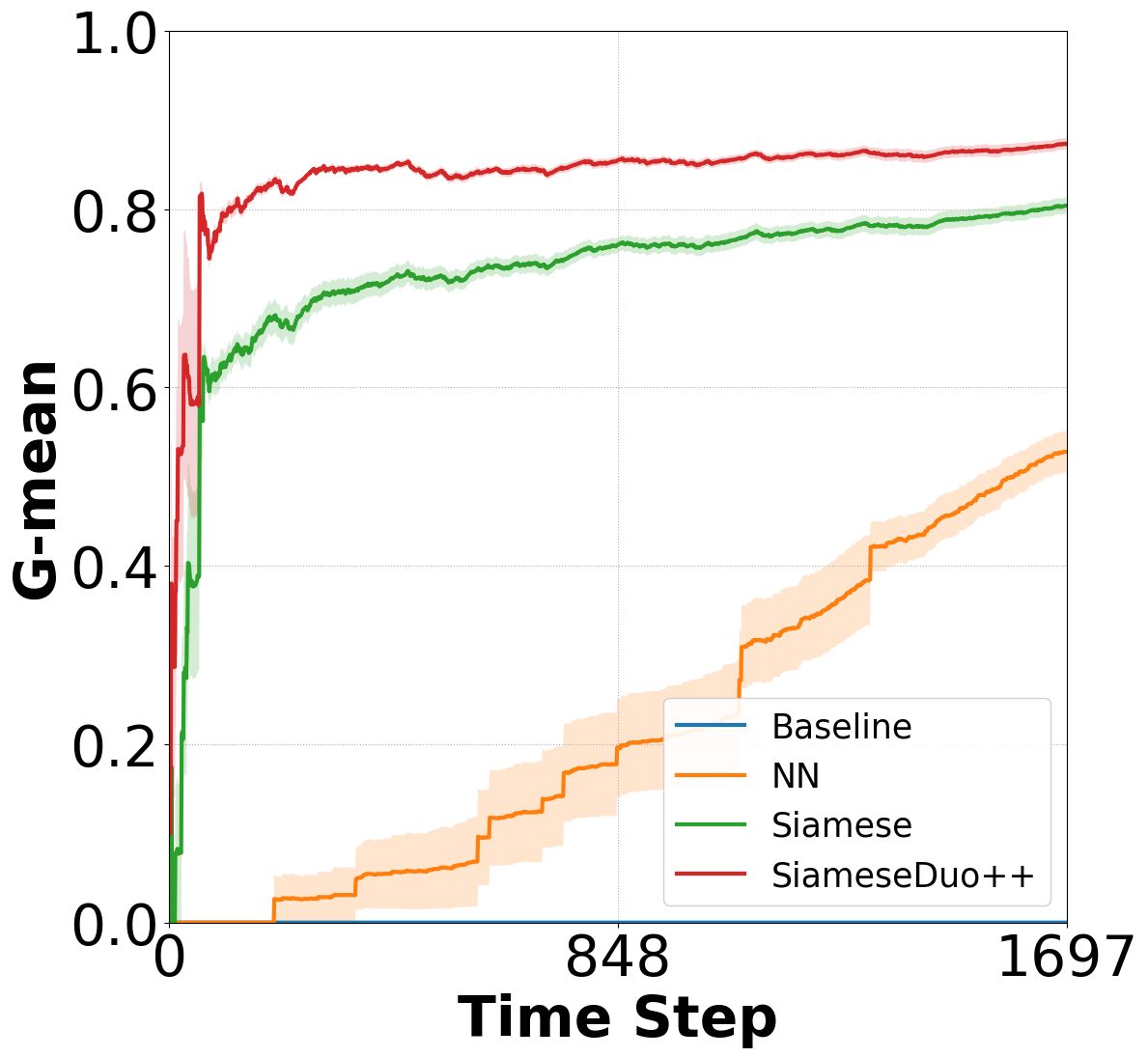}%
		\label{fig:augm_digits}}
	\subfloat[uWave gestures]{\includegraphics[scale=0.15]{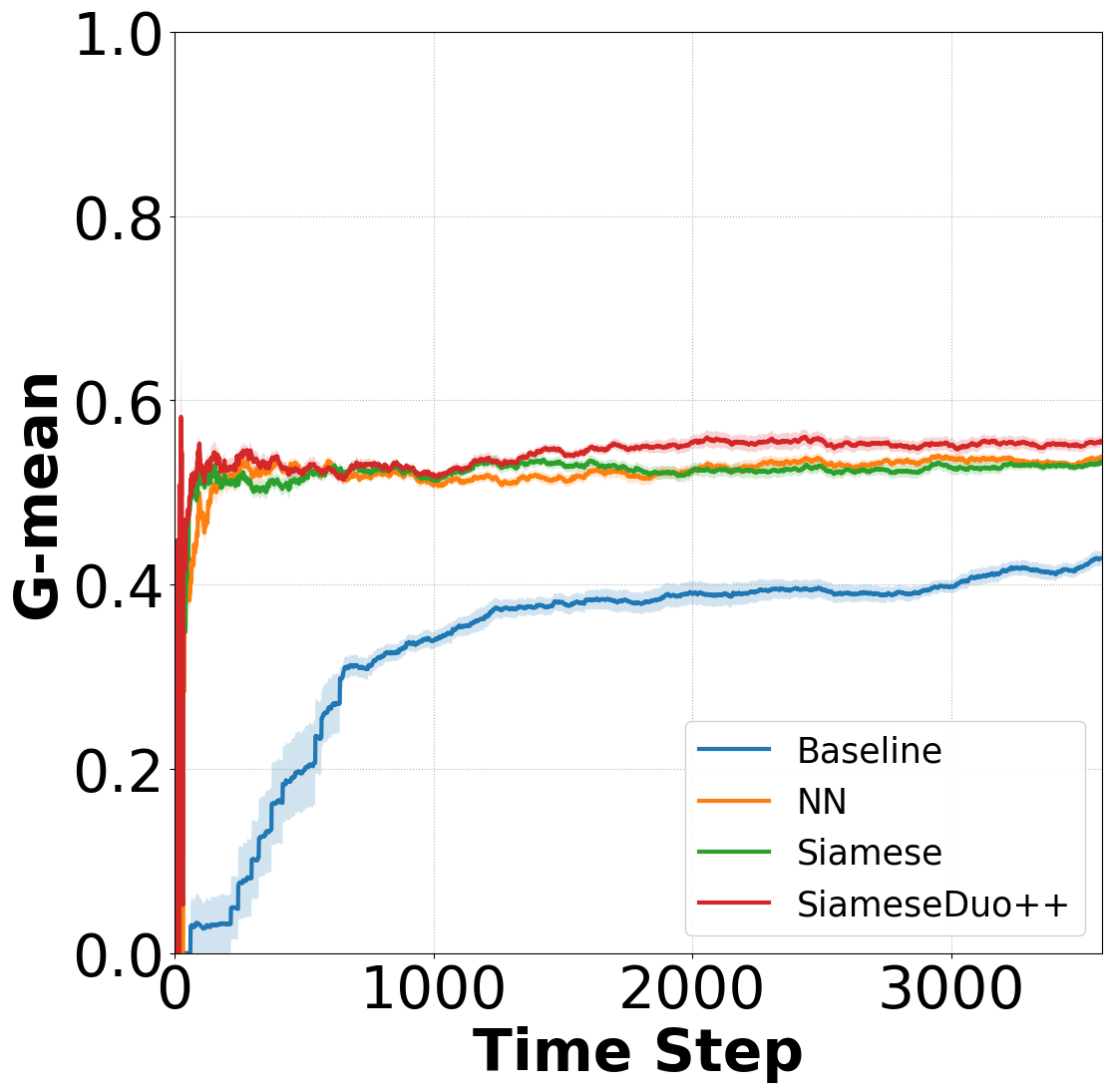}%
		\label{fig:augm_uwave}}
		
	\subfloat[Starlight curves]{\includegraphics[scale=0.15]{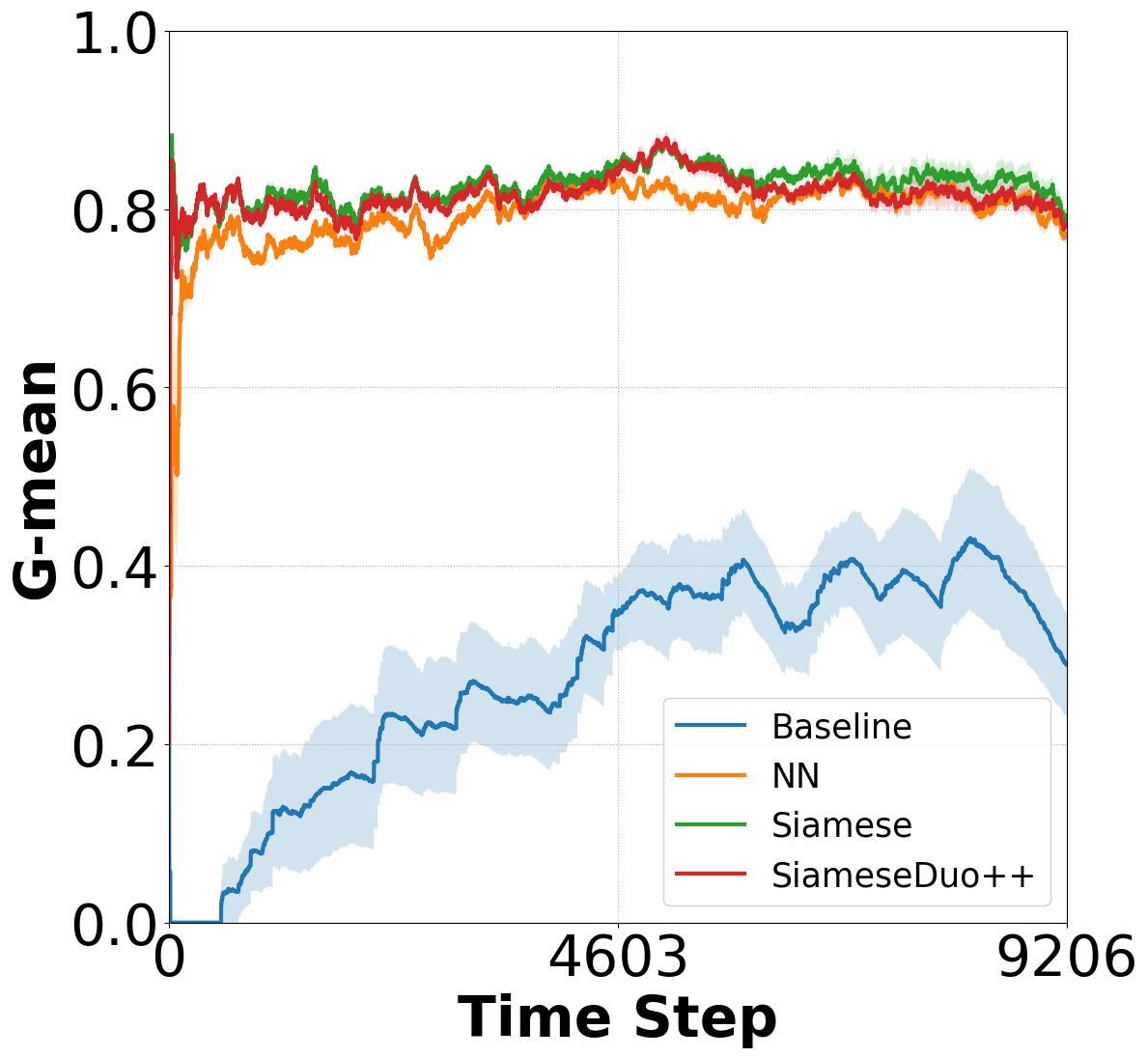}%
		\label{fig:augm_starlightcurves}}
	\subfloat[Electromyography]{\includegraphics[scale=0.15]{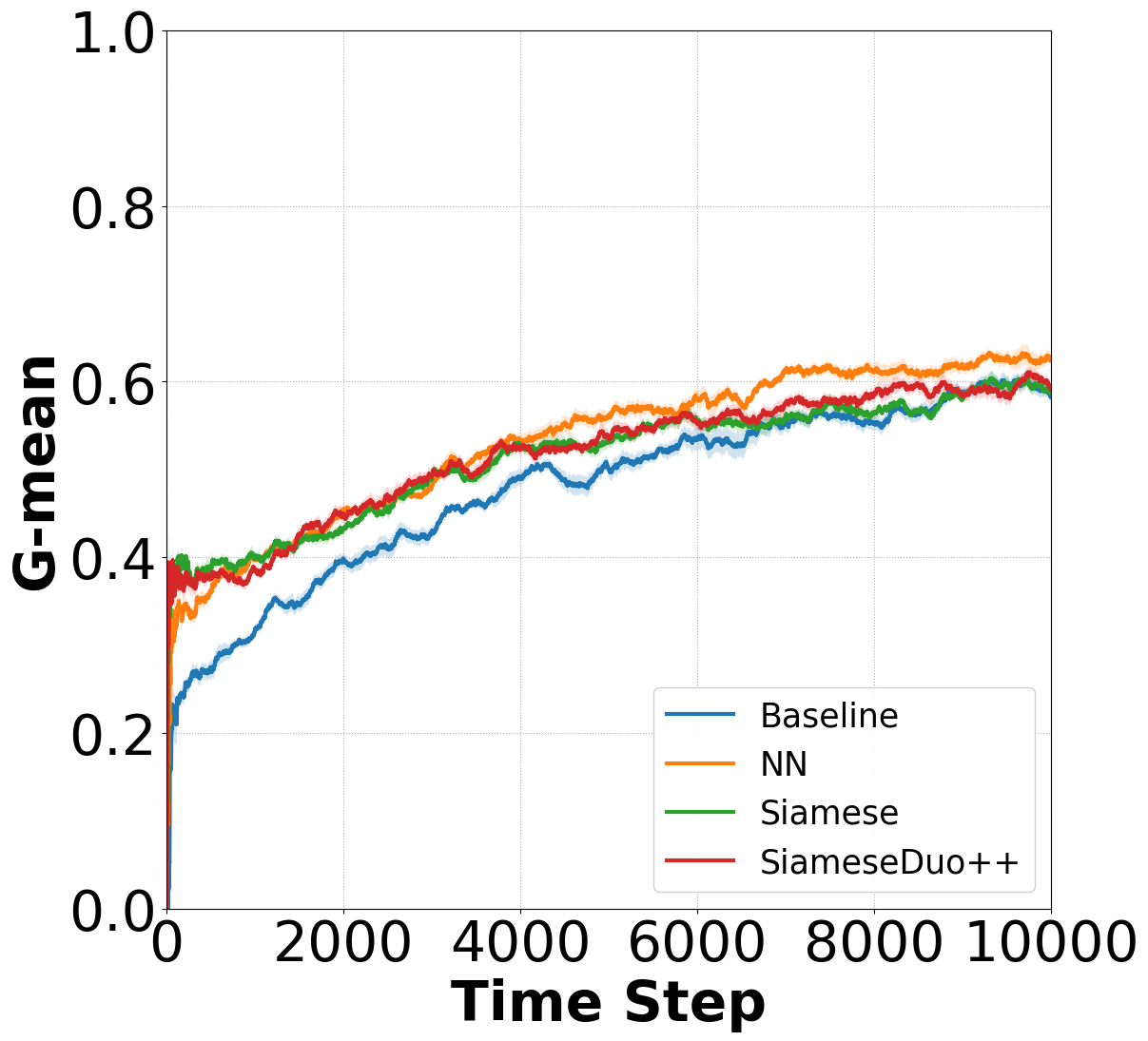}%
		\label{fig:augm_gestures}}
	
	\caption{SiameseDuo++'s performance in the real-world datasets.}
	\label{fig:real}
\end{figure*}

\textbf{Real data}. We discuss now the performance of all methods in the real-world datasets. The proposed method SiameseDuo has been shown to be effective in the Synthetic datasets using a tiny budget of 1\%. Given the more challenging conditions encountered in real-world dataset, a larger budget appears to be necessary for some datasets. Importantly though, the budget still remains low and never exceeds 10\%. Specifically, it is set to 1\% for Hand-written digits and Starlight curves, 5\% for uWave gestures, and 10\% for Keystroke and Electromyography. In all cases, SiameseDuo++ uses all three (interpolation, extrapolation, Gaussian noise) transformation functions. Fig.~\ref{fig:augm_keystroke} shows the results for Keystroke, where the proposed SiameseDuo++ significantly outperforms the rest. Notably, NN outperforms Siamese, while Baseline performs poorly. Fig.~\ref{fig:augm_digits} shows the results for Hand-written digits, where SiameseDuo++ significantly outperforms the rest. NN performs poorly, while Baseline yields a 0\% performance. Fig.~\ref{fig:augm_uwave} shows the results for uWave gestures, where again the proposed method outperforms the rest although to a less degree. In Fig.\ref{fig:augm_starlightcurves}, SiameseDuo++ and Siamese achieve the same performance. In Fig.\ref{fig:augm_gestures}, the NN method slightly outperforms the rest with SiameseDuo++ being second best.

Tables~\ref{tab:perfReal} and \ref{tab:perfReal_auc} show the mean G-mean and AUC respectively, and the standard deviation at the last time. The results in the two tables are aligned, except in the uWave gestures dataset. The difference is attributed to the fact that G-mean can better reflect the performance on minority classes, while the AUC can better reflect the performance on majority classes \cite{wang2020auc}. In the case of uWave gestures, the NN method predicts samples from the majority class better, which is reflected in Table~\ref{tab:perfReal_auc}.

\begin{figure}[t!]
	\centering
	
	\subfloat[Keystroke]{\includegraphics[scale=0.25]{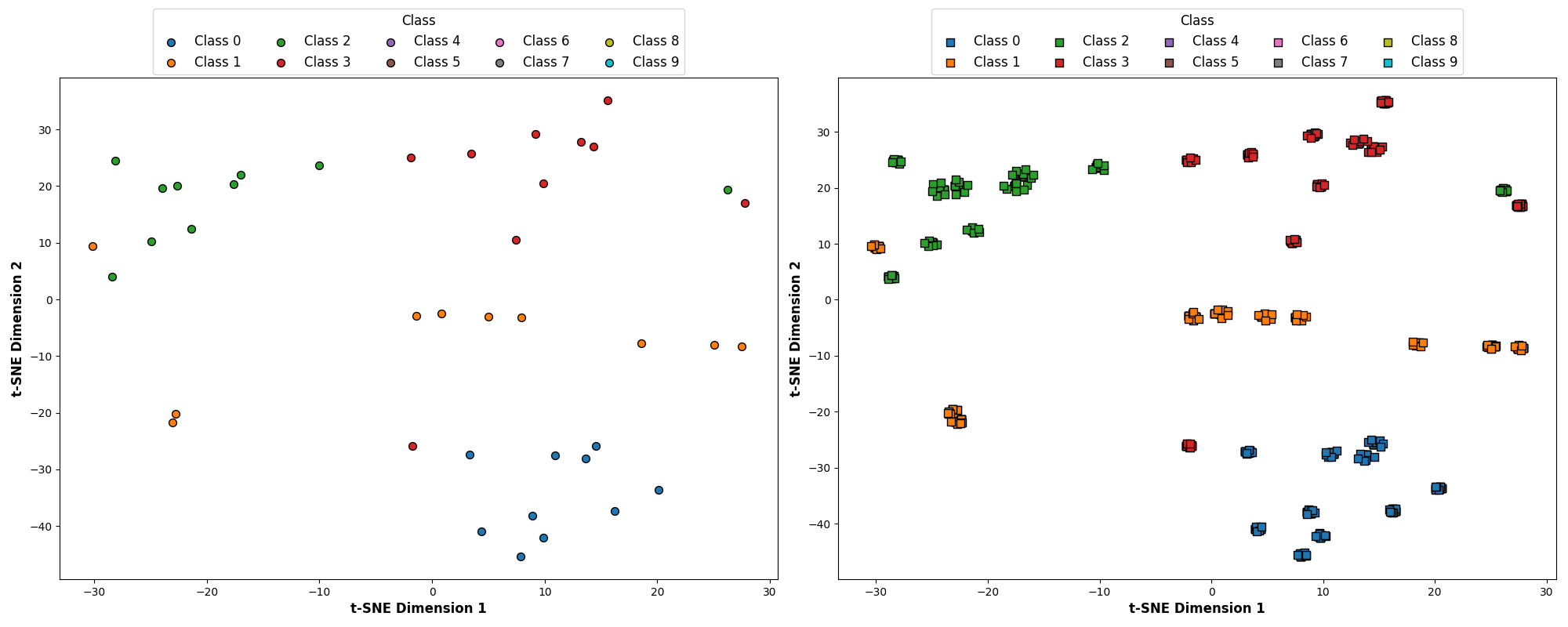}%
		\label{fig:space_keystroke}}
		
	\subfloat[Digits]{\includegraphics[scale=0.25]{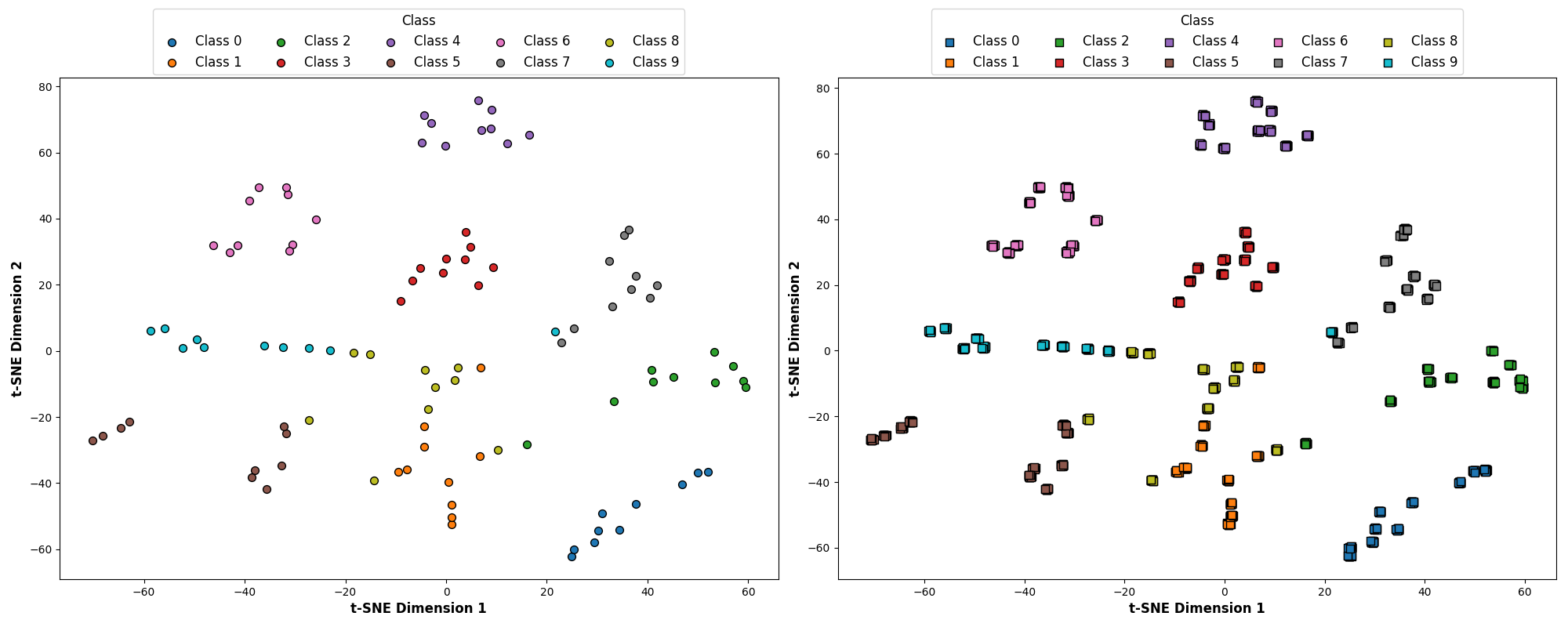}%
		\label{fig:space_digits}}
	
	\caption{Embedding space of some original points (left) and their augmentations (right).}
\end{figure}

Lastly, Fig.~\ref{fig:space_keystroke} depicts at time step $t=1000$ for Keystroke dataset, the embedding space of the original data points in $Q^t_{enc}$ (left) and their augmentations $Q^t_{gen}$ (right). The analogous results for the Digits datasets are shown in Fig.~\ref{fig:space_digits}.

\subsection{Discussion}
We have shown that in most scenarios, particularly those with challenging conditions, such as, with class imbalance or with the co-occurrence of imbalance and concept drift, the proposed SiameseDuo++ significantly outperforms the rest with respect to learning speed and/or performance. This is attributed to the pair creation process which creates a considerably larger number of training examples (the pairs), resulting into a form of oversampling, similarly to the Siamese method \cite{malialis2022nonstationary}. In addition, SiameseDuo++ performs explicit data augmentation to further increase the amount of training data, which further improves the learning speed and/or performance. It has also been shown that in some cases, such as, the balanced Electromyography dataset, the best performing method was a simple memory-based (NN method), which illustrates that neither a siamese network-based density sampling strategy (Siamese, SiameseDuo++) helped, nor data augmentation (SiameseDuo++). Therefore, this is attributed to two possible reasons. First, the active learning strategy. As discussed in Section 3.1.2, each has its pros and cons. For example, an uncertainty sampling strategy queries the most uncertain examples (which typically lie near decision boundaries), while a density sampling strategy queries the most representative examples (which typically lie in dense regions). This illustrates the necessity in some cases of a hybrid active learning strategy. Second, the quality of data augmentation. As discussed in Section 6.1, it has been observed that as the number of generated examples increases, either diminishing returns are observed or in some cases a smaller amount of augmentation is beneficial. This illustrates the necessity in some cases of monitoring the quality of augmented examples.

\section{Conclusions}\label{sec:conclusion}
We have proposed the SiameseDuo++ method, which uses active learning to incrementally train two siamese neural networks which operate in synergy, combined with data augmentation. The proposed active learning strategy and augmentation operate in the latent space. SiameseDuo++ operates with limited memory (e.g., 10 examples per class) and limited active learning budget (e.g., less than 10\%), and outperforms strong baselines and state-of-the-art methods in terms of learning speed and/or performance. Future directions are the following.

\begin{table}[t!]
	\centering
	\caption{Mean G-mean and standard deviation at time step 10000.}
	\label{tab:perfSynth}
	\begin{tabular}{llll}
		\hline
		& \textbf{Sea}         & \textbf{Circles}     & \textbf{Blobs}       \\
		\hline
		\textbf{Baseline}     & 0.01 (0.03)          & 0.05 (0.11)          & 0.02 (0.06)          \\
		\textbf{NN}           & 0.00 (0.00)          & \textbf{0.95 (0.01)} & 0.51 (0.07)          \\
		\textbf{Siamese}      & 0.06 (0.11)          & 0.90 (0.01)          & 0.16 (0.20)          \\
		\textbf{SiameseDuo++} & \textbf{0.68 (0.07)} & \textbf{0.94 (0.01)} & \textbf{0.56 (0.06)} \\
		\hline
	\end{tabular}
\end{table}

\begin{table*}[t!]
	\centering
	\caption{Mean G-mean and standard deviation at the last time step.}
	\label{tab:perfReal}
	\resizebox{\columnwidth}{!}{%
		\begin{tabular}{llllll}
			\hline
			& \textbf{Keystroke} & \textbf{Hand-written digits} & \textbf{uWave gestures} & \textbf{Starlight curves} & \textbf{Electromyography} \\ \hline
			\textbf{Baseline}     & 52.83 (4.87)       & 00.00 (0.00)                    & 42.84 (0.81)            & 28.91 (5.79)              & 58.31 (0.72)      \\
			\textbf{NN}           & 88.34 (0.31)       & 52.79 (2.32)                 & 53.74 (0.37)            & 76.94 (0.77)              & \textbf{62.80 (0.67)}      \\
			\textbf{Siamese}      & 89.94 (0.27)       & 80.37 (0.90)                 & 53.19 (0.39)            & \textbf{78.94 (0.97)}              & 58.69 (0.63)      \\
			\textbf{SiameseDuo++} & \textbf{91.68 (0.28)}       & \textbf{87.30 (0.62)}                 & \textbf{55.46 (0.61)}            & \textbf{78.10 (0.89)}              & 59.16 (0.58)      \\ \hline
		\end{tabular}
	}
\end{table*}

\begin{table}[t!]
	\centering
	\caption{Mean AUC and standard deviation at the last time step.}
	\label{tab:perfReal_auc}
	\resizebox{\columnwidth}{!}{%
		\begin{tabular}{llllll}
			\hline
			& \textbf{Keystroke}     & \textbf{Hand-written digits} & \textbf{uWave gestures} & \textbf{Starlight curves} & \textbf{Electromyography} \\ \hline
			\textbf{Baseline}     & 0.922 (0.015)          & 0.641 (0.037)                & 0.848 (0.020)           & 0.874 (0.020)             & 0.842 (0.016)             \\
			\textbf{NN}           & 0.962 (0.005)          & 0.968 (0.007)                & \textbf{0.870 (0.013)}  & 0.919 (0.006)             & \textbf{0.854 (0.013)}    \\
			\textbf{Siamese}      & 0.976 (0.003)          & 0.975 (0.005)                & 0.868 (0.012)           & \textbf{0.924 (0.010)}    & 0.817 (0.016)             \\
			\textbf{SiameseDuo++} & \textbf{0.983 (0.007)} & \textbf{0.984 (0.004)}       & 0.827 (0.009)           & \textbf{0.922 (0.018)}    & 0.783 (0.023)  \\ \hline           
		\end{tabular}
	}
\end{table}

\textbf{Quality monitoring of augmentation}. It has been observed that as the number of generated examples increases, either diminishing returns are observed or in some cases a smaller amount of augmentation is preferred. This is attributed to the fact that after some time, the generated examples do not contribute any new knowledge to the classifier, or worse, some generated examples do not reflect the true class distribution. Specialised mechanisms to monitor the quality of the generated examples should be proposed to alleviate this issue.

\textbf{Concept drift detection}. In this work two siamese neural networks are continually updated via incremental learning. In other words, they adapt to or learn the concept drift. Explicit mechanisms to detect drift would allow the siamese networks to train only when a change in the data distribution is detected.

\textbf{Hybrid learning paradigms}. The focus of this work has been on active stream learning. Future work will examine the potential benefits of combining in a seamless manner other paradigms, such as, semi-supervised learning.



  \bibliographystyle{elsarticle-num} 
  \bibliography{paper}

\begin{thebibliography}{10}
\expandafter\ifx\csname url\endcsname\relax
  \def\url#1{\texttt{#1}}\fi
\expandafter\ifx\csname urlprefix\endcsname\relax\def\urlprefix{URL }\fi
\expandafter\ifx\csname href\endcsname\relax
  \def\href#1#2{#2} \def\path#1{#1}\fi

\bibitem{kyriakides2014intelligent}
E.~Kyriakides, M.~Polycarpou (Eds.), Intelligent monitoring, control, and
  security of critical infrastructure systems, Vol. 565, Springer, 2014.

\bibitem{chang2023attention}
J.~Chang, J.~Wang, B.~Li, Y.~Zhao, D.~Li, Attention-based deep reinforcement
  learning for edge user allocation, IEEE Transactions on Network and Service
  Management (2023).

\bibitem{dal2015credit}
A.~Dal~Pozzolo, G.~Boracchi, O.~Caelen, C.~Alippi, G.~Bontempi, Credit card
  fraud detection and concept-drift adaptation with delayed supervised
  information, in: 2015 international joint conference on Neural networks
  (IJCNN), IEEE, 2015, pp. 1--8.

\bibitem{wang2018systematic}
S.~Wang, L.~L. Minku, X.~Yao, A systematic study of online class imbalance
  learning with concept drift, IEEE Transactions on Neural Networks and
  Learning Systems 29~(10) (2018) 4802--4821.

\bibitem{liu2022multi}
H.~Liu, C.~Zheng, D.~Li, Z.~Zhang, K.~Lin, X.~Shen, N.~N. Xiong, J.~Wang,
  Multi-perspective social recommendation method with graph representation
  learning, Neurocomputing 468 (2022) 469--481.

\bibitem{ditzler2015learning}
G.~Ditzler, M.~Roveri, C.~Alippi, R.~Polikar, Learning in nonstationary
  environments: A survey, IEEE Computational Intelligence Magazine 10~(4)
  (2015) 12--25.

\bibitem{liu2024integrating}
B.~Liu, D.~Li, J.~Wang, Z.~Wang, B.~Li, C.~Zeng, Integrating user short-term
  intentions and long-term preferences in heterogeneous hypergraph networks for
  sequential recommendation, Information Processing \& Management 61~(3) (2024)
  103680.

\bibitem{li2024homogeneous}
D.~Li, Y.~Gao, Z.~Wang, H.~Qiu, P.~Liu, Z.~Xiong, Z.~Zhang, Homogeneous graph
  neural networks for third-party library recommendation, Information
  Processing \& Management 61~(6) (2024) 103831.

\bibitem{gama2014survey}
J.~Gama, I.~{\v{Z}}liobait{\.e}, A.~Bifet, M.~Pechenizkiy, A.~Bouchachia, A
  survey on concept drift adaptation, ACM Computing Surveys 46~(4) (2014) 44.

\bibitem{lu2018learning}
J.~Lu, A.~Liu, F.~Dong, F.~Gu, J.~Gama, G.~Zhang, Learning under concept drift:
  A review, IEEE Transactions on Knowledge and Data Engineering 31~(12) (2018)
  2346--2363.

\bibitem{settles2009active}
B.~Settles, Active learning literature survey, Tech. rep., University of
  Wisconsin-Madison Department of Computer Sciences (2009).

\bibitem{nvidia}
NVIDIA-AI, \href{https://medium.com/nvidia-ai/}{Scalable active learning for
  autonomous driving}, accessed 21 Sep. 2023.
\newline\urlprefix\url{https://medium.com/nvidia-ai/}

\bibitem{sculley2011detecting}
D.~Sculley, M.~E. Otey, M.~Pohl, B.~Spitznagel, J.~Hainsworth, Y.~Zhou,
  Detecting adversarial advertisements in the wild, in: Proceedings of the 17th
  ACM SIGKDD international conference on Knowledge discovery and data mining,
  ACM, 2011, pp. 274--282.

\bibitem{zliobaite2013active}
I.~{\v{Z}}liobait{\.e}, A.~Bifet, B.~Pfahringer, G.~Holmes, Active learning
  with drifting streaming data, IEEE Transactions on Neural Networks and
  Learning Systems 25~(1) (2013) 27--39.

\bibitem{malialis2022nonstationary}
K.~Malialis, C.~G. Panayiotou, M.~M. Polycarpou, Nonstationary data stream
  classification with online active learning and siamese neural networks,
  Neurocomputing 512 (2022) 235--252.

\bibitem{losing2018incremental}
V.~Losing, B.~Hammer, H.~Wersing, Incremental on-line learning: A review and
  comparison of state of the art algorithms, Neurocomputing 275 (2018)
  1261--1274.

\bibitem{vzliobaite2015towards}
I.~{\v{Z}}liobait{\.e}, M.~Budka, F.~Stahl, Towards cost-sensitive adaptation:
  When is it worth updating your predictive model?, Neurocomputing 150 (2015)
  240--249.

\bibitem{mai2022online}
Z.~Mai, R.~Li, J.~Jeong, D.~Quispe, H.~Kim, S.~Sanner, Online continual
  learning in image classification: An empirical survey, Neurocomputing 469
  (2022) 28--51.

\bibitem{gunasekara2023survey}
N.~Gunasekara, B.~Pfahringer, H.~M. Gomes, A.~Bifet, Survey on online streaming
  continual learning., in: International Joint Conference on Artificial
  IntelligenceI, 2023, pp. 6628--6637.

\bibitem{dyer2013compose}
K.~B. Dyer, R.~Capo, R.~Polikar, Compose: A semisupervised learning framework
  for initially labeled nonstationary streaming data, IEEE Transactions on
  Neural Networks and Learning Systems 25~(1) (2013) 12--26.

\bibitem{khezri2021novel}
S.~Khezri, J.~Tanha, A.~Ahmadi, A.~Sharifi, A novel semi-supervised ensemble
  algorithm using a performance-based selection metric to non-stationary data
  streams, Neurocomputing 442 (2021) 125--145.

\bibitem{khezri2020stds}
S.~Khezri, J.~Tanha, A.~Ahmadi, A.~Sharifi, Stds: self-training data streams
  for mining limited labeled data in non-stationary environment, Applied
  Intelligence 50 (2020) 1448--1467.

\bibitem{yu2022learn}
E.~Yu, Y.~Song, G.~Zhang, J.~Lu, Learn-to-adapt: Concept drift adaptation for
  hybrid multiple streams, Neurocomputing 496 (2022) 121--130.

\bibitem{wang2021survey}
X.~Wang, Y.~Chen, W.~Zhu, A survey on curriculum learning, IEEE transactions on
  pattern analysis and machine intelligence 44~(9) (2021) 4555--4576.

\bibitem{alippi2008justI}
C.~Alippi, M.~Roveri, Just-in-time adaptive classifiers—part i: Detecting
  nonstationary changes, IEEE Transactions on Neural Networks 19~(7) (2008)
  1145--1153.

\bibitem{baena2006early}
M.~Baena-Garc{\i}a, J.~del Campo-{\'A}vila, R.~Fidalgo, A.~Bifet, R.~Gavalda,
  R.~Morales-Bueno, Early drift detection method, in: Fourth International
  Workshop on Knowledge Discovery from Data Streams, Vol.~6, 2006, pp. 77--86.

\bibitem{krawczyk2017ensemble}
B.~Krawczyk, L.~L. Minku, J.~Gama, J.~Stefanowski, M.~Wo{\'z}niak, Ensemble
  learning for data stream analysis: A survey, Information Fusion 37 (2017)
  132--156.

\bibitem{gomes2017survey}
H.~M. Gomes, J.~P. Barddal, F.~Enembreck, A.~Bifet, A survey on ensemble
  learning for data stream classification, ACM Computing Surveys (CSUR) 50~(2)
  (2017) 1--36.

\bibitem{aguiar2022survey}
G.~Aguiar, B.~Krawczyk, A.~Cano, A survey on learning from imbalanced data
  streams: taxonomy, challenges, empirical study, and reproducible experimental
  framework, Machine learning (2023) 1--79.

\bibitem{malialis2018queue}
K.~Malialis, C.~Panayiotou, M.~M. Polycarpou, Queue-based resampling for online
  class imbalance learning, in: International Conference on Artificial Neural
  Networks (ICANN), Springer, 2018, pp. 498--507.
\newblock \href {https://doi.org/10.1007/978-3-030-01418-6\_49}
  {\path{doi:10.1007/978-3-030-01418-6\_49}}.

\bibitem{malialis2020online}
K.~Malialis, C.~G. Panayiotou, M.~M. Polycarpou, Online learning with adaptive
  rebalancing in nonstationary environments, IEEE Transactions on Neural
  Networks and Learning Systems 32~(10) (2021) 4445--4459.
\newblock \href {https://doi.org/10.1109/TNNLS.2020.3017863}
  {\path{doi:10.1109/TNNLS.2020.3017863}}.

\bibitem{wang2015resampling}
S.~Wang, L.~L. Minku, X.~Yao, Resampling-based ensemble methods for online
  class imbalance learning, IEEE Transactions on Knowledge and Data Engineering
  27~(5) (2015) 1356--1368.

\bibitem{cano2022rose}
A.~Cano, B.~Krawczyk, Rose: robust online self-adjusting ensemble for continual
  learning on imbalanced drifting data streams, Machine Learning (2022) 1--39.

\bibitem{siahroudi2021online}
S.~K. Siahroudi, D.~Kudenko, An online learning algorithm for non-stationary
  imbalanced data by extra-charging minority class, in: Pacific-Asia Conference
  on Knowledge Discovery and Data Mining, Springer, 2021, pp. 603--615.

\bibitem{mirza2015ensemble}
B.~Mirza, Z.~Lin, N.~Liu, Ensemble of subset online sequential extreme learning
  machine for class imbalance and concept drift, Neurocomputing 149 (2015)
  316--329.

\bibitem{ren2018gradual}
S.~Ren, B.~Liao, W.~Zhu, Z.~Li, W.~Liu, K.~Li, The gradual resampling ensemble
  for mining imbalanced data streams with concept drift, Neurocomputing 286
  (2018) 150--166.
\newblock \href {https://doi.org/https://doi.org/10.1016/j.neucom.2018.01.063}
  {\path{doi:https://doi.org/10.1016/j.neucom.2018.01.063}}.

\bibitem{elwell2011incremental}
R.~Elwell, R.~Polikar, Incremental learning of concept drift in nonstationary
  environments, IEEE Transactions on Neural Networks 22~(10) (2011) 1517--1531.

\bibitem{malialis2022hybrid}
K.~Malialis, M.~Roveri, C.~Alippi, C.~G. Panayiotou, M.~M. Polycarpou, A hybrid
  active-passive approach to imbalanced nonstationary data stream
  classification, in: 2022 IEEE Symposium Series on Computational Intelligence
  (SSCI), IEEE, 2022, pp. 1021--1027.

\bibitem{alippi2017learning}
C.~Alippi, W.~Qi, M.~Roveri, Learning in nonstationary environments: A hybrid
  approach, in: International Conference on Artificial Intelligence and Soft
  Computing, Springer, 2017, pp. 703--714.

\bibitem{artelt2022unsupervised}
A.~Artelt, K.~Malialis, C.~G. Panayiotou, M.~M. Polycarpou, B.~Hammer,
  Unsupervised unlearning of concept drift with autoencoders, in: 2023 IEEE
  Symposium Series on Computational Intelligence (SSCI), IEEE, 2023, pp.
  703--710.

\bibitem{li2023autoencoder}
J.~Li, K.~Malialis, M.~M. Polycarpou, Autoencoder-based anomaly detection in
  streaming data with incremental learning and concept drift adaptation, in:
  2023 International Joint Conference on Neural Networks (IJCNN), 2023, pp.
  1--8.

\bibitem{li2024unsupervised}
J.~Li, K.~Malialis, C.~G. Panayiotou, M.~M. Polycarpou, Unsupervised
  incremental learning with dual concept drift detection for identifying
  anomalous sequences (2024).
\newblock \href {http://arxiv.org/abs/2403.03576} {\path{arXiv:2403.03576}}.

\bibitem{malialis2020data}
K.~Malialis, C.~G. Panayiotou, M.~M. Polycarpou, Data-efficient online
  classification with siamese networks and active learning, in: 2020
  International Joint Conference on Neural Networks (IJCNN), IEEE, 2020, pp.
  1--7.

\bibitem{martins2023meta}
V.~E. Martins, A.~Cano, S.~B. Junior, Meta-learning for dynamic tuning of
  active learning on stream classification, Pattern Recognition 138 (2023)
  109359.

\bibitem{koch2015siamese}
G.~Koch, R.~Zemel, R.~Salakhutdinov, Siamese neural networks for one-shot image
  recognition, in: ICML Deep Learning Workshop, Vol.~2, 2015.

\bibitem{ienco2014high}
D.~Ienco, I.~{\v{Z}}liobait{\.e}, B.~Pfahringer, High density-focused
  uncertainty sampling for active learning over evolving stream data, in: 3rd
  International Workshop on Big Data, Streams and Heterogeneous Source Mining:
  Algorithms, Systems, Programming Models and Applications, 2014.

\bibitem{liu2021online}
S.~Liu, S.~Xue, J.~Wu, C.~Zhou, J.~Yang, Z.~Li, J.~Cao, Online active learning
  for drifting data streams, IEEE Transactions on Neural Networks and Learning
  Systems 34~(1) (2023) 186--200.

\bibitem{yan2021clustering}
X.~Yan, A.~Homaifar, M.~Sarkar, A.~Girma, E.~Tunstel, A clustering-based
  framework for classifying data streams, in: International Joint Conference on
  Artificial Conference, 2021.

\bibitem{freund1997selective}
Y.~Freund, H.~S. Seung, E.~Shamir, N.~Tishby, Selective sampling using the
  query by committee algorithm, Machine Learning 28~(2-3) (1997) 133--168.

\bibitem{korycki2019active}
L.~Korycki, A.~Cano, B.~Krawczyk, Active learning with abstaining classifiers
  for imbalanced drifting data streams, in: IEEE International Conference on
  Big Data, 2019, pp. 2334--2343.

\bibitem{wozniak2023active}
M.~Wo{\'z}niak, P.~Zyblewski, P.~Ksieniewicz, Active weighted aging ensemble
  for drifted data stream classification, Information Sciences 630 (2023)
  286--304.

\bibitem{wu2020comprehensive}
Z.~Wu, S.~Pan, F.~Chen, G.~Long, C.~Zhang, S.~Y. Philip, A comprehensive survey
  on graph neural networks, IEEE transactions on neural networks and learning
  systems 32~(1) (2020) 4--24.

\bibitem{cao2024graph}
H.~Cao, S.~Du, J.~Hu, Y.~Yang, S.-J. Horng, T.~Li, Graph deep active learning
  framework for data deduplication, Big Data Mining and Analytics 7~(3) (2024)
  753--764.

\bibitem{huang2024adaptive}
Y.~Huang, Y.~Pi, Y.~Shi, W.~Guo, S.~Wang, Adaptive graph active learning with
  mutual information via policy learning, Expert Systems with Applications 255
  (2024) 124773.

\bibitem{ge2025iterative}
Y.~Ge, D.~Yang, A.~L. Bertozzi, Iterative active learning strategies for
  subgraph matching, Pattern Recognition 158 (2025) 110797.

\bibitem{shorten2019survey}
C.~Shorten, T.~M. Khoshgoftaar, A survey on image data augmentation for deep
  learning, J. of Big Data 6~(1) (2019) 1--48.

\bibitem{taylor2018improving}
L.~Taylor, G.~Nitschke, Improving deep learning with generic data augmentation,
  in: IEEE Symposium Series on Computational Intelligence, IEEE, 2018, pp.
  1542--1547.

\bibitem{zhong2020random}
Z.~Zhong, L.~Zheng, G.~Kang, A.~Li, Y.~Yang, Random erasing data augmentation,
  in: AAAI Conference on Artificial Intelligence, Vol.~34, 2020, pp.
  13001--13008.

\bibitem{jackson2019style}
P.~T.~G. Jackson, A.~A. Abarghouei, S.~Bonner, T.~P. Breckon, B.~Obara, Style
  augmentation: data augmentation via style randomization, in: CVPR Workshop,
  Vol.~6, 2019, pp. 10--11.

\bibitem{inoue2018data}
H.~Inoue, Data augmentation by pairing samples for images classification, arXiv
  preprint arXiv:1801.02929 (2018).

\bibitem{guennec2016data}
L.~A. Guennec, S.~Malinowski, R.~Tavenard, Data augmentation for time series
  classification using convolutional neural networks, in: ECML/PKDD Workshop on
  Advanced Analytics and Learning on Temporal Data, 2016.

\bibitem{umtt2017data}
T.~T. Um, F.~M.~J. Pfister, D.~Pichler, S.~Endo, M.~Lang, S.~Hirche,
  U.~Fietzek, D.~Kulić, Data augmentation of wearable sensor data for
  parkinson’s disease monitoring using convolutional neural networks, in:
  19th ACM International Conference on Multimodal Interaction, 2017, pp.
  216--220.

\bibitem{cheung2021modals}
T.~Cheung, D.~Yeung, Modals: Modality-agnostic automated data augmentation in
  the latent space, in: International Conference on Learning Representations,
  2021.

\bibitem{kumar2019closer}
V.~Kumar, H.~Glaude, C.~de~Lichy, W.~Campbell, A closer look at feature space
  data augmentation for few-shot intent classification, in: 2nd Workshop on
  Deep Learning Approaches for Low-Resource NLP (DeepLo), 2019, pp. 1--10.

\bibitem{devries2017dataset}
T.~DeVries, G.~W. Taylor, Dataset augmentation in feature space, arXiv preprint
  arXiv:1702.05538 (2017).

\bibitem{tran2019bayesian}
T.~Tran, T.~Do, I.~Reid, G.~Carneiro, Bayesian generative active deep learning,
  in: International Conference on Machine Learning, 2019, pp. 6295--6304.

\bibitem{hong2020deep}
S.~Hong, H.~Ha, J.~Kim, M.~Choi, Deep active learning with augmentation-based
  consistency estimation, arXiv preprint arXiv:2011.02666 (2020).

\bibitem{kim2021lada}
Y.~Kim, K.~Song, J.~Jang, I.~Moon, Lada: Look-ahead data acquisition via
  augmentation for deep active learning, Advances in Neural Information
  Processing Systems 34 (2021) 22919--22930.

\bibitem{malialis2022data}
K.~Malialis, D.~Papatheodoulou, S.~Filippou, C.~G. Panayiotou, M.~M.
  Polycarpou, Data augmentation on-the-fly and active learning in data stream
  classification, in: 2022 IEEE Symposium Series on Computational Intelligence
  (SSCI), IEEE, 2022, pp. 1408--1414.

\bibitem{rumelhart1986learning}
D.~E. Rumelhart, G.~E. Hinton, R.~J. Williams, Learning representations by
  back-propagating errors, nature 323~(6088) (1986) 533--536.

\bibitem{wang2012multiclass}
S.~Wang, X.~Yao, Multiclass imbalance problems: Analysis and potential
  solutions, IEEE Transactions on Systems, Man, and Cybernetics, Part B
  (Cybernetics) 42~(4) (2012) 1119--1130.

\bibitem{souza2015data}
V.~M.~A. Souza, D.~F. Silva, J.~Gama, G.~E. A. P.~A. Batista, Data stream
  classification guided by clustering on nonstationary environments and extreme
  verification latency, in: Proceedings of the 2015 SIAM International
  Conference on Data Mining, 2015, pp. 873--881.

\bibitem{misc_optical_recognition_of_handwritten_digits_80}
E.~Alpaydin, C.~Kaynak, {Optical Recognition of Handwritten Digits}, UCI
  Machine Learning Repository, {DOI}: https://doi.org/10.24432/C50P49 (1998).

\bibitem{LIU2009657}
J.~Liu, L.~Zhong, J.~Wickramasuriya, V.~Vasudevan, uwave: Accelerometer-based
  personalized gesture recognition and its applications, Pervasive and Mobile
  Computing 5~(6) (2009) 657--675, perCom 2009.
\newblock \href {https://doi.org/https://doi.org/10.1016/j.pmcj.2009.07.007}
  {\path{doi:https://doi.org/10.1016/j.pmcj.2009.07.007}}.

\bibitem{Rebbapragada_2008}
U.~Rebbapragada, P.~Protopapas, C.~E. Brodley, C.~Alcock,
  \href{https://doi.org/10.1007%2Fs10994-008-5093-3}{Finding anomalous periodic
  time series}, Machine Learning 74~(3) (2008) 281--313.
\newblock \href {https://doi.org/10.1007/s10994-008-5093-3}
  {\path{doi:10.1007/s10994-008-5093-3}}.
\newline\urlprefix\url{https://doi.org/10.1007%2Fs10994-008-5093-3}

\bibitem{gestures}
K.~Yashuk, \href{https://www.kaggle.com/kyr7plus/emg-4}{Classify gestures by
  reading muscle activity}, accessed 26 Jan, 2022.
\newline\urlprefix\url{https://www.kaggle.com/kyr7plus/emg-4}

\bibitem{he2015delving}
K.~He, X.~Zhang, S.~Ren, J.~Sun, Delving deep into rectifiers: Surpassing
  human-level performance on imagenet classification, in: Proceedings of the
  IEEE International Conference on Computer Vision, 2015, pp. 1026--1034.

\bibitem{kingma2014adam}
D.~P. Kingma, J.~Ba, Adam: A method for stochastic optimization, in:
  Proceedings of the 3rd International Conference on Learning Representations
  (ICLR), 2015.

\bibitem{maas2013rectifier}
A.~L. Maas, A.~Y. Hannun, A.~Y. Ng, Rectifier nonlinearities improve neural
  network acoustic models, in: Proceedings of the 30th International Conference
  on Machine Learning, 2013.

\bibitem{he2008learning}
H.~He, E.~A. Garcia, Learning from imbalanced data, IEEE Transactions on
  Knowledge and Data Engineering~(9) (2008) 1263--1284.

\bibitem{sun2006boosting}
Y.~Sun, M.~S. Kamel, Y.~Wang, Boosting for learning multiple classes with
  imbalanced class distribution, in: Sixth International Conference on Data
  Mining (ICDM'06), IEEE, 2006, pp. 592--602.

\bibitem{gama2013evaluating}
J.~Gama, R.~Sebasti{\~a}o, P.~P. Rodrigues, On evaluating stream learning
  algorithms, Machine Learning 90~(3) (2013) 317--346.

\bibitem{aguiar2024survey}
G.~Aguiar, B.~Krawczyk, A.~Cano, A survey on learning from imbalanced data
  streams: taxonomy, challenges, empirical study, and reproducible experimental
  framework, Machine learning 113~(7) (2024) 4165--4243.

\bibitem{wang2020auc}
S.~Wang, L.~L. Minku, Auc estimation and concept drift detection for imbalanced
  data streams with multiple classes, in: 2020 international joint conference
  on neural networks (IJCNN), 2020, pp. 1--8.

\end{thebibliography}


%
%
%
\end{document}